\documentclass[oneside,10pt]{article}

\usepackage{times,url,mathrsfs}
\usepackage{amsmath}
\usepackage{amsfonts}
\usepackage{graphicx}
\usepackage{subfigure}
\newtheorem{theorem}{Theorem}
\newtheorem{lemma}{Lemma}

\begin{document}
\title{Hashing Algorithms for Large-Scale Learning}

\author{ Ping Li \\
         Department of Statistical Science\\
       Cornell University\\
         Ithaca, NY 14853\\
       pingli@cornell.edu
       \and
       Anshumali Shrivastava\\
         Department of Computer Science\\
         Cornell University\\
         Ithaca, NY 14853\\
         anshu@cs.cornell.edu
       \vspace{0.2in}\and
       Joshua Moore\\
         Department of Computer Science\\
         Cornell University\\
         Ithaca, NY 14853\\
         jlmo@cs.cornell.edu
         \and
         Arnd Christian K\"{o}nig\\
         Microsoft Research\\
        Microsoft Corporation\\
        Redmond, WA 98052\\
        chrisko@microsoft.com
        }

\date{}
\maketitle

\begin{abstract}

\noindent In this paper, we first demonstrate that  $b$-bit minwise hashing, whose estimators are  positive definite kernels, can be naturally integrated with learning algorithms such as SVM and logistic regression. We adopt a simple scheme to transform the nonlinear (resemblance) kernel into linear (inner product) kernel; and hence large-scale  problems can be solved extremely efficiently. Our method provides a simple effective solution to large-scale learning in massive and extremely high-dimensional datasets, especially when data do not fit in memory.\\

\noindent We then compare $b$-bit minwise hashing with the Vowpal Wabbit (VW) algorithm (which is related the Count-Min (CM) sketch). Interestingly, VW has the same variances as random projections. Our theoretical and empirical comparisons illustrate that usually $b$-bit minwise hashing is significantly more accurate (at the same storage) than VW (and random projections) in binary data.  Furthermore, $b$-bit minwise hashing can be combined with VW to achieve further improvements in terms of training speed, especially when $b$ is large.

\end{abstract}

\section{Introduction}

With the advent of the Internet, many machine learning applications are faced with very large and inherently high-dimensional datasets, resulting in challenges in scaling up training algorithms and storing the data.
Especially in the context of search and machine translation, corpus sizes used in industrial practice have long exceeded the main memory capacity of single machine. For example, \cite{Proc:Weinberger_ICML2009} experimented with a dataset with potentially 16 trillion ($1.6\times10^{13}$) unique features. \cite{GoogleBlog} discusses training sets with (on average) $10^{11}$ items and $10^9$ distinct features, requiring novel algorithmic approaches and architectures.  As a consequence, there has been a renewed emphasis on scaling up machine learning techniques by using massively parallel architectures; however, methods relying solely on parallelism can be expensive (both with regards to hardware requirements and energy costs) and often induce significant additional communication and data distribution overhead.

This work approaches the challenges posed by  large datasets  by leveraging techniques from the area of {\em similarity search}\/~\cite{Article:Andoni_CACM08}, where similar increase in dataset sizes has made the storage and computational requirements for computing exact distances prohibitive, thus making data representations that allow compact storage and efficient approximate distance computation necessary.

The method of $b$-bit minwise hashing~\cite{Proc:Li_Konig_WWW10,Proc:Li_Konig_NIPS10,Article:Li_Konig_CACM11} is a very recent progress for efficiently (in both time and space) computing  {\em resemblances} among extremely high-dimensional (e.g., $2^{64}$) binary vectors. In this paper, we show that $b$-bit minwise hashing can be seamlessly integrated with linear Support Vector Machine (SVM)~\cite{Proc:Joachims_KDD06,Proc:Shalev-Shwartz_ICML07,Article:Fan_JMLR08,Proc:Hsieh_ICML08,Proc:Yu_KDD10} and logistic regression solvers. In~\cite{Proc:Yu_KDD10}, the authors addressed a critically important problem of training linear SVM when the data can not fit in memory. In this paper, our work also addresses the same problem using a very different approach.

\subsection{Ultra High-Dimensional Large Datasets and Memory Bottleneck }

In the context of search, a standard procedure to represent documents (e.g., Web pages) is to use $w$-shingles (i.e., $w$ contiguous words), where  $w$ can be as large as 5 (or 7) in several studies~\cite{Proc:Broder,Proc:Broder_WWW97,Proc:Fetterly_WWW03}. This procedure can generate datasets of extremely high dimensions. For example, suppose we only consider $10^5$ common English words. Using $w=5$ may require the size of dictionary $\Omega$ to be $D=|\Omega| = 10^{25} = 2^{83}$. In  practice,  $D=2^{64}$ often suffices, as the number of available documents may not be large enough to exhaust the dictionary. For $w$-shingle data, normally only  abscence/presence (0/1) information is used, as it is known that word frequency distributions within documents approximately follow a power-law~\cite{Baayen}, meaning that most single terms occur rarely, thereby making a $w$-shingle unlikely to occur more than once in a document. Interestingly, even when the data  are not too high-dimensional,  empirical studies~\cite{Article:Chapelle_99,Proc:Hein_AISTATS05,Proc:Jiang_CIVR07} achieved good performance with binary-quantized data.

When the data can fit in memory, linear SVM is often extremely efficient after the data are loaded into the memory. It is however often the case that, for very large datasets, the data loading time dominates the computing time for training the SVM~\cite{Proc:Yu_KDD10}. A much more severe problem arises when the data can not fit in memory. This situation can be very common in  practice. The publicly available {\em webspam} dataset needs about 24GB disk space (in LIBSVM input data format), which exceeds the memory capacity of many desktop PCs. Note that {\em webspam}, which contains only 350,000 documents represented by 3-shingles, is still a small dataset compared to the industry applications~\cite{GoogleBlog}.

\subsection{A Brief Introduction of Our Proposal}

We propose a  solution which leverages {\em b-bit minwise hashing}.  Our approach  assume the data vectors are binary, very high-dimensional, and relatively sparse, which is generally true of text documents represented via shingles. We  apply $b$-bit minwise hashing  to obtain a  compact representation of the original data.  In order to use the technique for efficient learning, we have to address several  issues:
\begin{itemize}
\item We need to prove  that the matrices generated by $b$-bit minwise hashing are indeed positive definite, which will provide the solid foundation for our proposed solution.
\item If we use $b$-bit minwise hashing to estimate the resemblance, which is nonlinear, how can we effectively convert this nonlinear problem into a linear problem?
\item Compared to other hashing techniques such as random projections, Count-Min (CM) sketch~\cite{Article:Cormode_05}, or  Vowpal Wabbit (VW)~\cite{Proc:Weinberger_ICML2009}, does our approach exhibits advantages?
\end{itemize}
It turns out that our proof in the next section that {$b$-bit hashing matrices} are positive definite  naturally provides the construction for converting the otherwise nonlinear SVM problem into linear SVM.\\

\cite{Proc:Yu_KDD10} proposed solving the memory bottleneck by partitioning the data into  blocks, which are repeatedly loaded into memory as their approach updates the model coefficients. However, the computational bottleneck is still at the memory because loading the data blocks for many iterations consumes a large number of disk I/Os. Clearly, one should note that our method is not really a competitor of the approach in~\cite{Proc:Yu_KDD10}. In fact, both approaches may work together to solve extremely large problems.

\section{Review Minwise Hashing and b-Bit Minwise Hashing}\label{sec_minwise}

{\em Minwise hashing}~\cite{Proc:Broder,Proc:Broder_WWW97} has been successfully applied to a very wide range of real-world problems especially in the context of search~\cite{Proc:Broder,Proc:Broder_WWW97,Proc:Bendersky_WSDM09,Article:Forman09,Proc:Cherkasova_KDD09,Proc:Buehrer_WSDM08,Article:Urvoy08,Proc:Nitin_WSDM08,
Article:Dourisboure09,Proc:Chierichetti_KDD09,Proc:Gollapudi_WWW09,Article:Kalpakis08,Proc:Najork_WSDM09}, for efficiently computing set similarities.

 Minwise hashing mainly works well with binary data, which can be viewed either as 0/1  vectors or as sets. Given two sets, $S_1, \  S_2 \subseteq \Omega = \{0, 1, 2, ..., D-1\}$, a widely used (normalized)  measure of similarity is the {\em resemblance} $R$:
\begin{align}\notag
&R = \frac{|S_1 \cap S_2|}{|S_1 \cup S_2|} = \frac{a}{f_1 + f_2 - a}, \hspace{0.3in}
\text{where}\ \  f_1 = |S_1|, \  f_2 = |S_2|, \ a = |S_1\cap S_2|.
\end{align}
In this method, one applies a random permutation $\pi: \Omega\rightarrow\Omega$ on $S_1$ and $S_2$. The collision probability is simply
\begin{align}
\mathbf{Pr}\left(\text{min}({\pi}(S_1)) = \text{min}({\pi}(S_2)) \right) = \frac{|S_1
  \cap S_2|}{|S_1 \cup S_2|}=R.
\end{align}
One can repeat the permutation $k$ times: $\pi_1$, $\pi_2$, ..., $\pi_k$ to estimate $R$ without bias, as
 \begin{align}
&\hat{R}_{M} = \frac{1}{k}\sum_{j=1}^{k}1\{{\min}({\pi_j}(S_1)) =
  {\min}({\pi_j}(S_2))\}, \\\label{eqn_Var_M}
&\text{Var}\left(\hat{R}_{M}\right) = \frac{1}{k}R(1-R).
\end{align}

The common practice of minwise hashing is to store each hashed value, e.g., ${\min}({\pi}(S_1))$ and ${\min}({\pi}(S_2))$, using 64 bits~\cite{Proc:Fetterly_WWW03}. The storage (and computational) cost  will be prohibitive in  truly large-scale (industry) applications~\cite{Proc:Manku_WWW07}.\\

{\em b-bit minwise hashing}~\cite{Proc:Li_Konig_WWW10,Proc:Li_Konig_NIPS10,Article:Li_Konig_CACM11} provides a strikingly simple solution to this (storage and computational) problem  by storing only the lowest $b$ bits (instead of 64 bits) of each hashed value. For convenience, we define the minimum values under $\pi$: $z_1 = \min\left(\pi\left(S_1\right)\right)$ and $z_2 = \min\left(\pi\left(S_2\right)\right)$, and define $e_{1,i} = i$th lowest bit of $z_1$, and $e_{2,i} = i$th lowest bit of $z_2$.
\begin{theorem}\cite{Proc:Li_Konig_WWW10}\label{The_basic}
Assume $D$ is large.
\begin{align}\label{eqn_basic}
&P_b=\mathbf{Pr}\left(\prod_{i=1}^b1\left\{e_{1,i} = e_{2,i}\right\}\right) = C_{1,b} + \left(1-C_{2,b}\right) R\\\notag
&r_1 = \frac{f_1}{D}, \hspace{0.1in} r_2 = \frac{f_2}{D}, \ \ f_1 = |S_1|,\  \ f_2 =|S_2|\\\notag
&C_{1,b} = A_{1,b} \frac{r_2}{r_1+r_2} + A_{2,b}\frac{r_1}{r_1+r_2},\hspace{0.3in} \ C_{2,b} = A_{1,b} \frac{r_1}{r_1+r_2} + A_{2,b}\frac{r_2}{r_1+r_2},\\\notag
&A_{1,b} = \frac{r_1\left[1-r_1\right]^{2^b-1}}{1-\left[1-r_1\right]^{2^b}},\hspace{1in}
A_{2,b} = \frac{r_2\left[1-r_2\right]^{2^b-1}}{1-\left[1-r_2\right]^{2^b}}.\Box
\end{align}
\end{theorem}
This (approximate)  formula (\ref{eqn_basic}) is remarkably accurate, even for very small $D$. Some numerical comparisons with the exact probabilities are provided in Appendix~\ref{app_basic_error}.

We can then estimate $P_b$ (and $R$) from $k$ independent permutations: $\pi_1$, $\pi_2$, ..., $\pi_k$,
\begin{align}\label{eqn_R_b}
&\hat{R}_b = \frac{\hat{P}_b - C_{1,b}}{1-C_{2,b}},\hspace{0.6in} \hat{P}_{b} = \frac{1}{k}\sum_{j=1}^{k}\left\{ \prod_{i=1}^b1\{e_{1,i,\pi_j} = e_{2,i,\pi_j}\}\right\},\\\label{eqn_Var_b}
&\text{Var}\left(\hat{R}_b\right) = \frac{\text{Var}\left(\hat{P}_b\right)}{\left[1-C_{2,b}\right]^2}
=\frac{1}{k}\frac{\left[C_{1,b}+(1-C_{2,b})R\right]\left[1-C_{1,b}-(1-C_{2,b})R\right]}{\left[1-C_{2,b}\right]^2}
\end{align}
We will show that we can apply $b$-bit  hashing for learning without explicitly estimating $R$ from  (\ref{eqn_basic}).

\section{Kernels from Minwise Hashing and $b$-Bit Minwise Hashing}

This section proves some theoretical properties of matrices generated by resemblance, minwise hashing, or $b$-bit minwise hashing, which are all positive definite matrices. Our proof not only provides a solid theoretical foundation for using $b$-bit hashing in learning, but also illustrates our idea behind the construction  for integrating $b$-bit hashing with linear learning algorithms.\\

\noindent\textbf{Definition}: A symmetric $n\times n$ matrix $\mathbf{K}$ satisfying $\sum_{ij}c_ic_jK_{ij}\geq0$,
for all real vectors $c$ is called {\em positive definite (PD)}. Note that here we do not differentiate PD from {\em nonnegative definite}.

\begin{theorem}\label{thm_PD}
Consider $n$ sets $S_1$, $S_2$, ..., $S_n\in\Omega=\{0, 1, ..., D-1\}$. Apply one permutation $\pi$ to each set and define $z_i = \min\{\pi(S_i)\}$. The following three matrices are all PD.
\begin{enumerate}
\item
The {\em resemblance matrix} $\mathbf{R}\in\mathbb{R}^{n\times n}$, whose $(i,j)$-th entry is the resemblance between set $S_i$ and set $S_j$:
$R_{ij} = \frac{|S_i\cap S_j|}{|S_i\cup S_j|} = \frac{|S_i\cap S_j|}{|S_i| + |S_j|-|S_i\cap S_j|}$

\item The {\em minwise hashing matrix} $\mathbf{M}\in\mathbb{R}^{n\times n}$: $M_{ij} = 1\{z_i = z_j\}$
\item The {\em b-bit minwise hashing matrix} $\mathbf{M}^{(b)} \in \mathbb{R}^{n\times n}$:
$M^{(b)}_{ij} = \prod_{t=1}^b1\left\{e_{i,t} = e_{j,t}\right\}$,
where $e_{i,t}$ is the $t$-th lowest bit of $z_i$.
\end{enumerate}

Consequently, consider $k$  independent permutations and denote $\mathbf{M}^{(b)}_{(s)}$ the {b-bit minwise hashing matrix} generated by the $s$-th permutation. Then the summation $\sum_{s=1}^k \mathbf{M}^{(b)}_{(s)}$ is also PD.\\

\textbf{Proof:}\  A matrix $\mathbf{A}$ is PD if it can be written as an inner product $\mathbf{B^TB}$. Because
\begin{align}
M_{ij} = 1\{z_i= z_j\} = \sum_{t=0}^{D-1} 1\{z_i = t\}\times1\{z_j = t\},
\end{align}
$M_{ij}$ is the inner product of two D-dim vectors. Thus, $\mathbf{M}$ is PD.

Similarly, the {b-bit minwise hashing matrix}  $\mathbf{M}^{(b)}$  is PD because
\begin{align}\label{eqn_M_b}
M^{(b)}_{ij} = \sum_{t=0}^{2^b-1} 1\{z_i = t\}\times1\{z_j = t\}.
\end{align}

The resemblance matrix $\mathbf{R}$  is PD because $R_{ij}= \mathbf{Pr}\{M_{ij} =1 \} = E\left(M_{ij}\right)$ and $M_{ij}$ is the $(i,j)$-th element of the PD matrix $\mathbf{M}$. Note that the expectation is a linear operation. $\Box$
\end{theorem}

Our proof that the $b$-bit minwise hashing matrix $\mathbf{M}^{(b)}$ is PD provides us with a  simple strategy to expand a nonlinear (resemblance) kernel
into a linear (inner product) kernel. After concatenating the $k$ vectors resulting from (\ref{eqn_M_b}), the new (binary) data vector after the expansion will be of dimension $2^b\times k$ with exactly $k$ ones.

\section{Integrating  $b$-Bit Minwise Hashing with (Linear) Learning Algorithms}

Linear algorithms such as linear SVM and logistic regression have become very powerful and extremely popular. Representative software packages include SVM$^\text{perf}$~\cite{Proc:Joachims_KDD06}, Pegasos~\cite{Proc:Shalev-Shwartz_ICML07}, Bottou's SGD SVM~\cite{URL:Bottou_SGD}, and LIBLINEAR~\cite{Article:Fan_JMLR08}.

Given a dataset $\{(\mathbf{x}_i, y_i)\}_{i=1}^n$, $\mathbf{x}_i\in\mathbb{R}^{D}$, $y_i\in\{-1,1\}$, the $L_2$-regularized linear SVM solves the following optimization  problem:
\begin{align}
\min_{\mathbf{w}}\ \ \frac{1}{2}\mathbf{w^Tw} + C \sum_{i=1}^n \max \left\{1 - y_i\mathbf{w^Tx_i},\ 0\right\},
\end{align}
and the $L_2$-regularized logistic regression solves a  similar problem:
\begin{align}
\min_{\mathbf{w}}\ \ \frac{1}{2}\mathbf{w^Tw} + C \sum_{i=1}^n \log \left(1 + e^{-y_i\mathbf{w^Tx_i}}\right).
\end{align}
Here $C>0$ is an important penalty parameter. Since our purpose is to demonstrate the effectiveness of our proposed scheme using $b$-bit hashing, we simply provide results for a wide range of $C$ values and  assume that the best performance is achievable if we conduct cross-validations.

In our approach, we apply $k$ independent random permutations on each feature vector $\mathbf{x}_i$ and store the lowest $b$ bits of each hashed value. This way, we obtain a new dataset which can be stored using merely $nbk$ bits. At run-time,  we expand each new data point into a $2^b\times k$-length vector.\\

For example, suppose $k=3$ and the  hashed values are  originally $\{12013, 25964, 20191\}$, whose binary digits are $\{010111011101101, \ 110010101101100, 100111011011111\}$. Consider $b=2$. Then the binary digits are stored as $\{01, 00, 11\}$ (which corresponds to $\{1, 0, 3\}$ in decimals). At run-time, we need to expand them into a vector of length $2^bk = 12$, to be $\{ 0, 0, 1, 0, \ \ 0, 0, 0, 1, \ \ 1, 0, 0, 0\}$,
which will be the new feature vector fed to a solver:
\begin{align}\notag
\begin{array}{lrrr}
\text{Original hashed values } (k=3): &12013 &25964 &20191\\
\text{Original binary representations}: &010111011101101& 110010101101100& 100111011011111\\
\text{Lowest $b=2$ binary digits}: &01& 00& 11\\
\text{Expanded $2^b=4$ binary digits }: &0 0 1 0 & 0 0 0 1 & 1 0 0 0\\
\text{New feature vector fed to a solver}: &&\{0, 0, 1, 0, 0, 0, 0, 1, 1, 0, 0, 0\}
\end{array}
\end{align}

Clearly, this expansion is directly inspired by the proof that the $b$-bit minwise hashing matrix is PD in Theorem~\ref{thm_PD}. Note that the total storage cost is still just $nbk$ bits and each new data vector (of length $2^b\times k$) has exactly $k$ 1's.

Also, note that in this procedure we actually do not  explicitly estimate the resemblance $R$ using (\ref{eqn_basic}).

\section{Experimental Results on Webspam Dataset}

Our experimental settings follow the work in~\cite{Proc:Yu_KDD10} very closely. The authors of~\cite{Proc:Yu_KDD10} conducted experiments on three datasets, of which the {\em webspam} dataset is public and reasonably high-dimensional ($n=350000$, $D=16609143$). Therefore, our experiments focus on  {\em webspam}. Following~\cite{Proc:Yu_KDD10}, we randomly selected $20\%$ of samples for testing and used the remaining $80\%$ samples for training.

We chose LIBLINEAR as the tool to demonstrate the effectiveness of our algorithm.  All experiments were conducted on  workstations with Xeon(R) CPU (W5590@3.33GHz) and 48GB RAM, under Windows 7 System. Thus, in our case, the original data (about 24GB in LIBSVM format) fit in memory. In applications for which the data do not fit in memory, we expect that $b$-bit hashing will be even more substantially advantageous, because the hashed data are relatively very small. In fact, our experimental results will show that for this dataset, using $k=200$ and $b=8$ can achieve the same testing accuracy as using the original data. The effective storage for the reduced dataset (with 350K examples, using $k=200$ and $b=8$) would be merely about 70MB.

\subsection{ Experimental Results on Nonlinear (Kernel) SVM}

We implemented a new resemblance kernel function   and tried to use LIBSVM to train the {\em webspam} dataset. We waited for \textbf{over one week}\footnote{We will let the program run unless it is accidentally terminated (e.g., due to power outage).} but LIBSVM still had not output any results. Fortunately, using $b$-bit minswise hashing to estimate the resemblance kernels, we were able to obtain some results. For example, with $C=1$ and $b=8$, the training time of LIBSVM ranged from 1938 seconds ($k=30$) to 13253 seconds ($k=500$). In particular, when $k\geq200$, the  test accuracies essentially matched the best test results given by LIBLINEAR on the original {\em webspam} data.

Therefore, there is a significant benefit of {\em data reduction} provided by $b$-bit minwise hashing, for training nonlinear SVM. This experiment also demonstrates that it is very important (and fortunate) that we are able to transform this nonlinear problem into a linear problem.

\subsection{ Experimental Results on Linear SVM}

Since there is an important tuning parameter $C$ in linear SVM and logistic regression, we conducted our extensive experiments for a wide range of $C$ values (from $10^{-3}$ to $10^2$) with fine spacings in $[0.1,\ 10]$.

\begin{figure}[h!]

\mbox{
\includegraphics[width=1.7in]{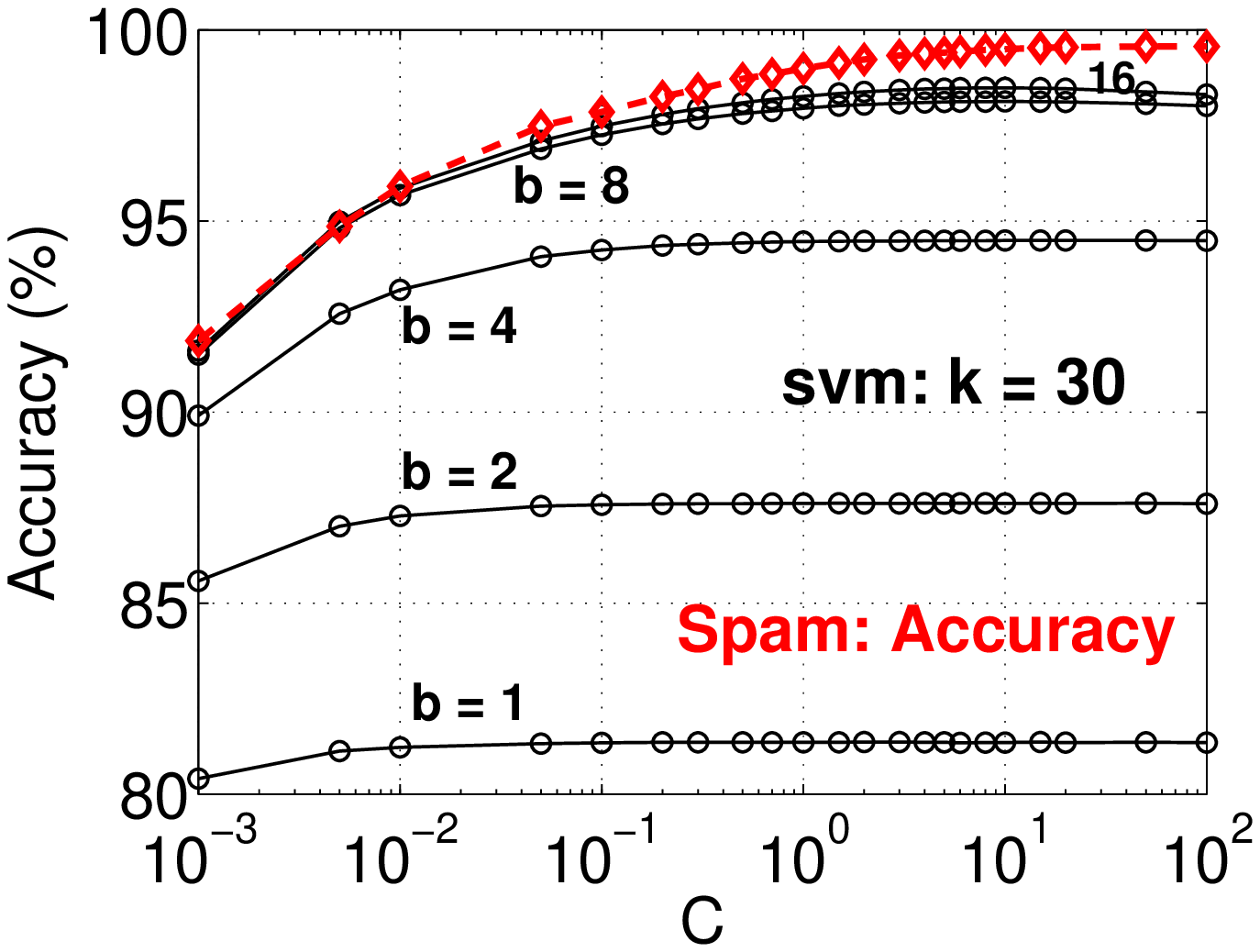}\hspace{-0.1in}
\includegraphics[width=1.7in]{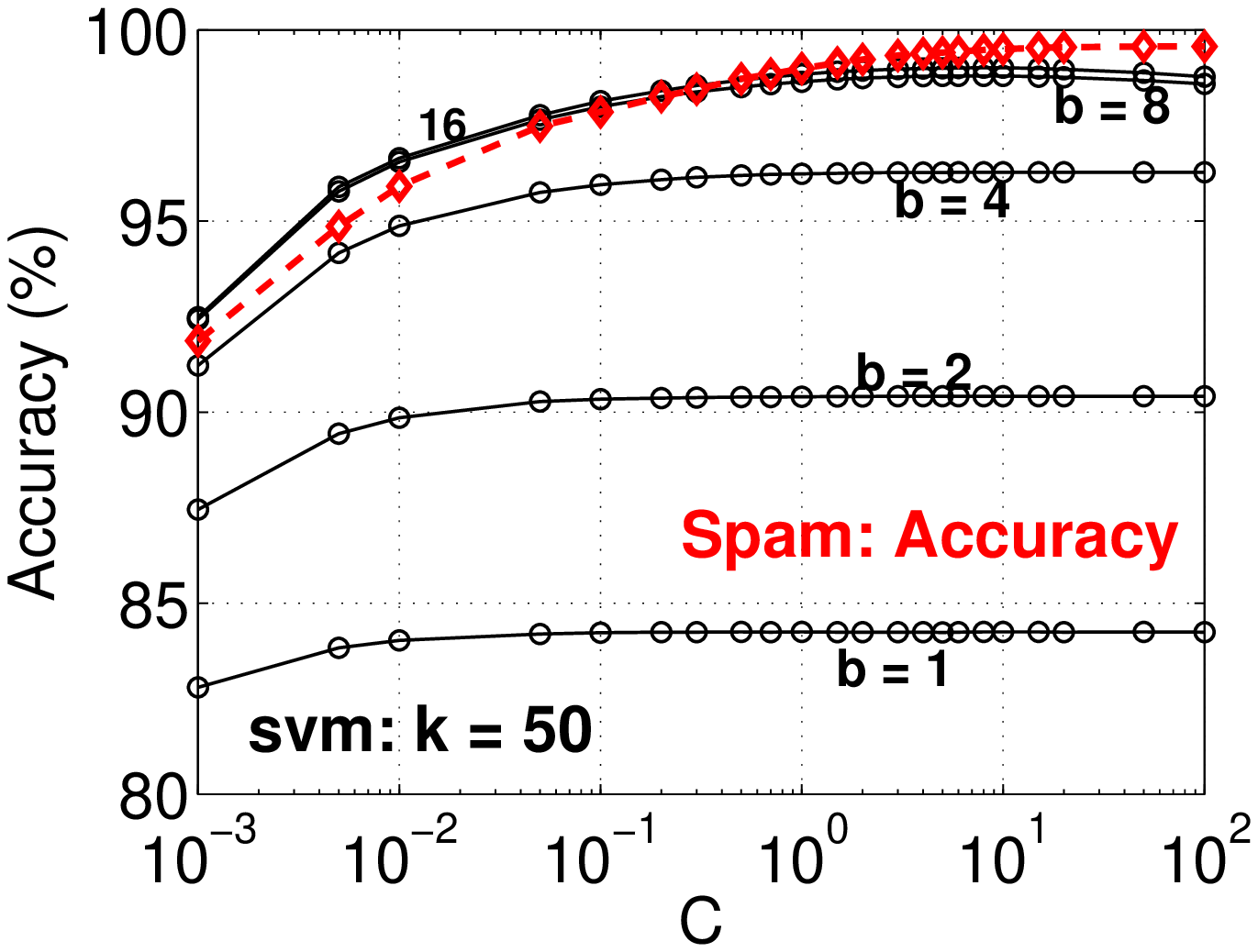}\hspace{-0.1in}
\includegraphics[width=1.7in]{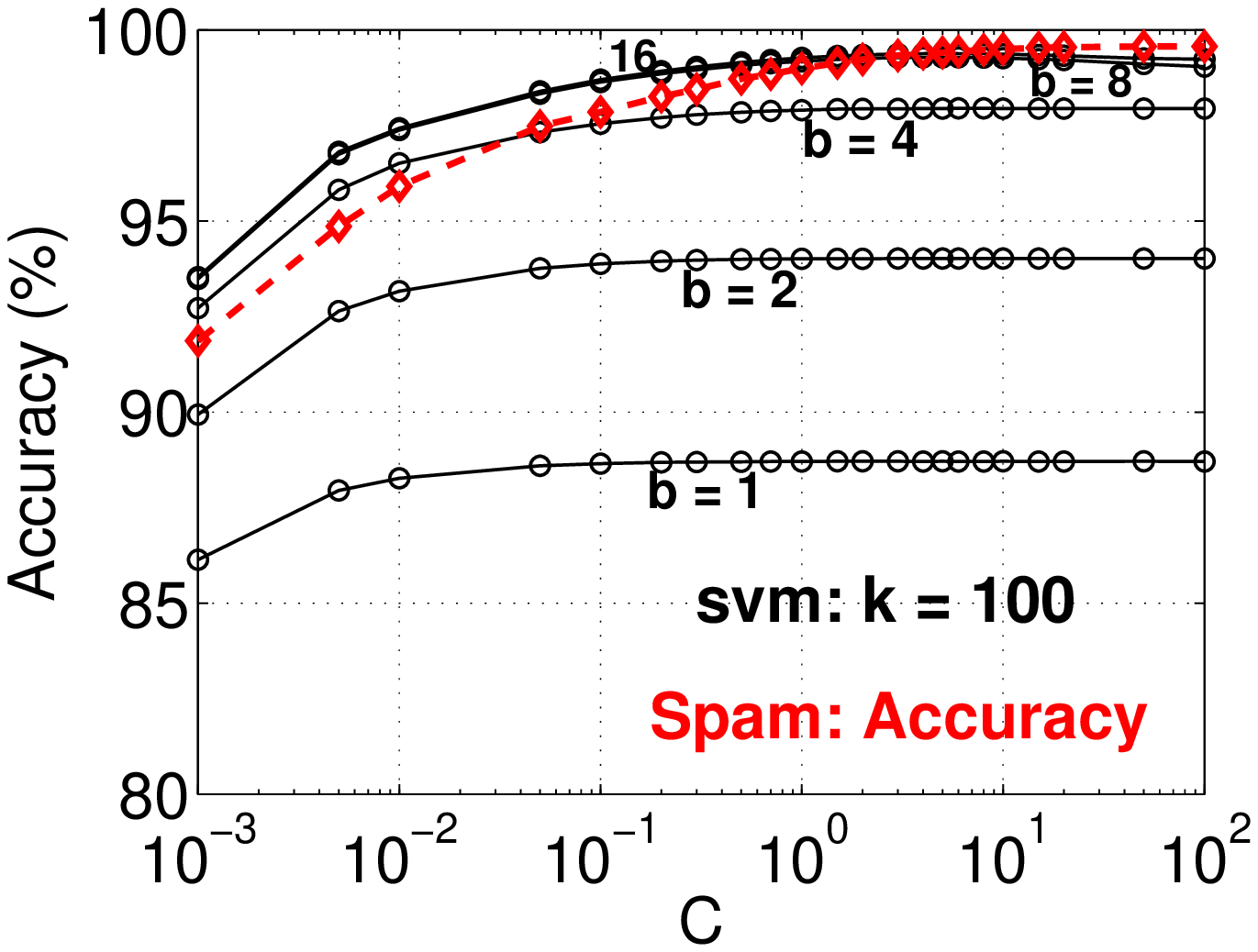}\hspace{-0.1in}
\includegraphics[width=1.7in]{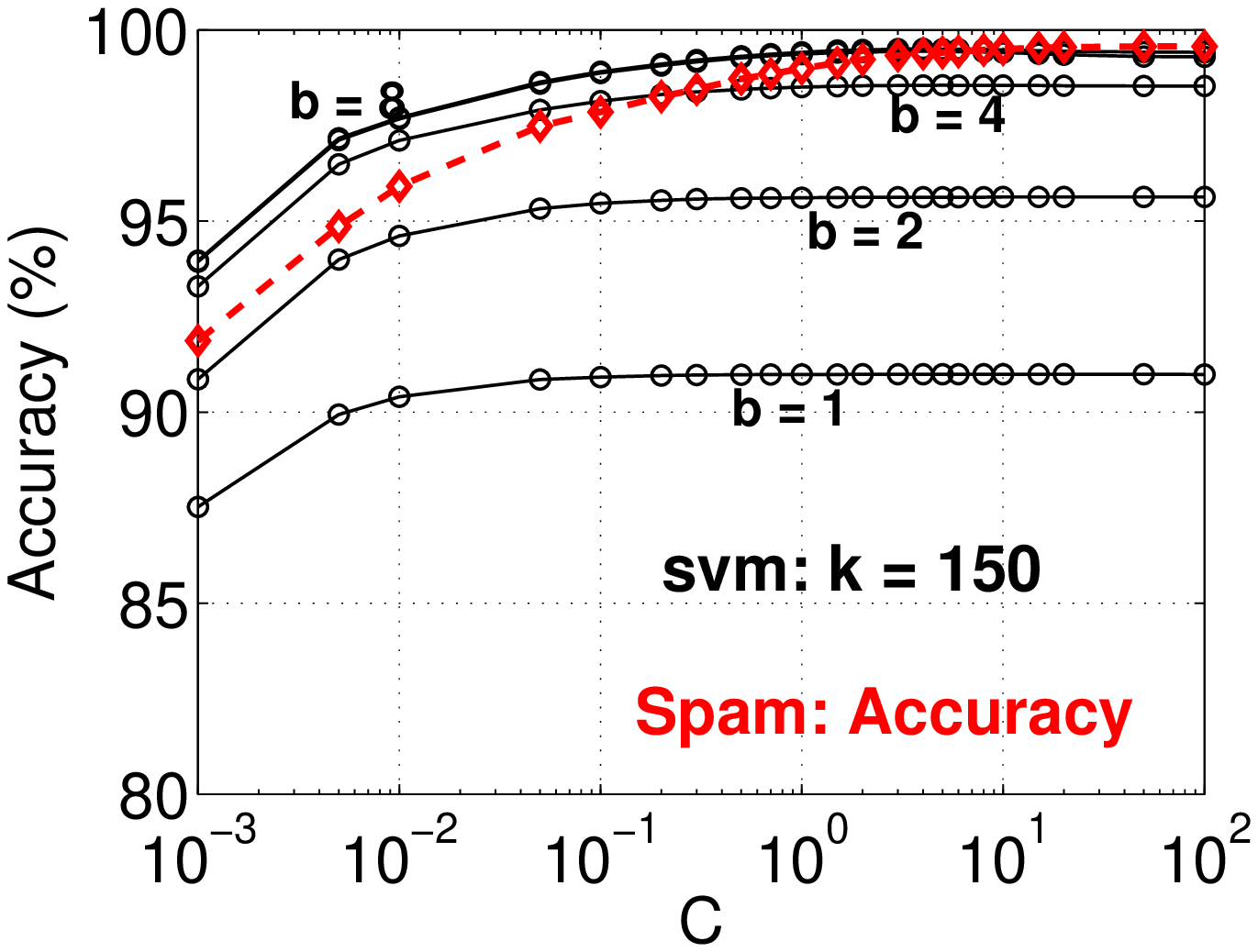}}

\mbox{
\includegraphics[width=1.7in]{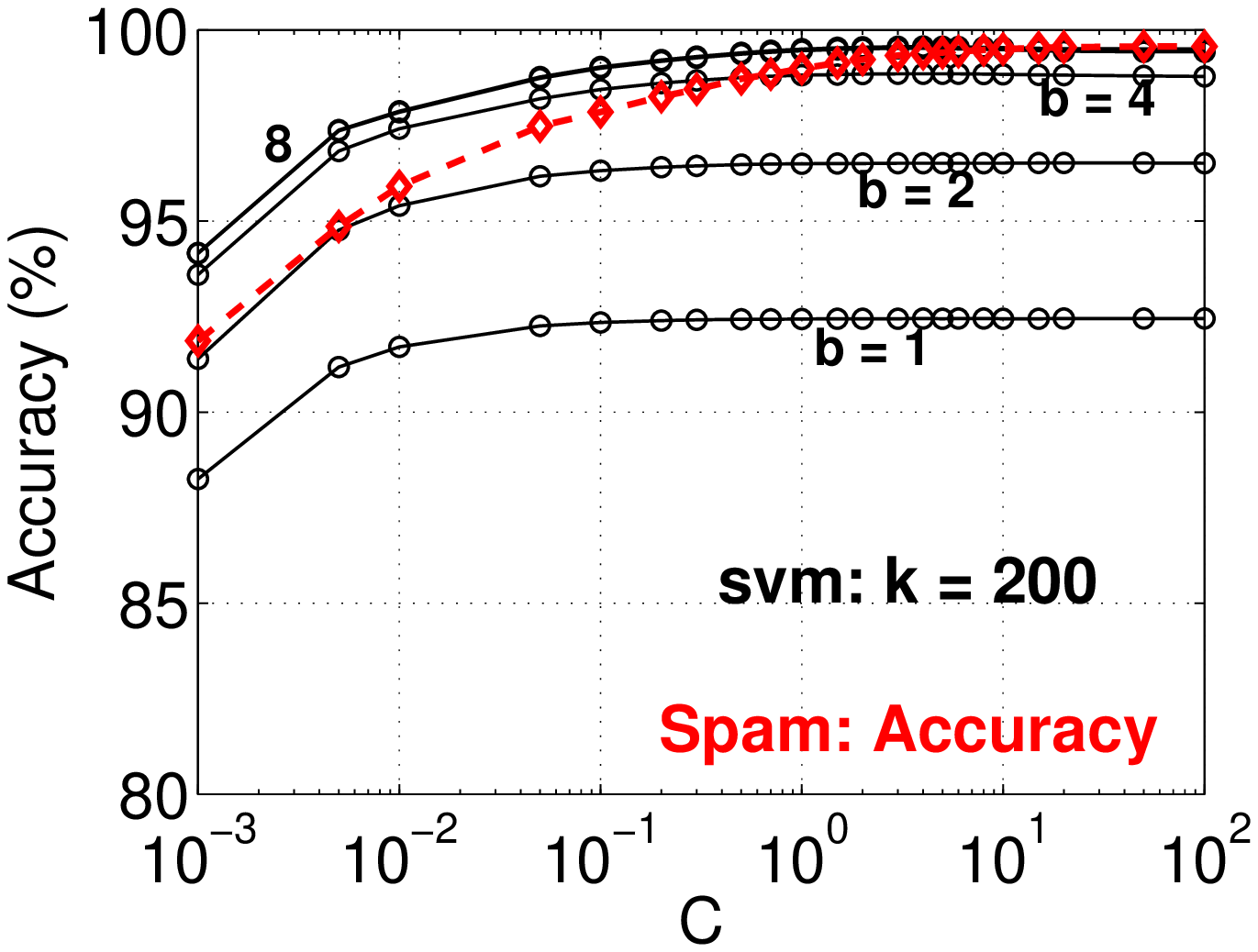}\hspace{-0.1in}
\includegraphics[width=1.7in]{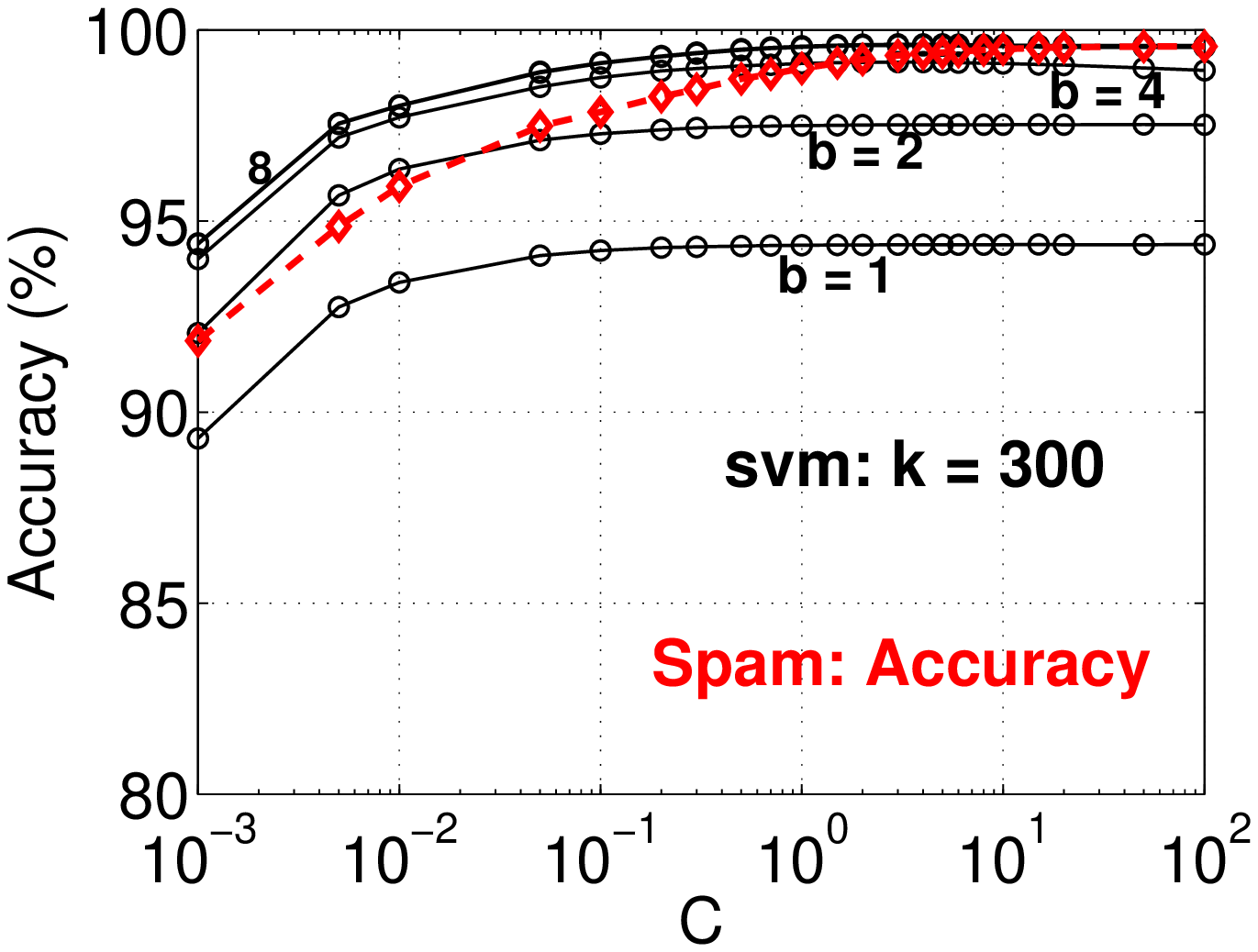}\hspace{-0.1in}
\includegraphics[width=1.7in]{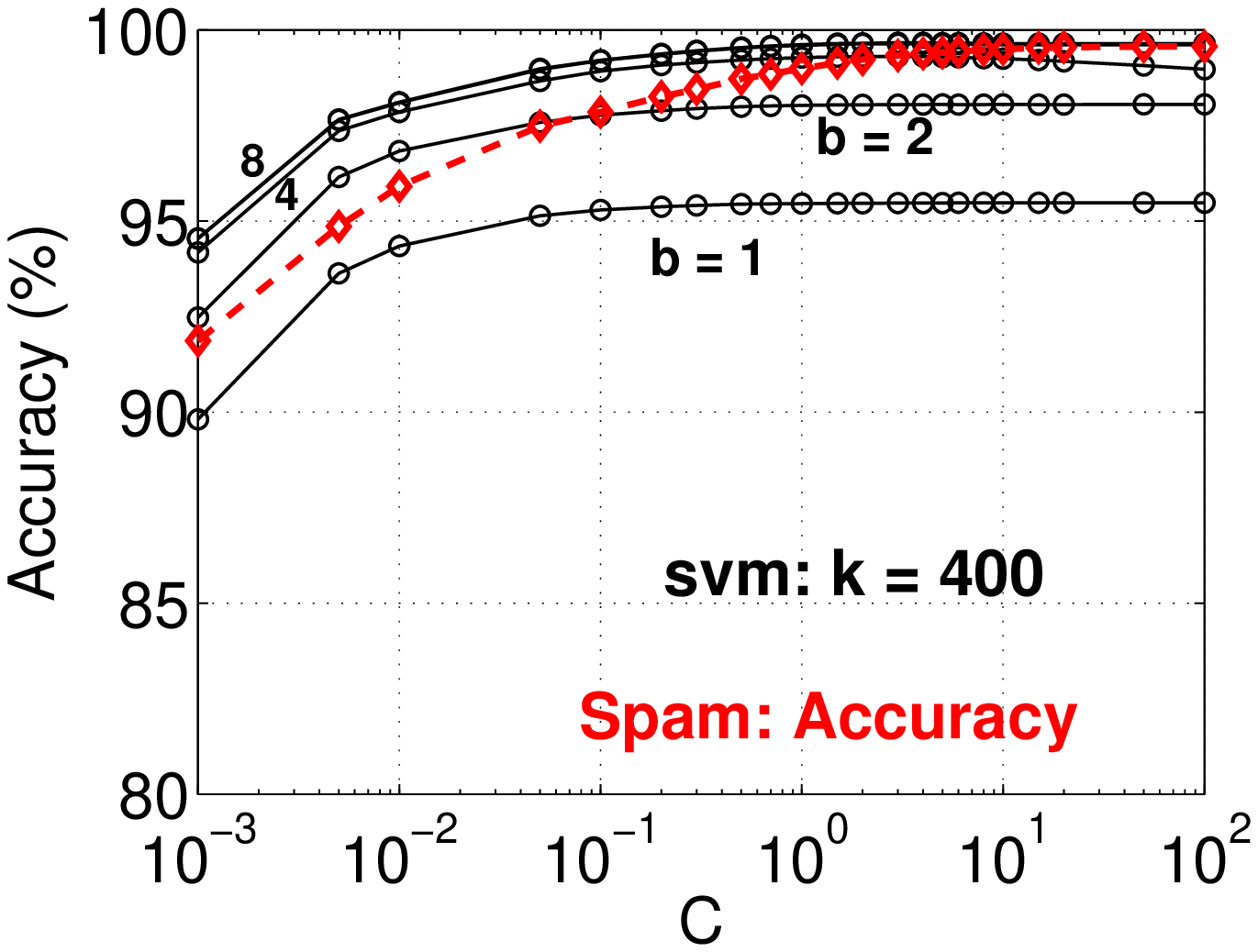}\hspace{-0.1in}
\includegraphics[width=1.7in]{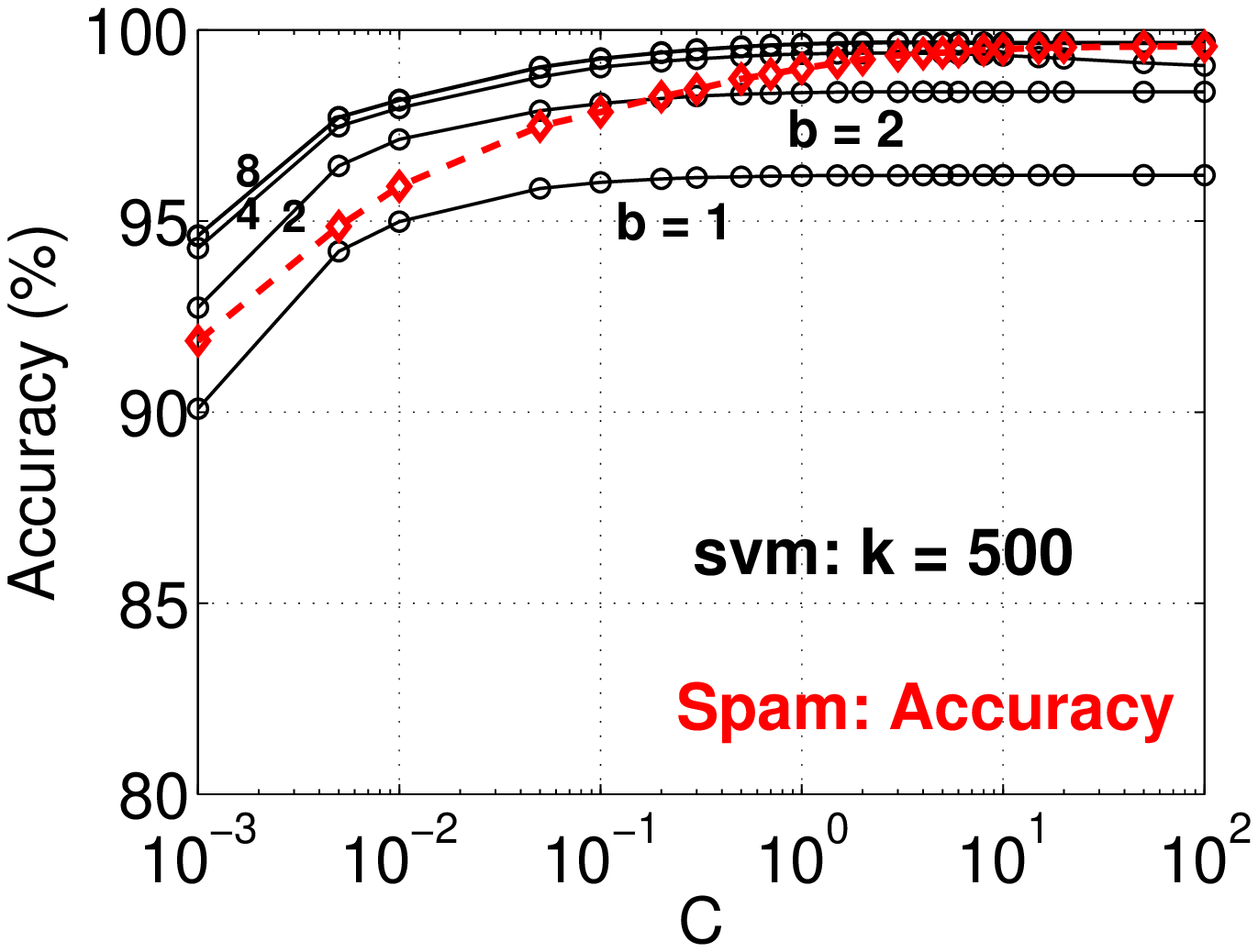}}

\vspace{-0.15in}

\caption{\textbf{Linear SVM test accuracy} (averaged over 50 repetitions). With $k\geq 100$ and $b\geq 8$. $b$-bit hashing (solid) achieves very similar accuracies as using the original data (dashed, red if color is available). Note that after $k\geq 150$, the curves for $b=16$ overlap the curves for $b=8$. }\label{fig_acc}
\end{figure}

\begin{figure}[h!]

\mbox{
\includegraphics[width=1.7in]{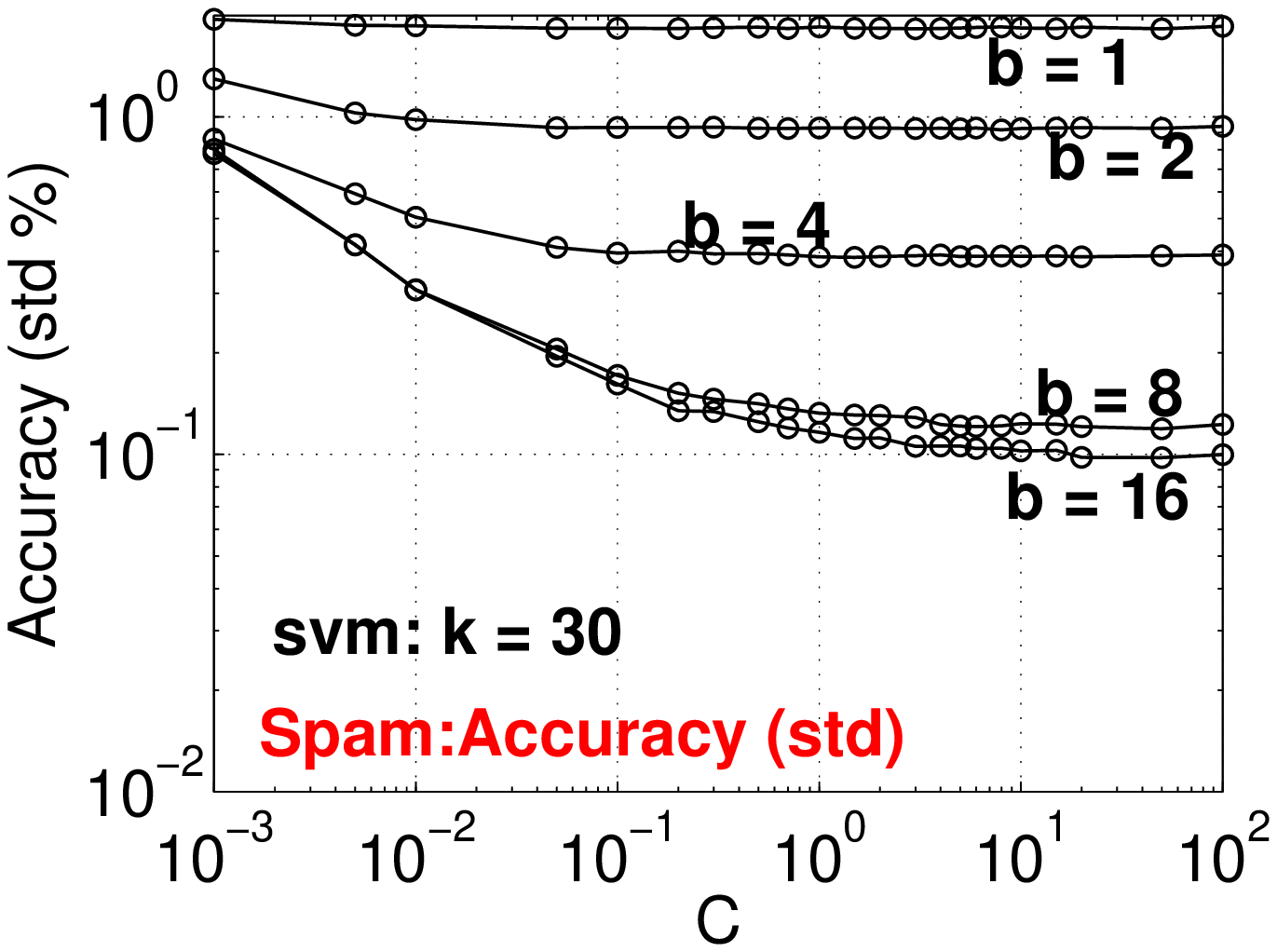}\hspace{-0.1in}
\includegraphics[width=1.7in]{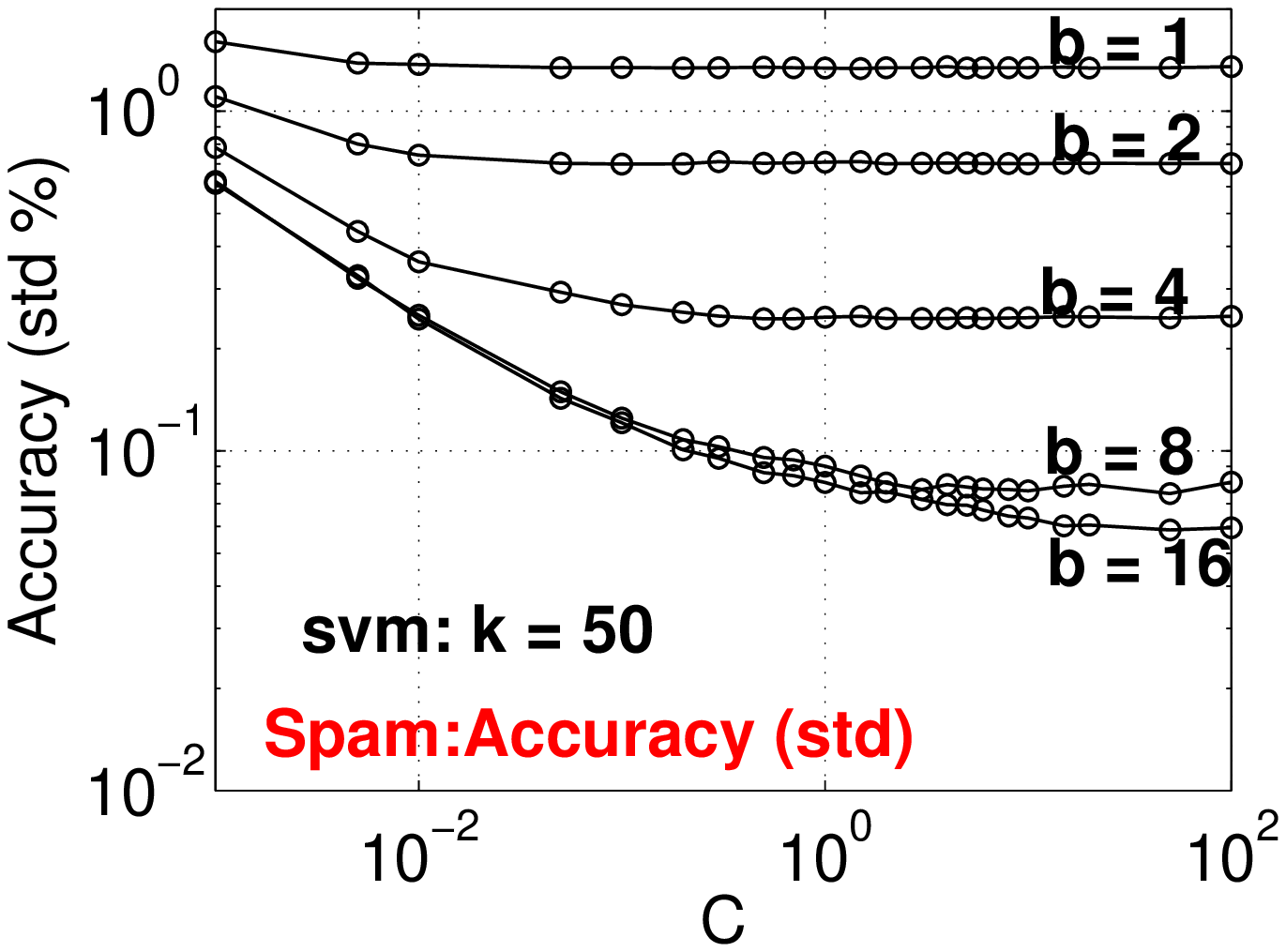}\hspace{-0.1in}
\includegraphics[width=1.7in]{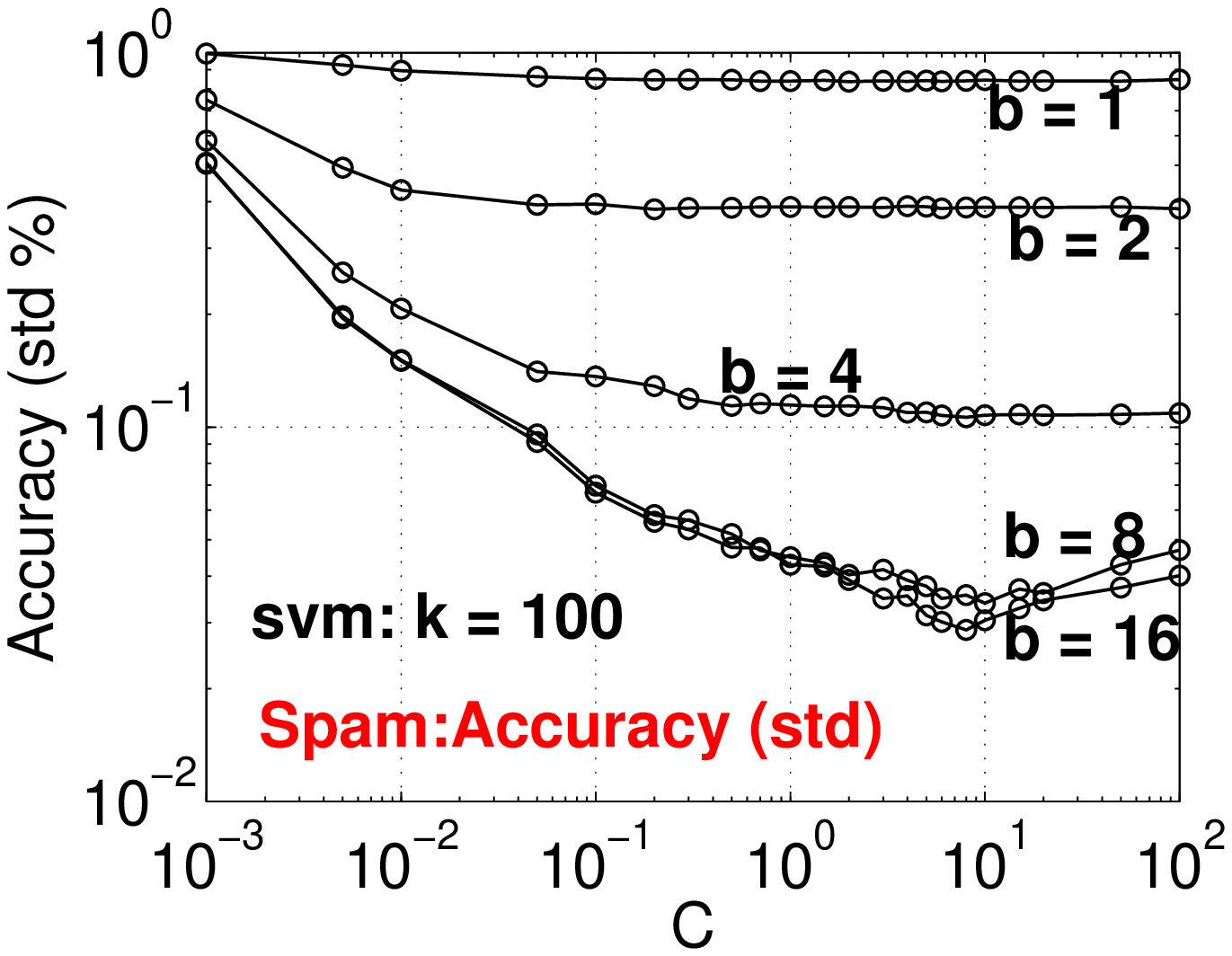}\hspace{-0.1in}
\includegraphics[width=1.7in]{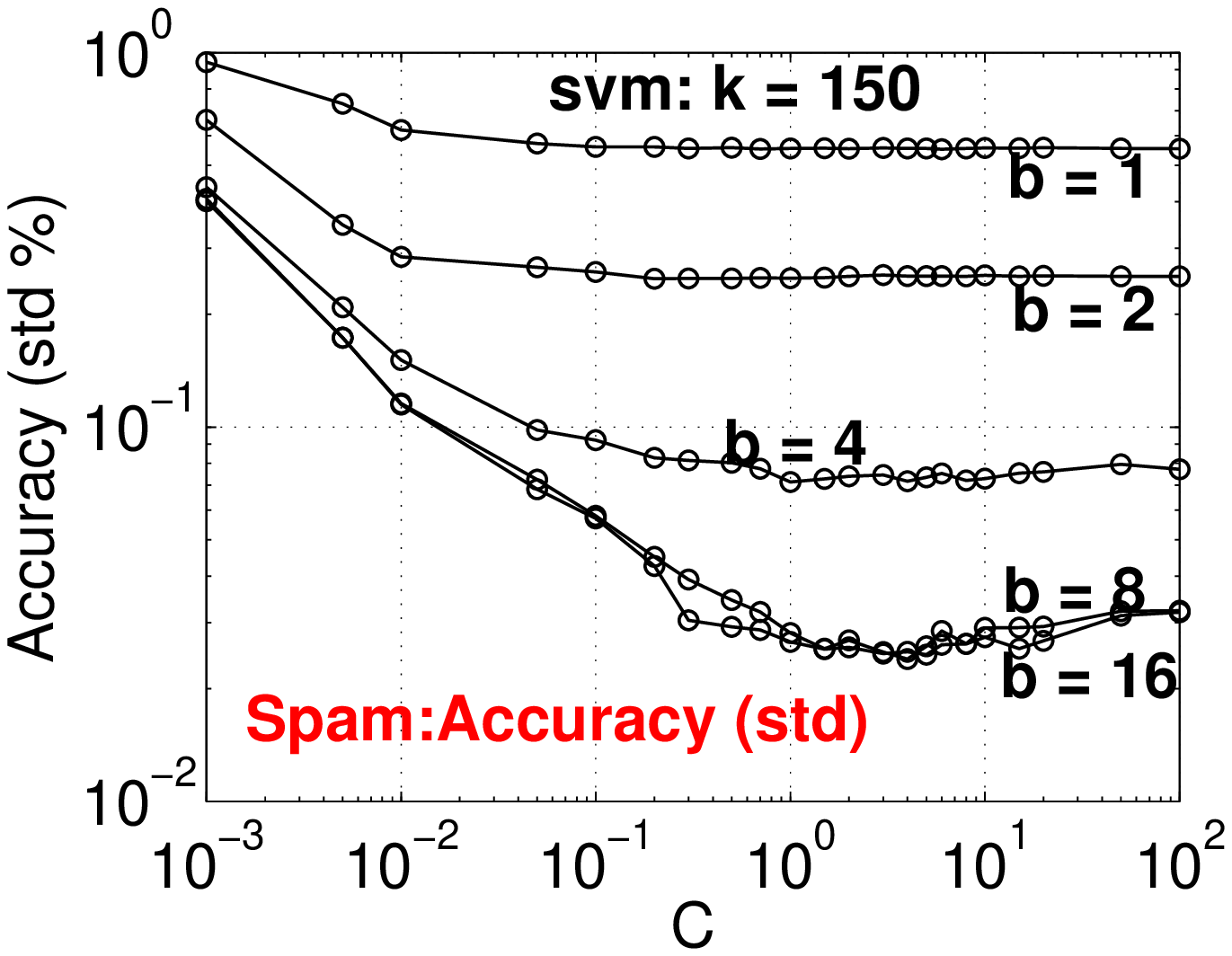}}

\mbox{
\includegraphics[width=1.7in]{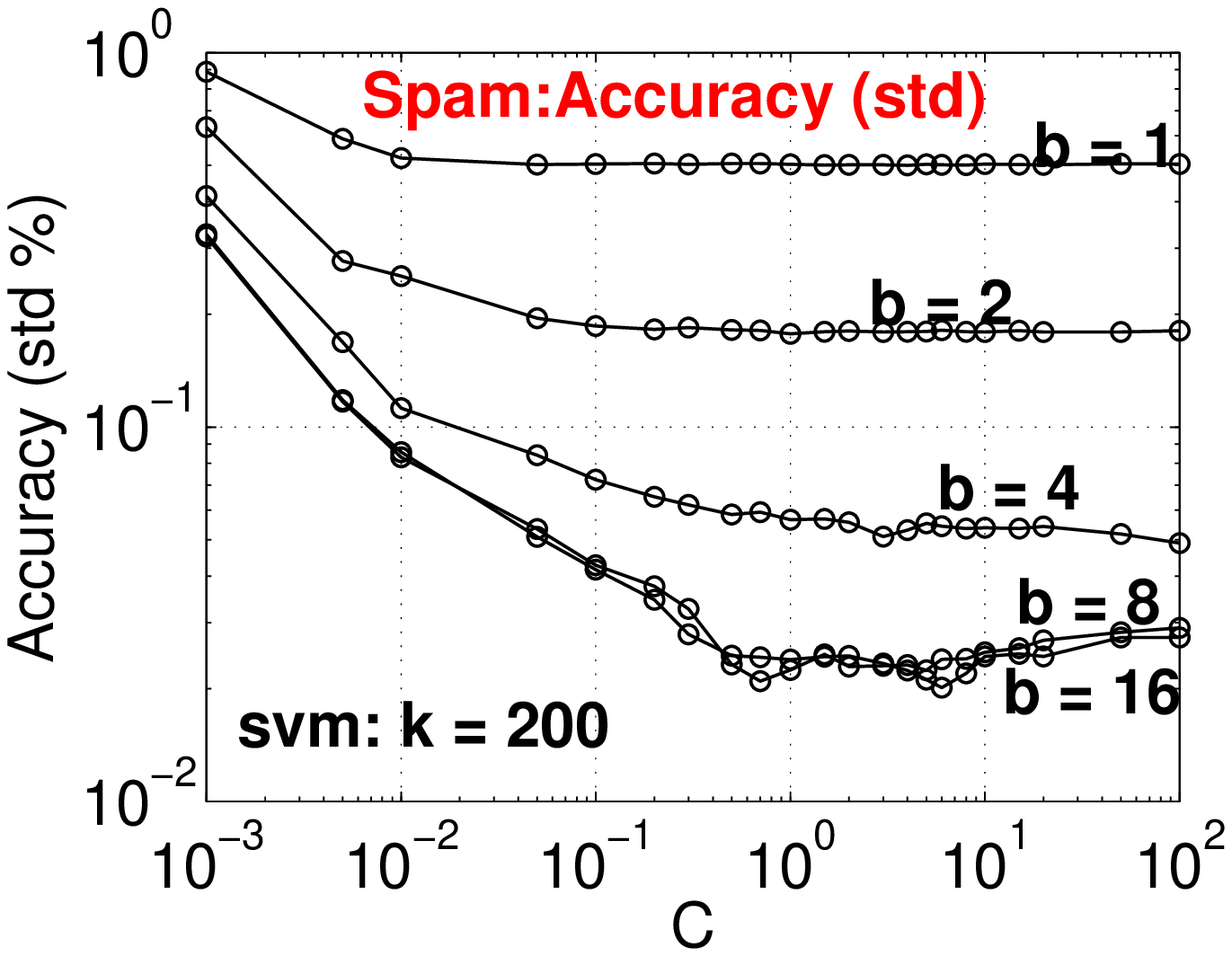}\hspace{-0.1in}
\includegraphics[width=1.7in]{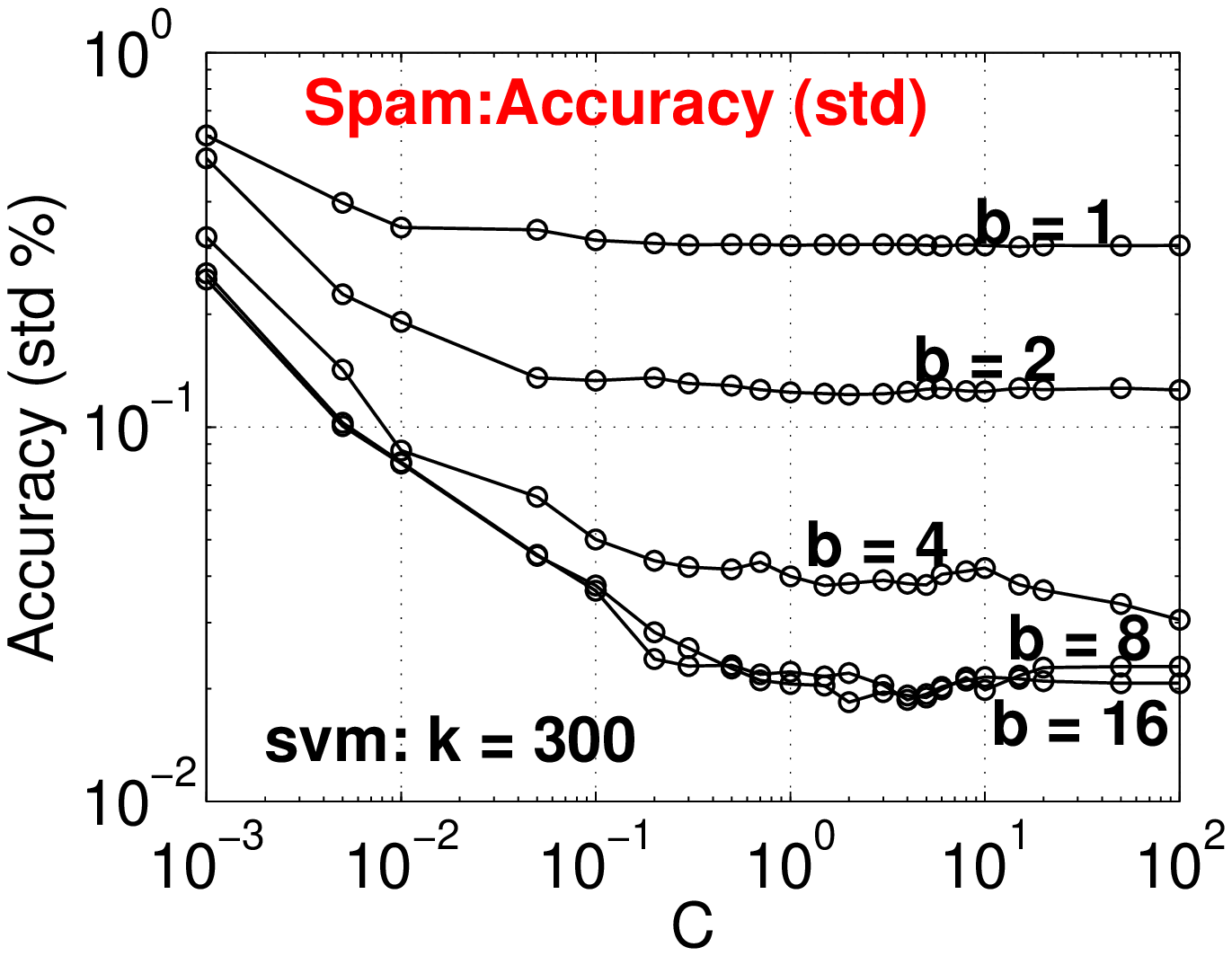}\hspace{-0.1in}
\includegraphics[width=1.7in]{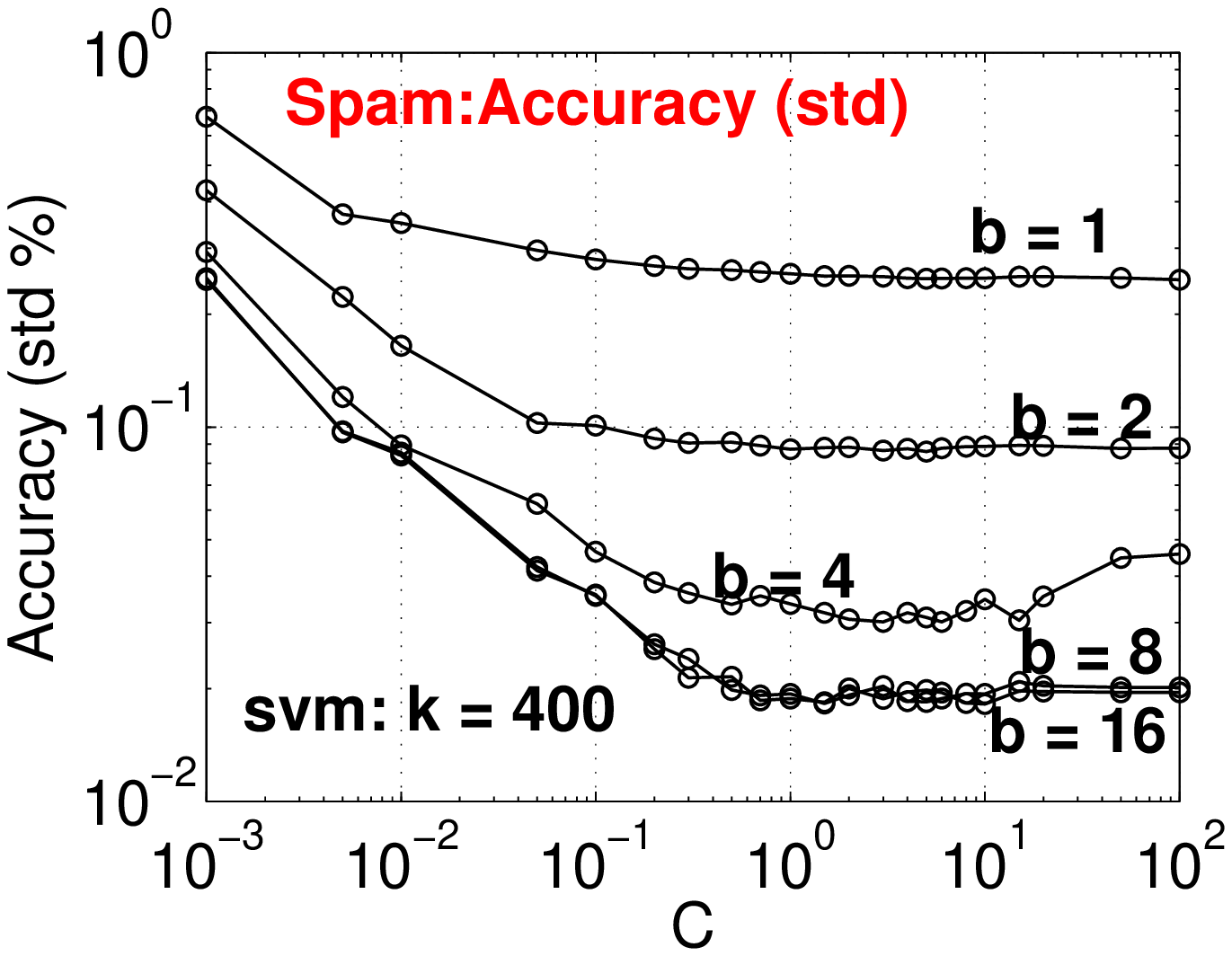}\hspace{-0.1in}
\includegraphics[width=1.7in]{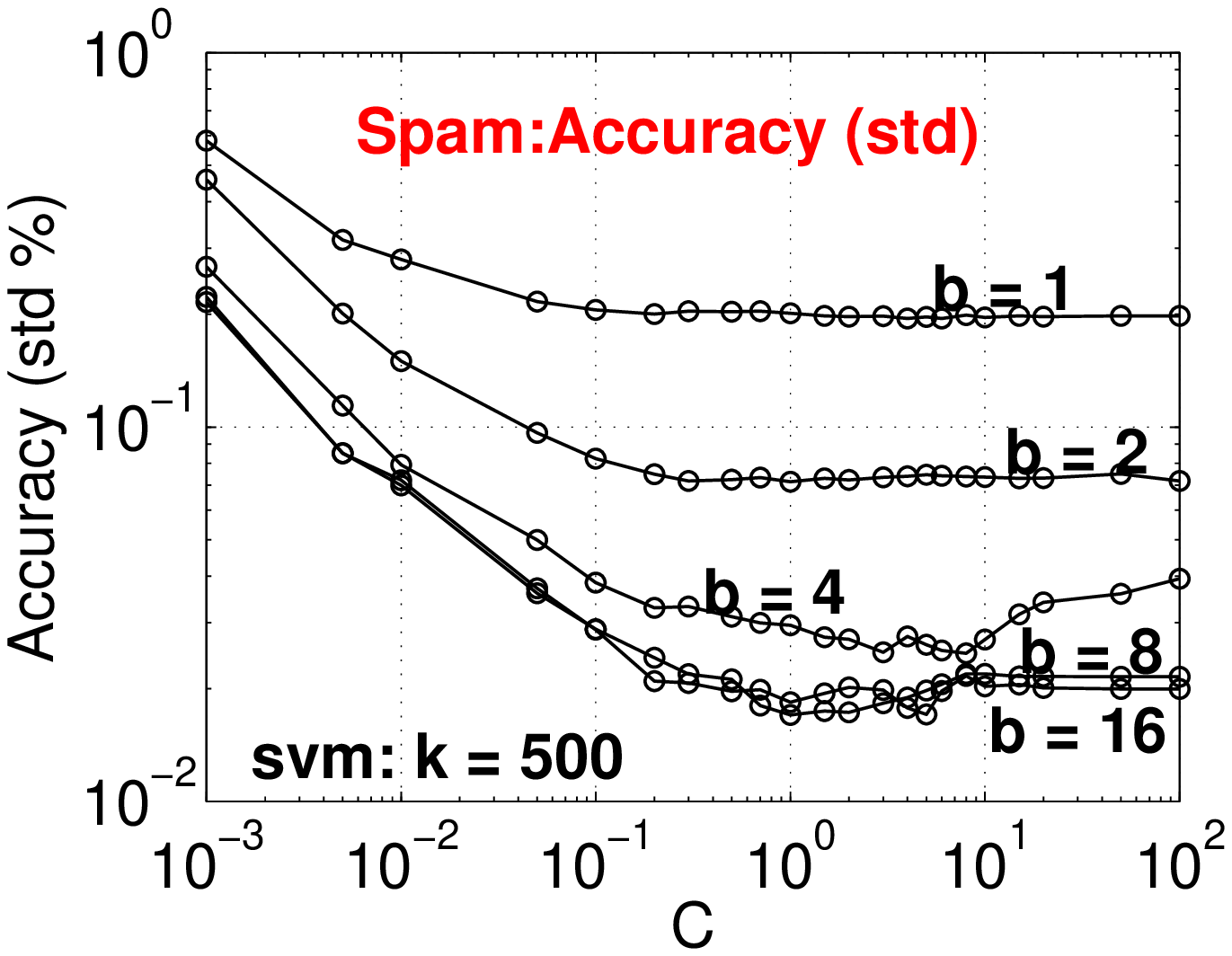}}

\vspace{-0.15in}

\caption{\textbf{Linear SVM test accuracy (std)}. The standard deviations are computed from 50 repetitions. When $b\geq8$, the standard deviations become extremely small (e.g., $0.02\%$).  }\label{fig_acc_std}
\end{figure}

We mainly experimented with $k=30$ to $k=500$, and $b=1$, 2, 4, 8, and 16. Figures~\ref{fig_acc} (average) and~\ref{fig_acc_std} (std, standard deviation) provide the test accuracies.  Figure~\ref{fig_acc} demonstrates that using $b\geq 8$ and $k\geq 150$ achieves about the same test accuracies as using the original data. Since our method is randomized, we repeated every experiment 50 times. We report both the mean and std values. Figure~\ref{fig_acc_std} illustrates that the stds are very small, especially with $b\geq4$. In other words, our algorithm produces stable predictions. For this dataset, the best performances were usually achieved when  $C\geq 1$.

\clearpage

\begin{figure}[h!]

\mbox{
\includegraphics[width=1.7in]{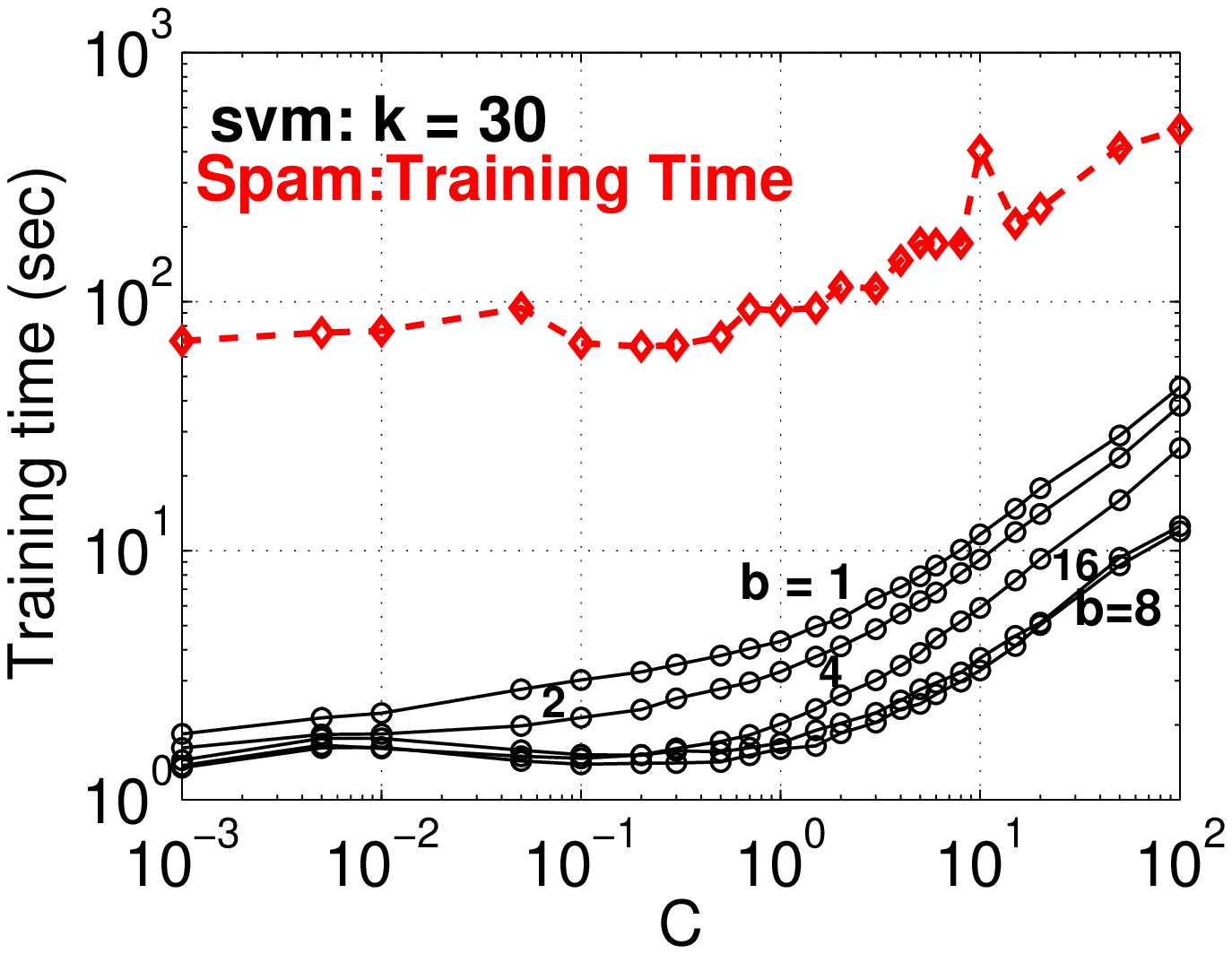}\hspace{-0.1in}
\includegraphics[width=1.7in]{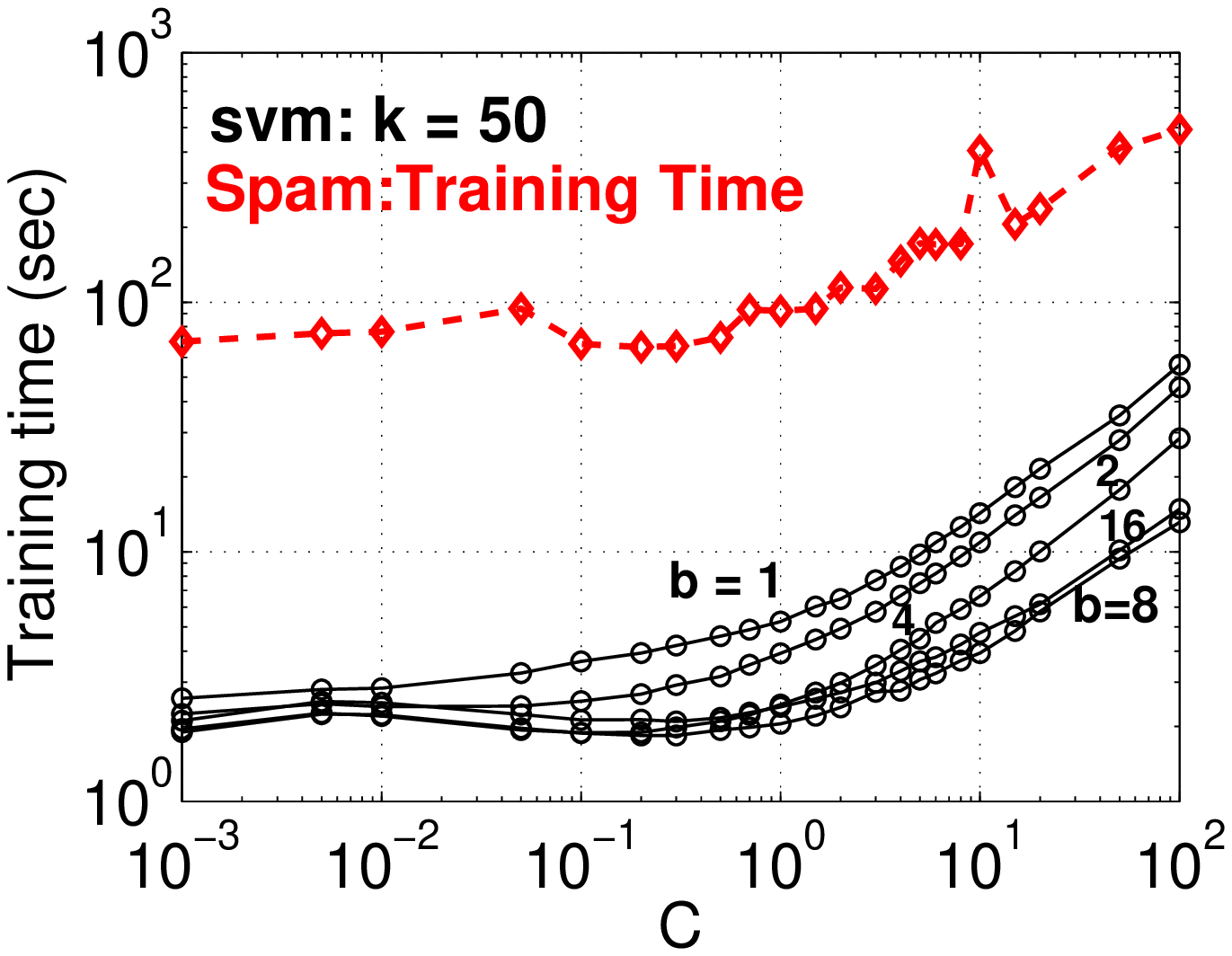}\hspace{-0.1in}
\includegraphics[width=1.7in]{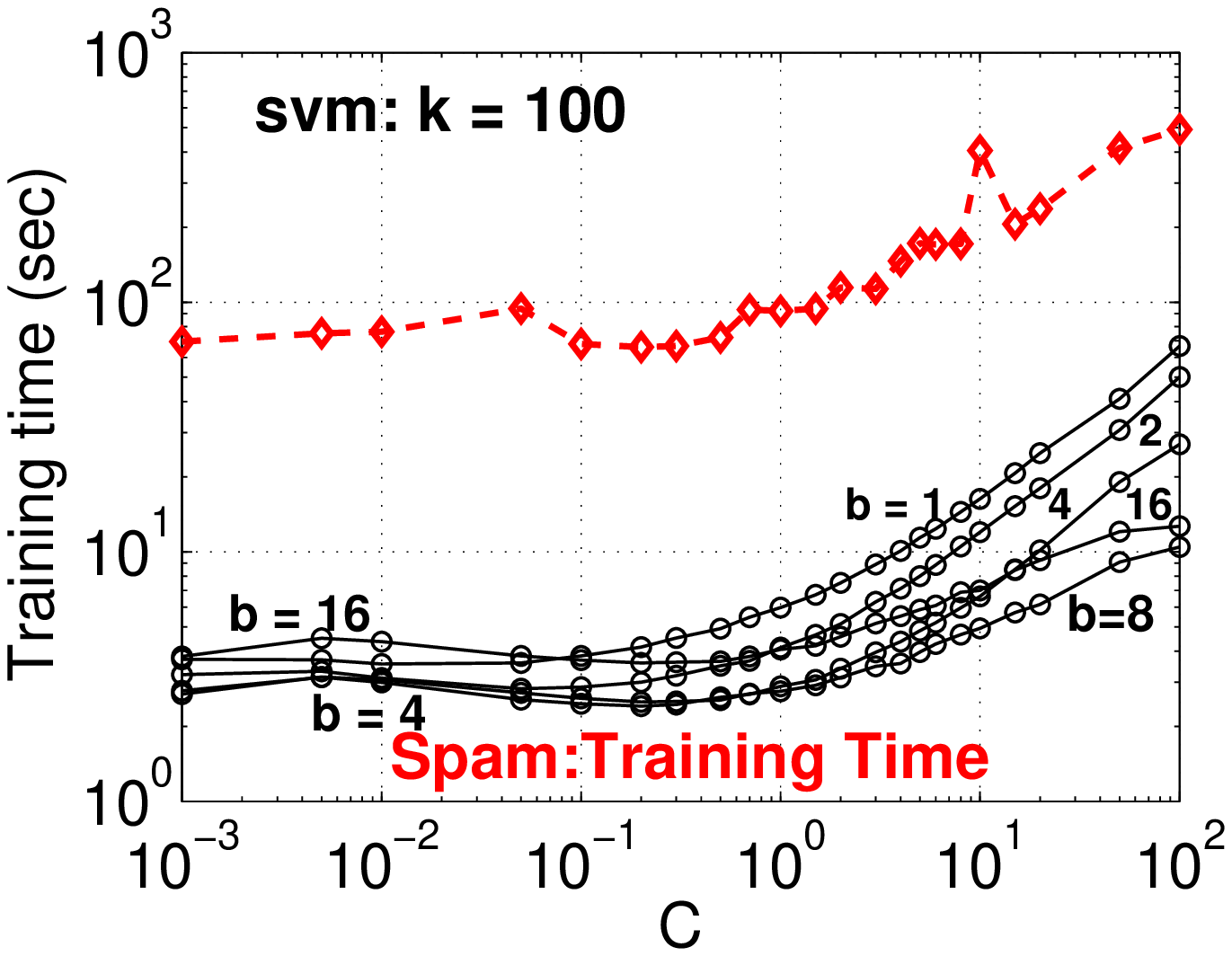}\hspace{-0.1in}
\includegraphics[width=1.7in]{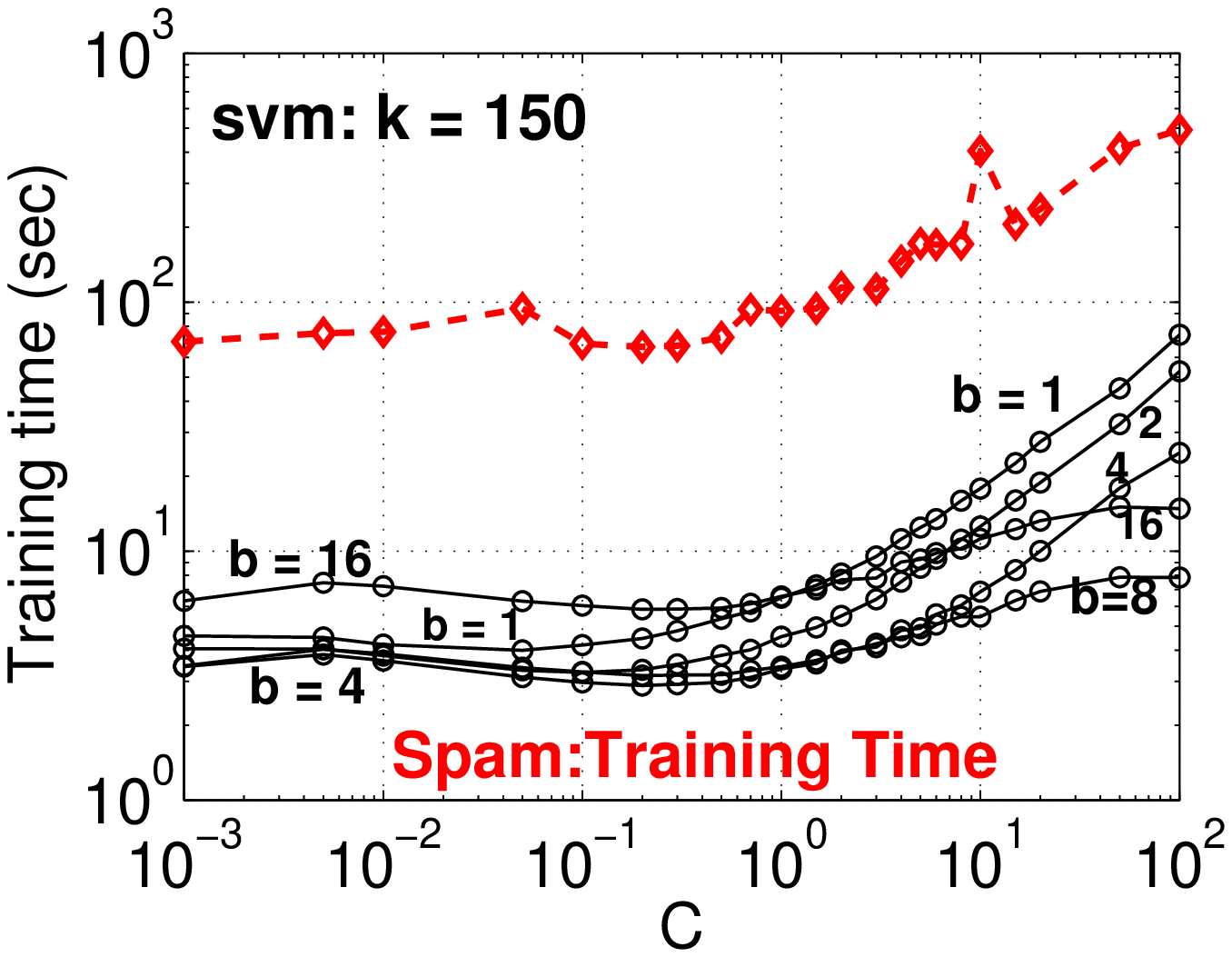}}

\mbox{
\includegraphics[width=1.7in]{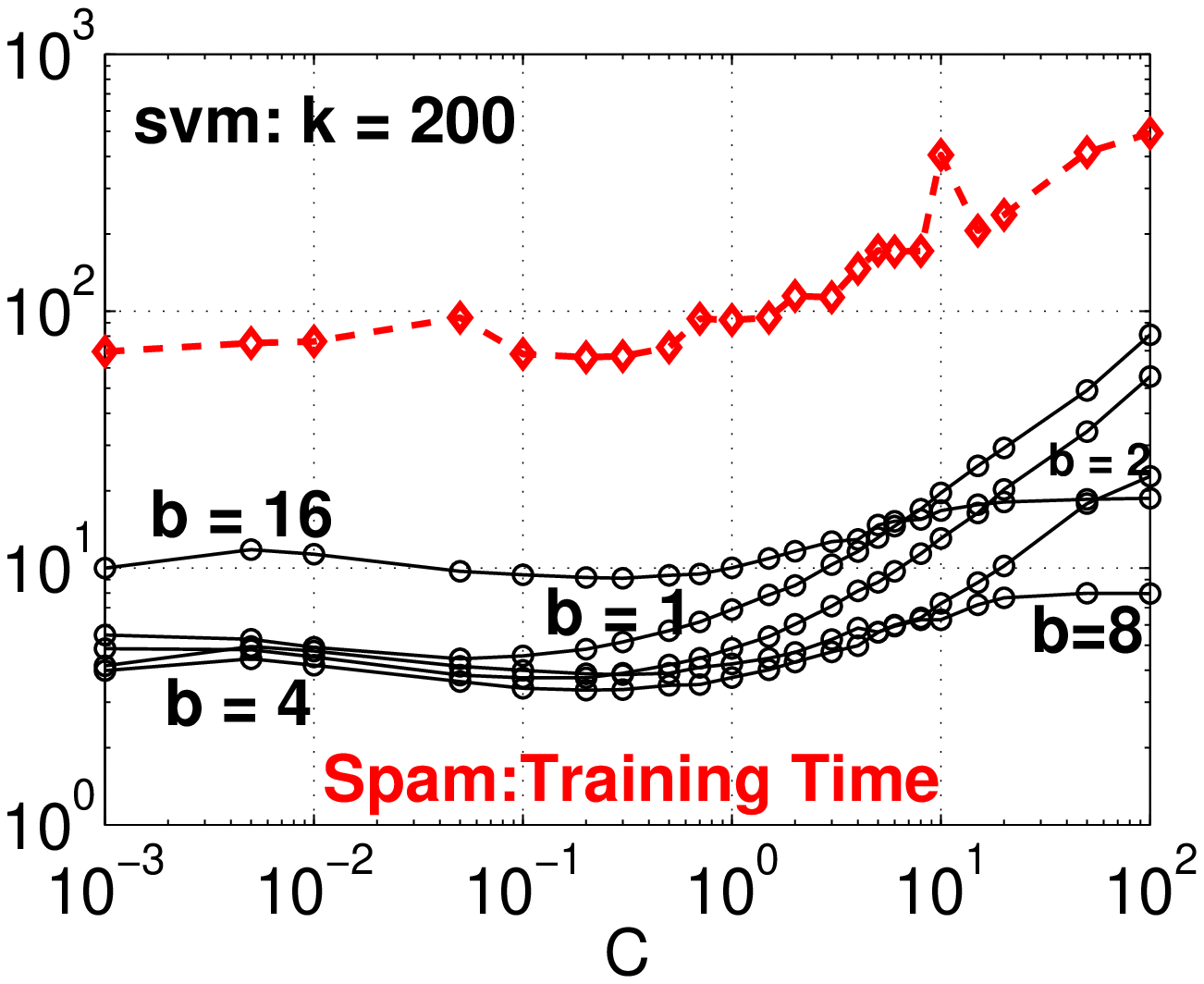}\hspace{-0.1in}
\includegraphics[width=1.7in]{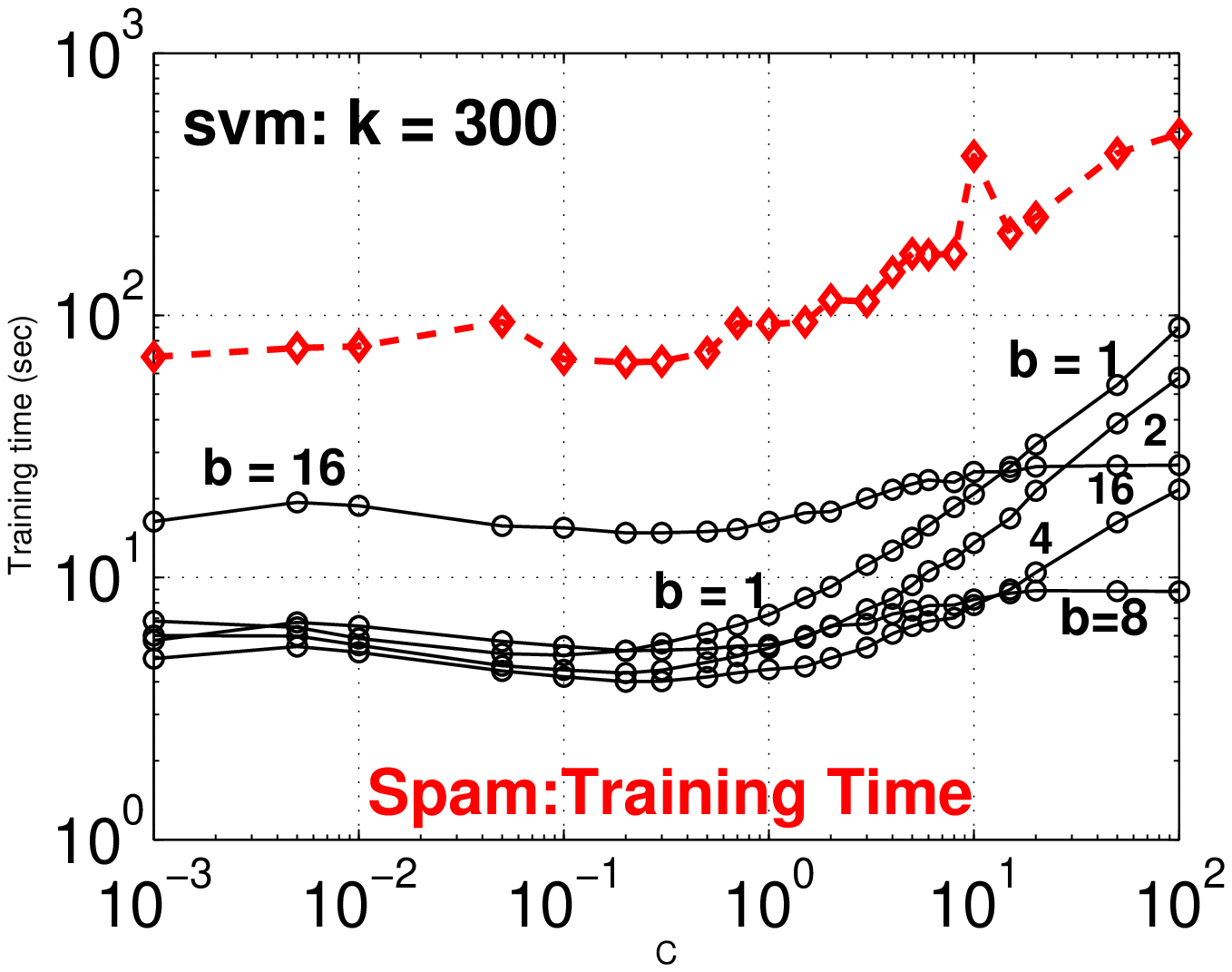}\hspace{-0.1in}
\includegraphics[width=1.7in]{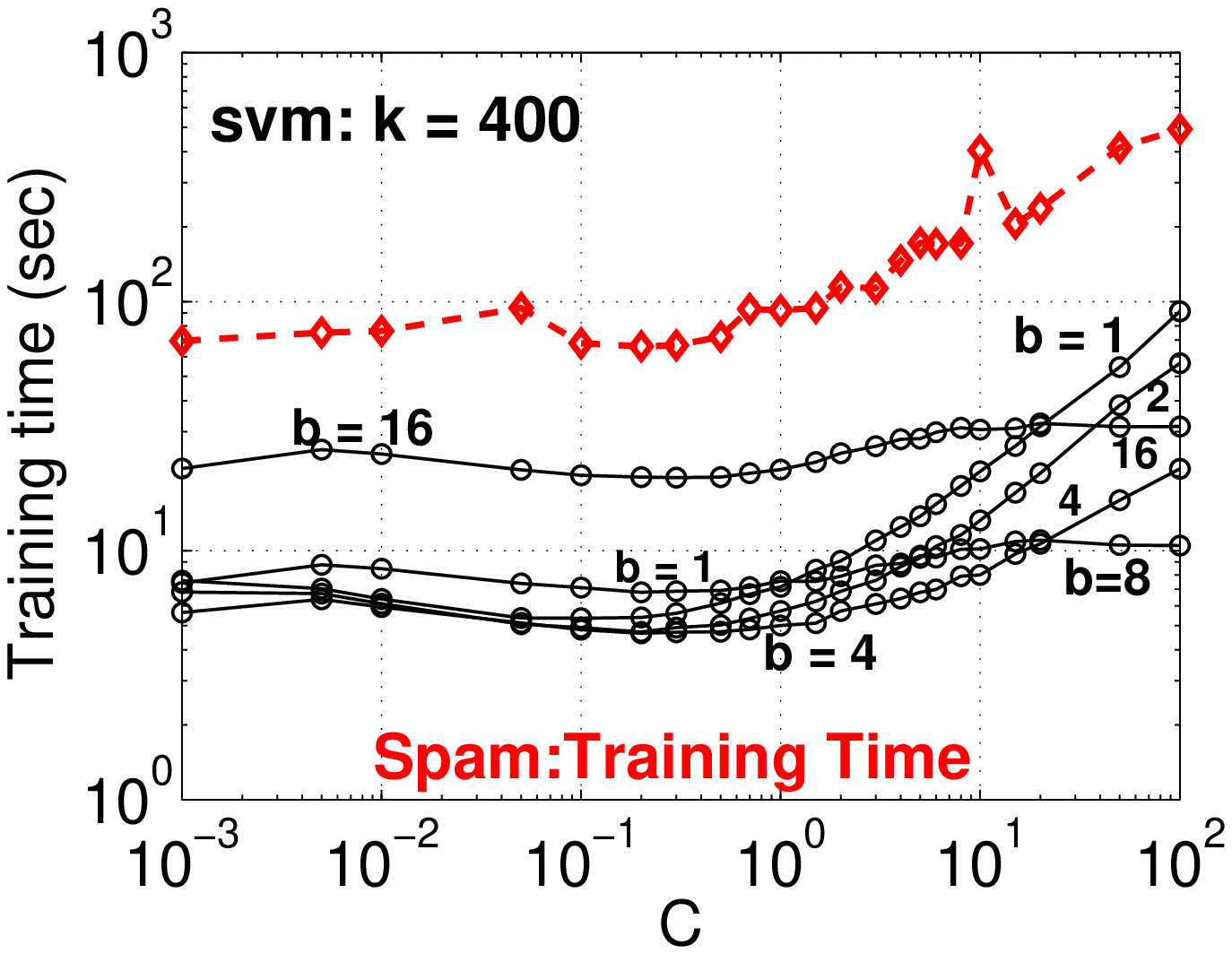}\hspace{-0.1in}
\includegraphics[width=1.7in]{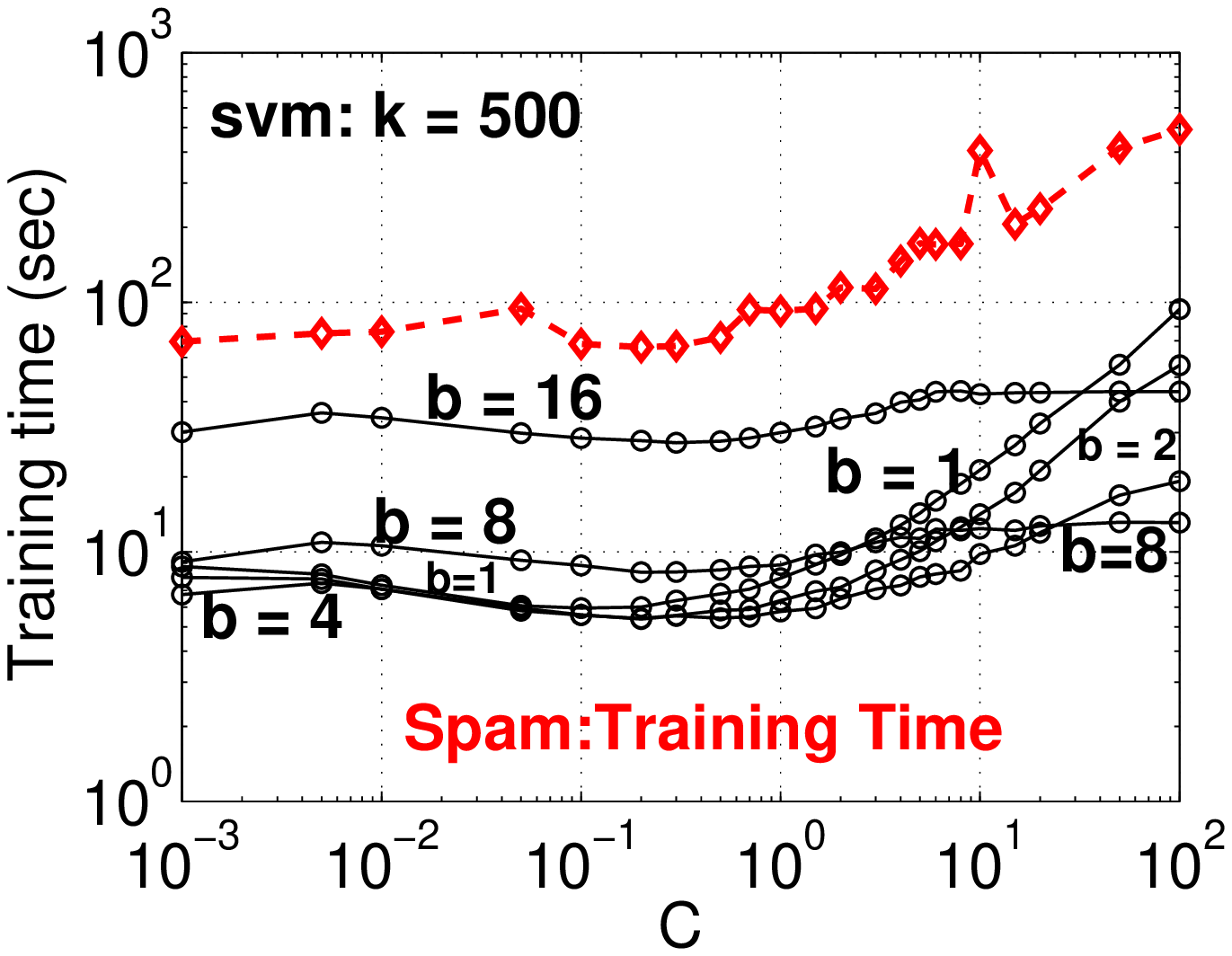}}

\vspace{-0.15in}
\caption{\textbf{Linear SVM Training time}. Compared with the training time of using the original data (dashed, red if color is available), we can see that our method with $b$-bit hashing only needs a very small fraction of the original cost. }\label{fig_training}
\end{figure}

Compared with the original training time (about 100 seconds), we can see from Figure~\ref{fig_training} that our method only need about $3\sim7$ seconds near $C=1$ (about 3 seconds for $b=8$). Note that here the training time did not include the data loading time. Loading the original data took about 12 minutes while loading the hashed data took only about 10 seconds. Of course, there is a cost for processing (hashing) the data, which we find is  efficient, confirming prior studies~\cite{Proc:Broder}. In fact, data processing can be conducted during data collection, as is the standard practice in search. In other words, prior to conducting the learning procedure, the data may be already processed and stored by ($b$-bit) minwise hashing, which can be used for multiple tasks including learning, clustering, duplicate detection, near-neighbor search, etc.\\

\begin{figure}[h!]

\mbox{
\includegraphics[width=1.7in]{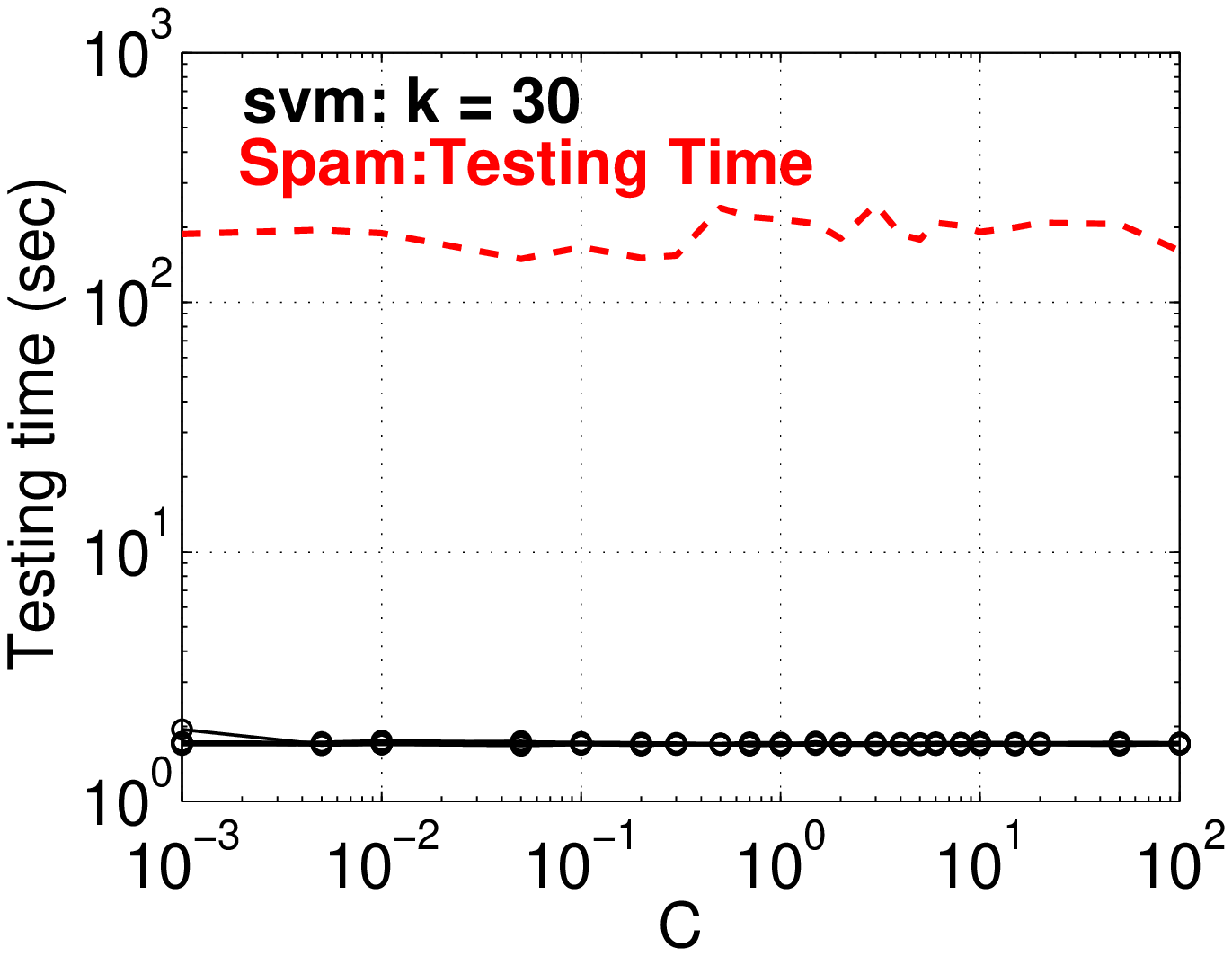}\hspace{-0.1in}
\includegraphics[width=1.7in]{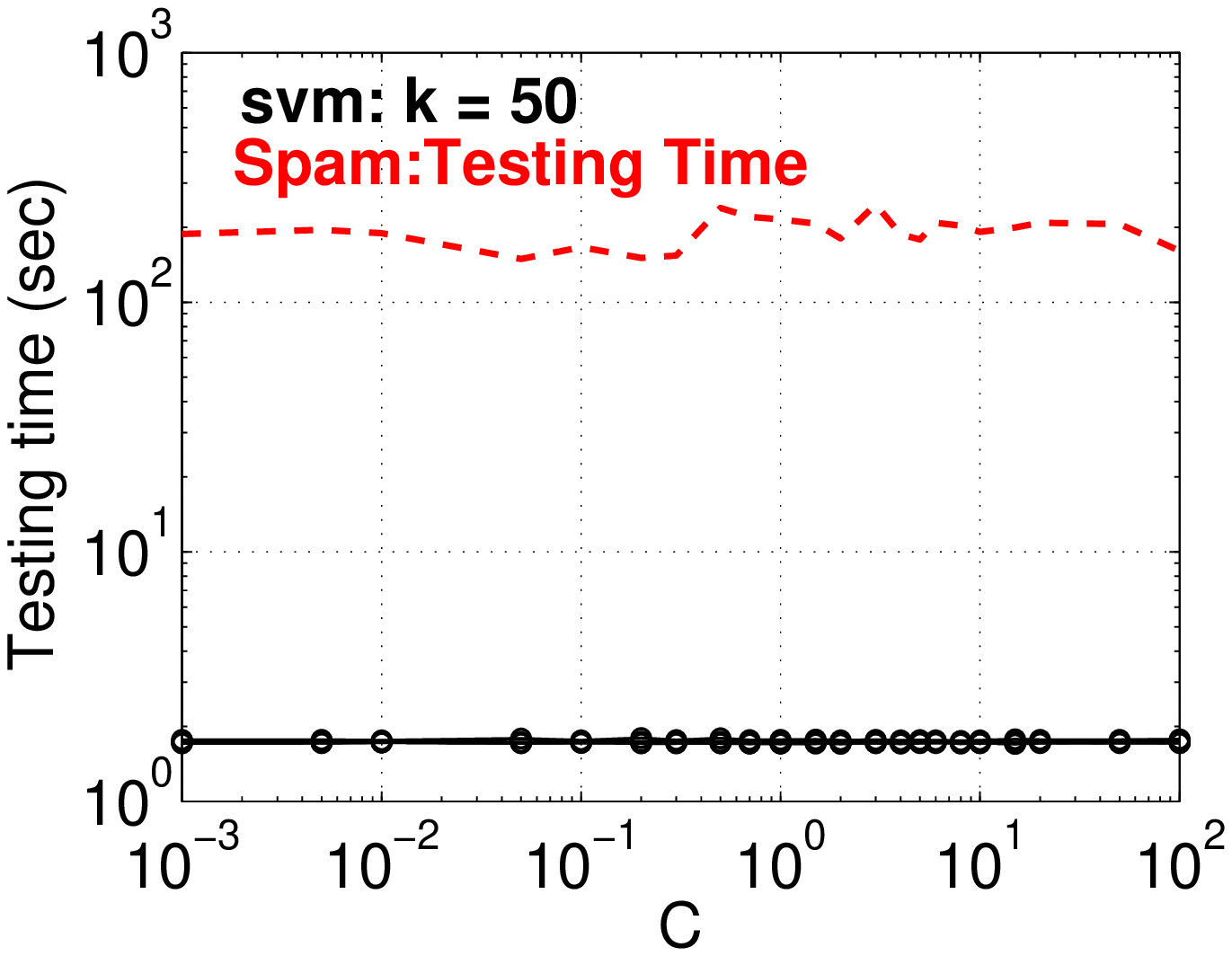}\hspace{-0.1in}
\includegraphics[width=1.7in]{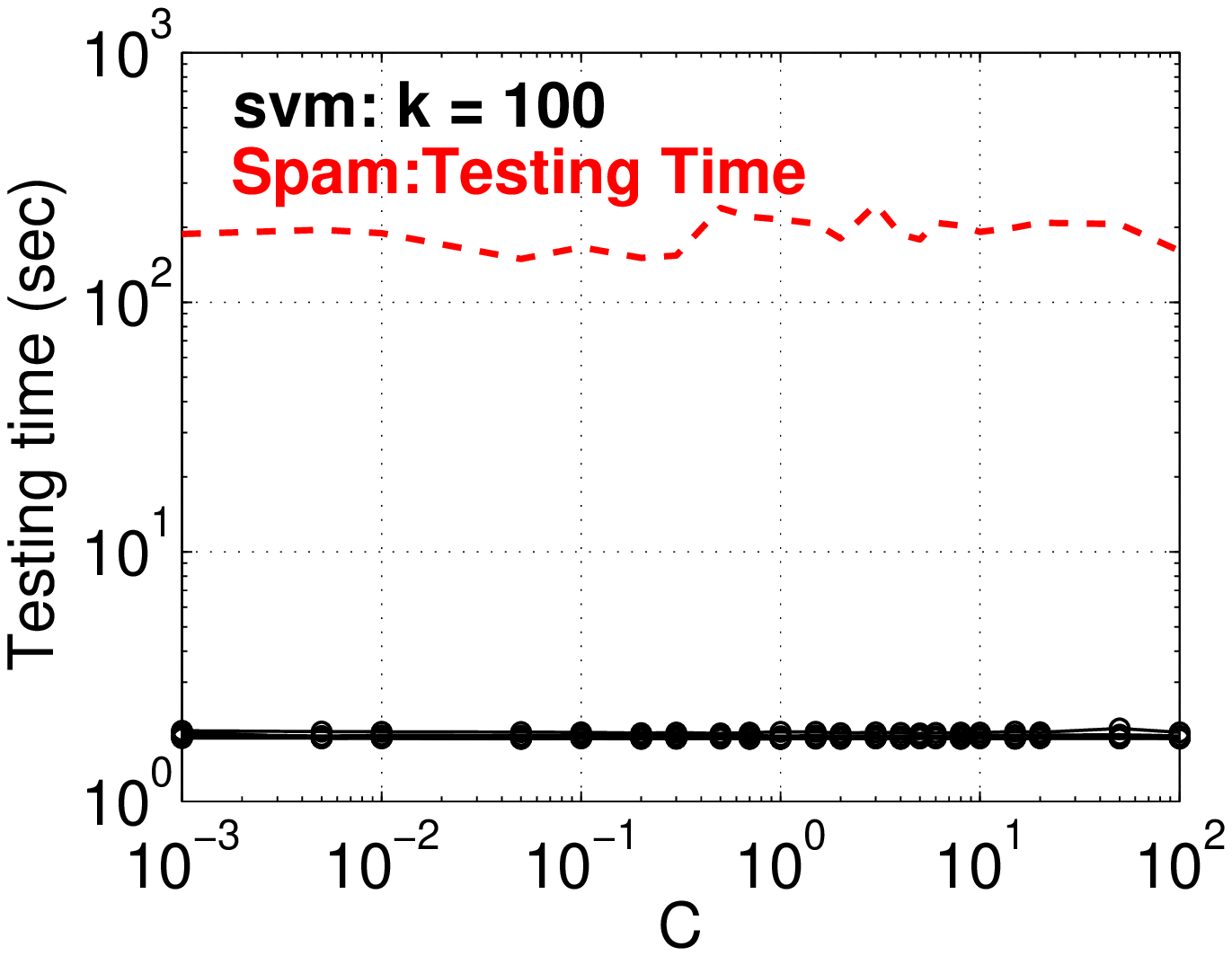}\hspace{-0.1in}
\includegraphics[width=1.7in]{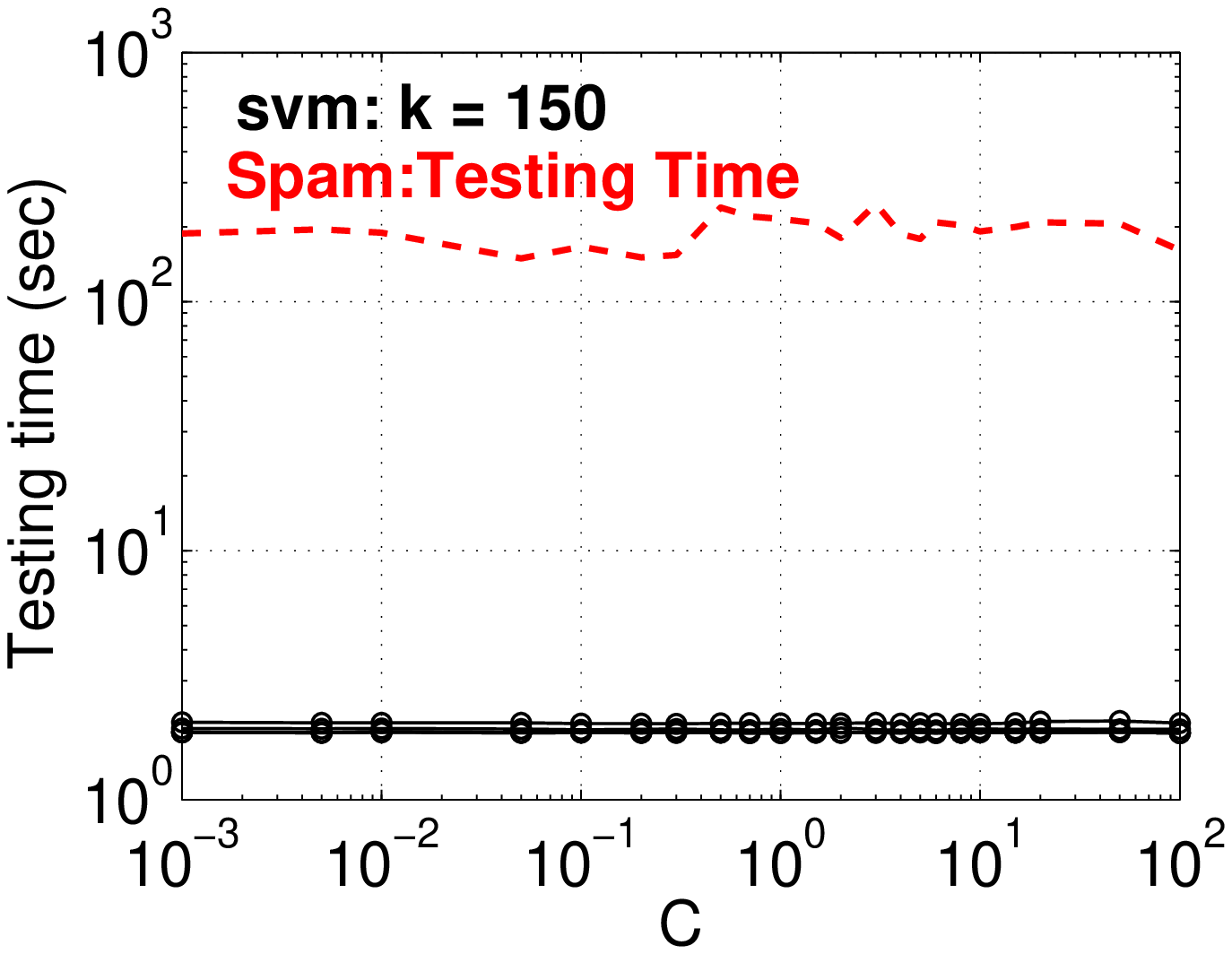}}

\mbox{
\includegraphics[width=1.7in]{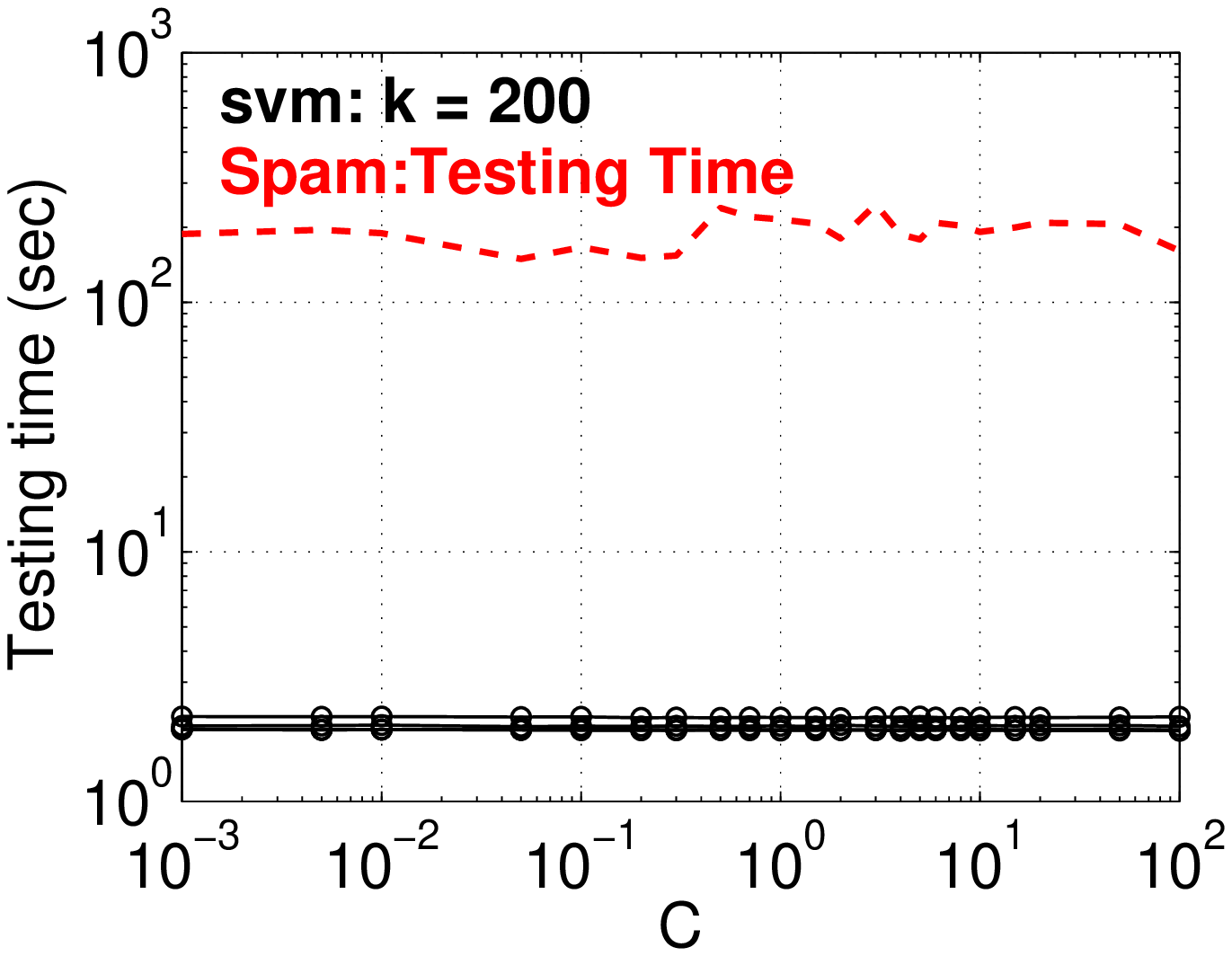}\hspace{-0.1in}
\includegraphics[width=1.7in]{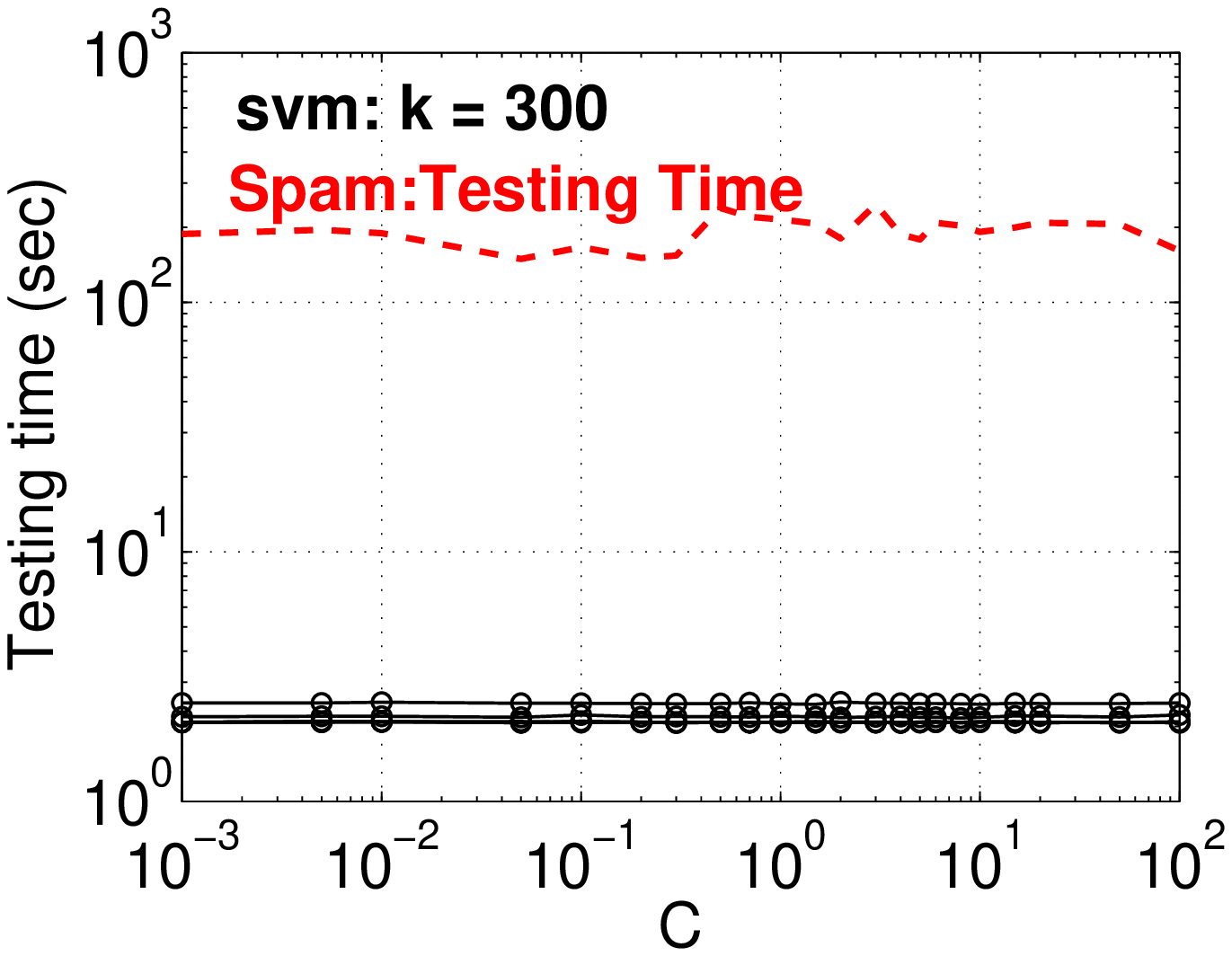}\hspace{-0.1in}
\includegraphics[width=1.7in]{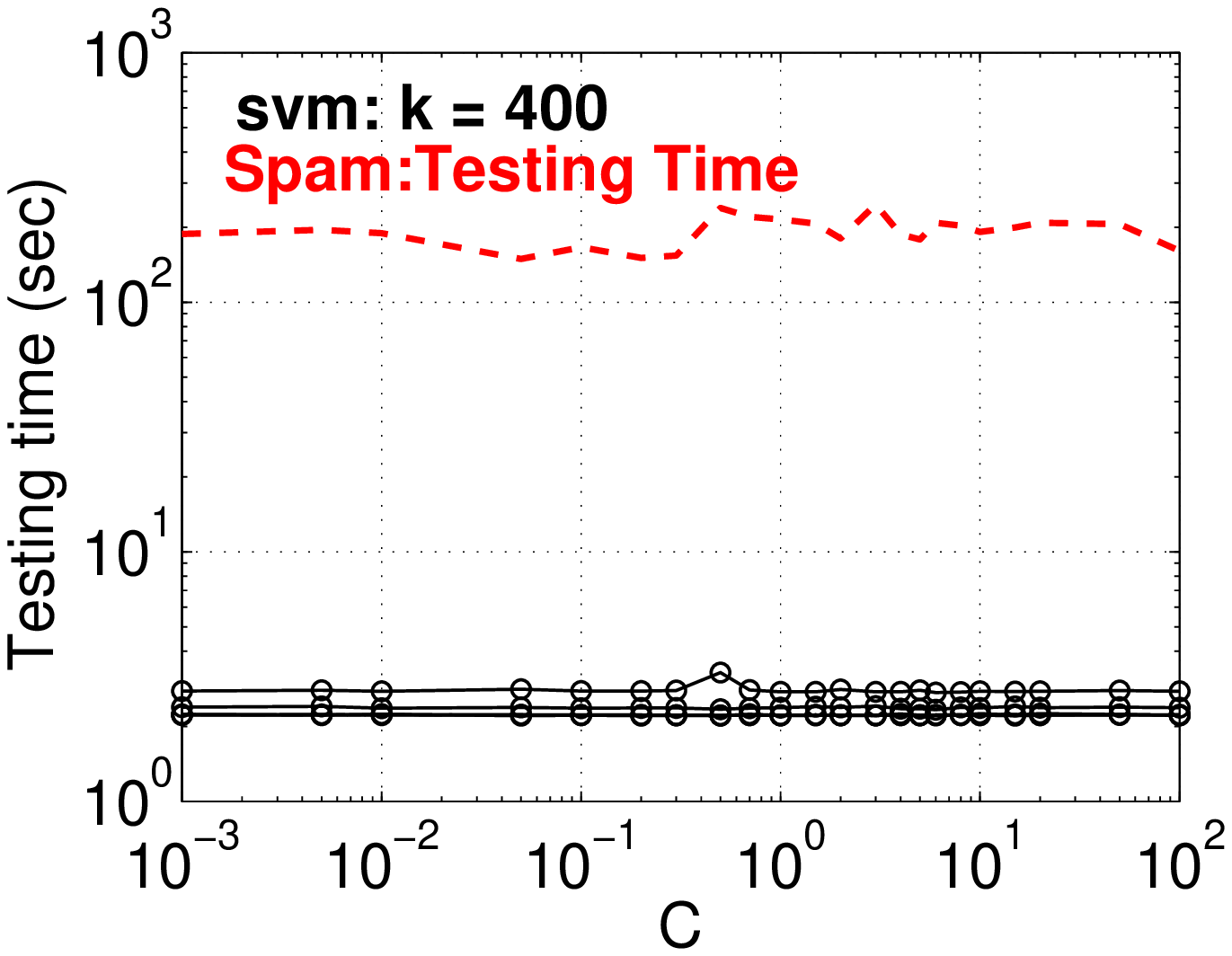}\hspace{-0.1in}
\includegraphics[width=1.7in]{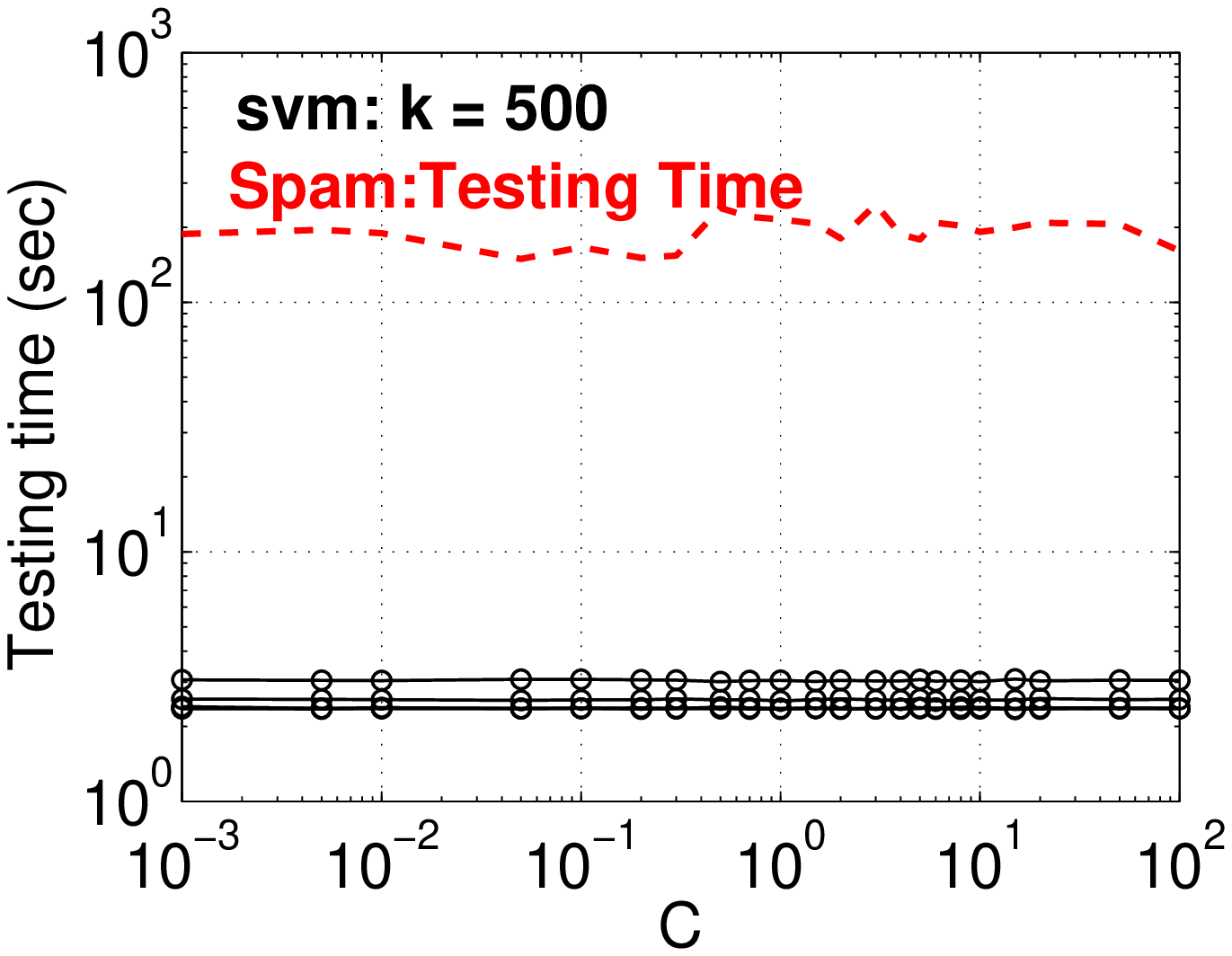}}

\vspace{-0.15in}

\caption{\textbf{Linear SVM  testing time}. The original costs are plotted using dashed (red, if color is available) curves.}\label{fig_testing}
\end{figure}

Compared with the original testing time (about $100\sim 200$ seconds), we can see from Figure~\ref{fig_testing} that the testing time of our method is merely about  1 or 2 seconds. Note that the testing time includes both the data loading time and computing time, as designed by LIBLINEAR. The efficiency of testing may be very important in practice, for example, when the classifier is deployed in an user-facing application (such as search), while the cost of training or pre-processing (such as hashing) may be less critical and can often be conducted off-line.


\subsection{ Experimental Results on Logistic Regression}

\begin{figure}[h!]

\mbox{
\includegraphics[width=1.7in]{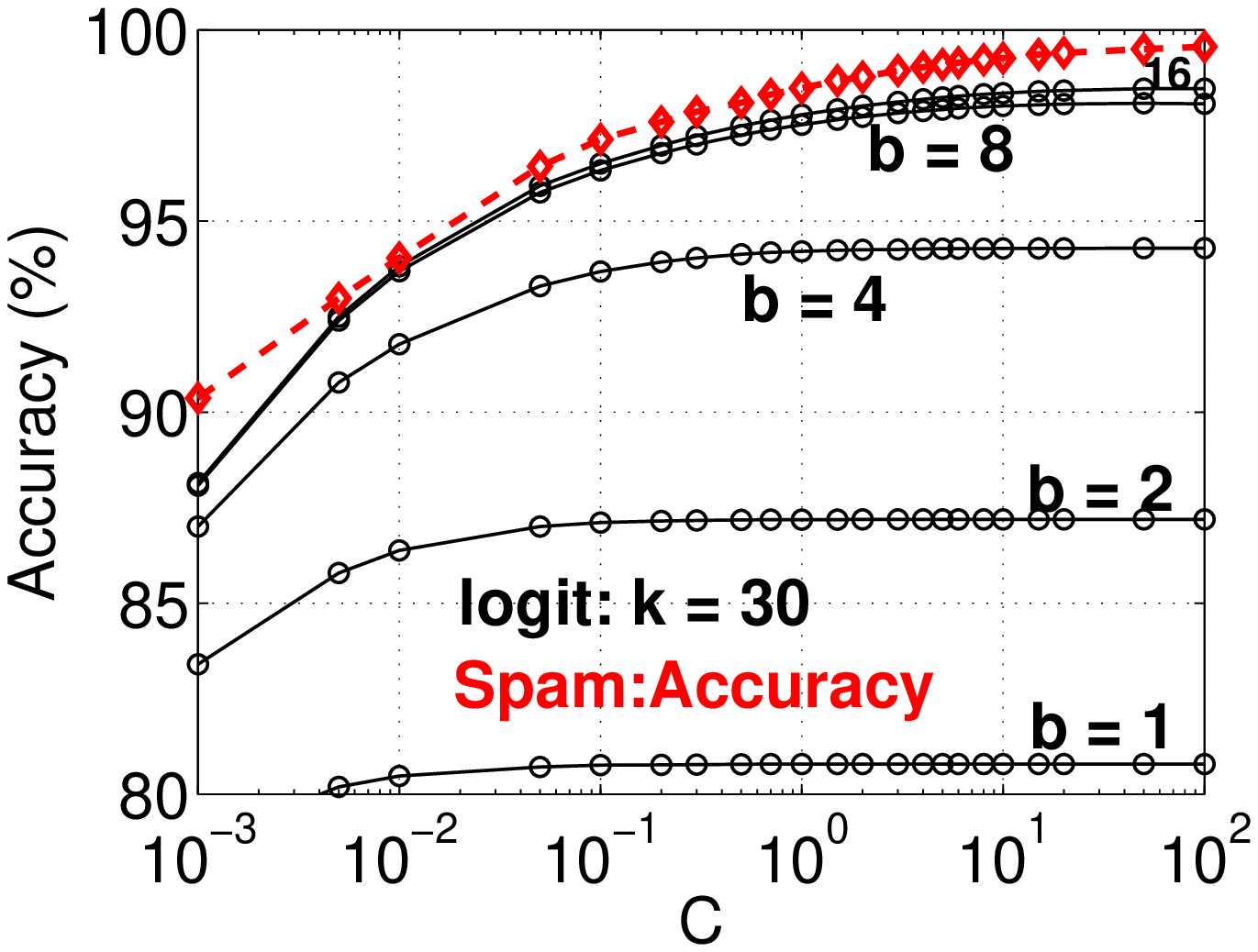}\hspace{-0.1in}
\includegraphics[width=1.7in]{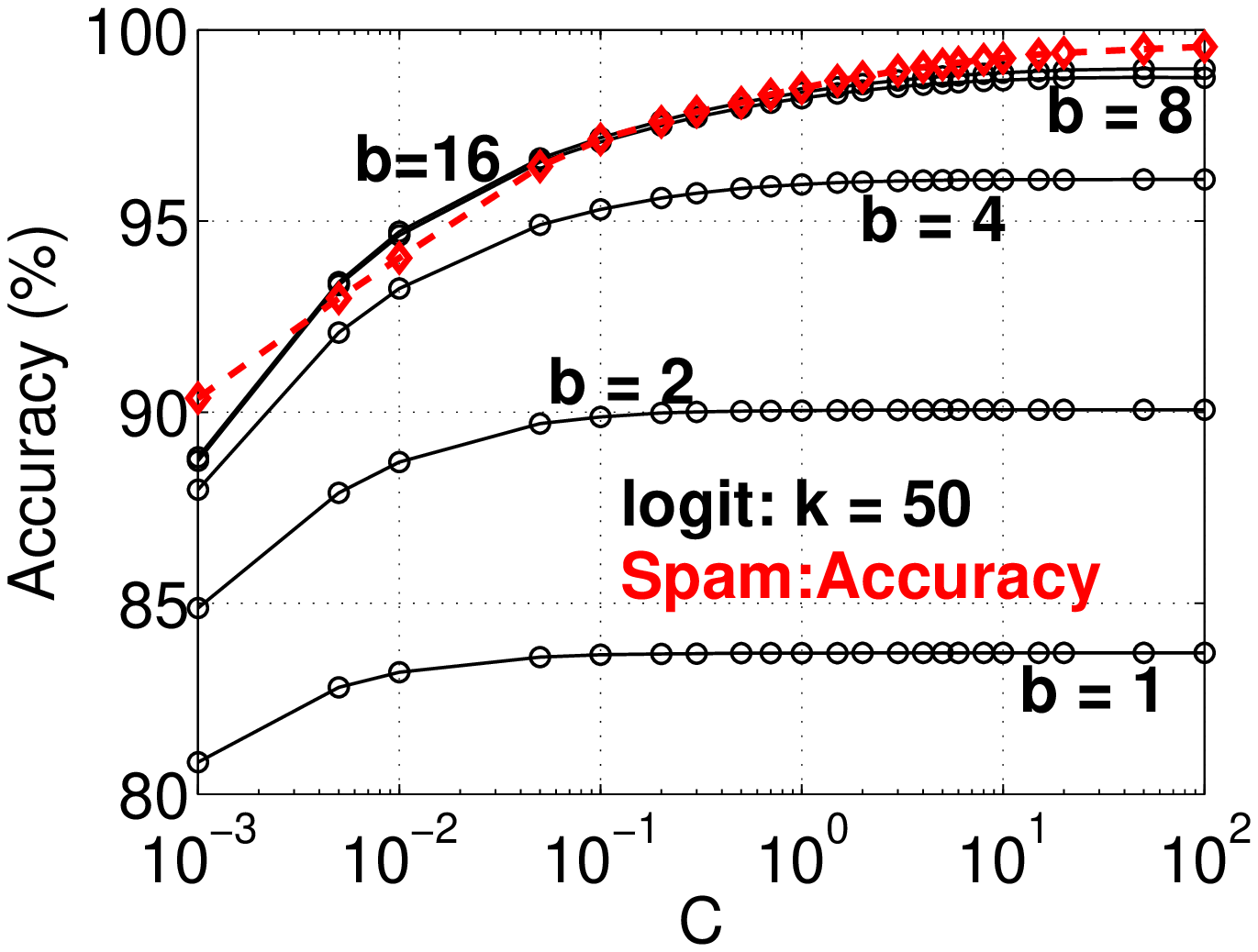}\hspace{-0.1in}
\includegraphics[width=1.7in]{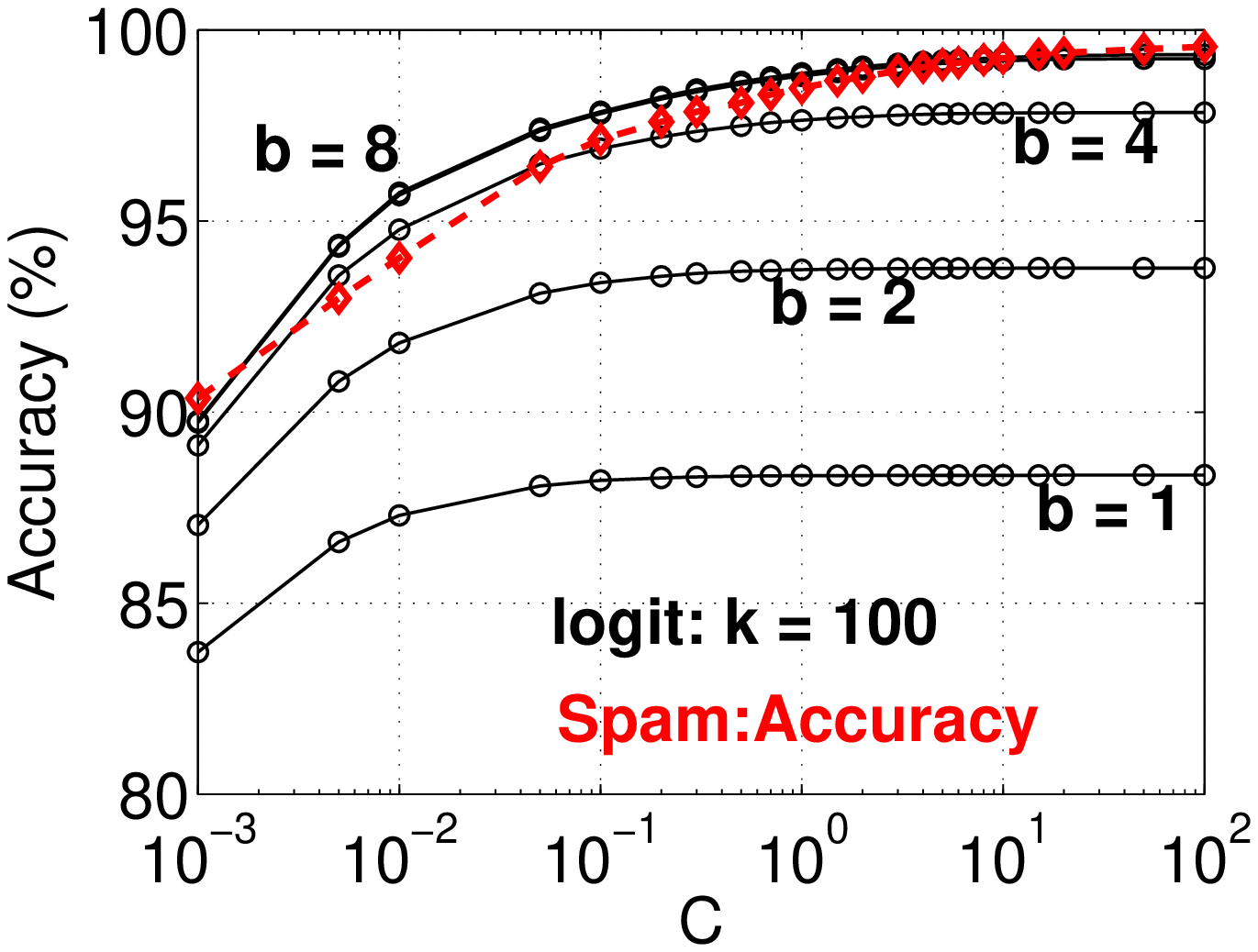}\hspace{-0.1in}
\includegraphics[width=1.7in]{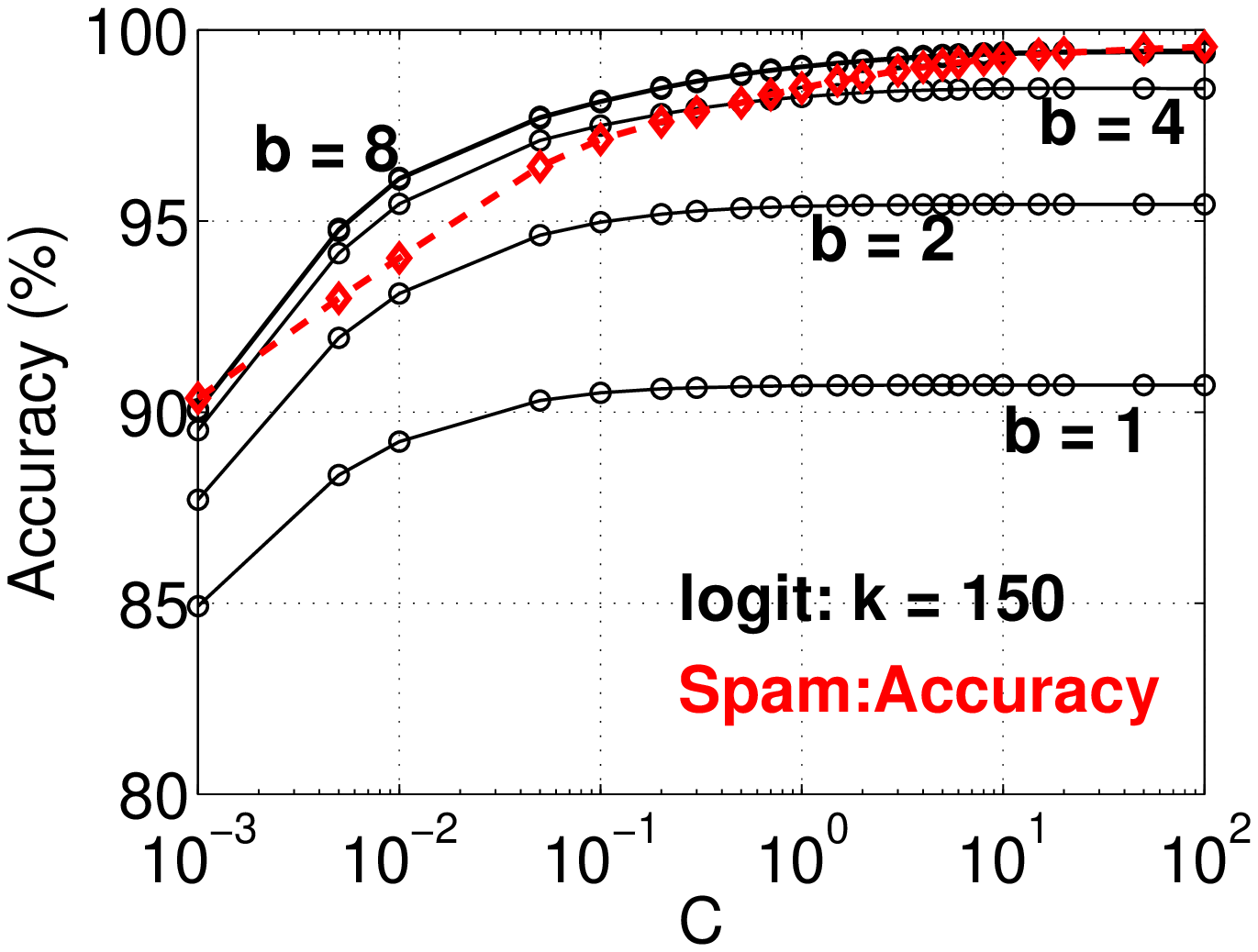}}

\mbox{
\includegraphics[width=1.7in]{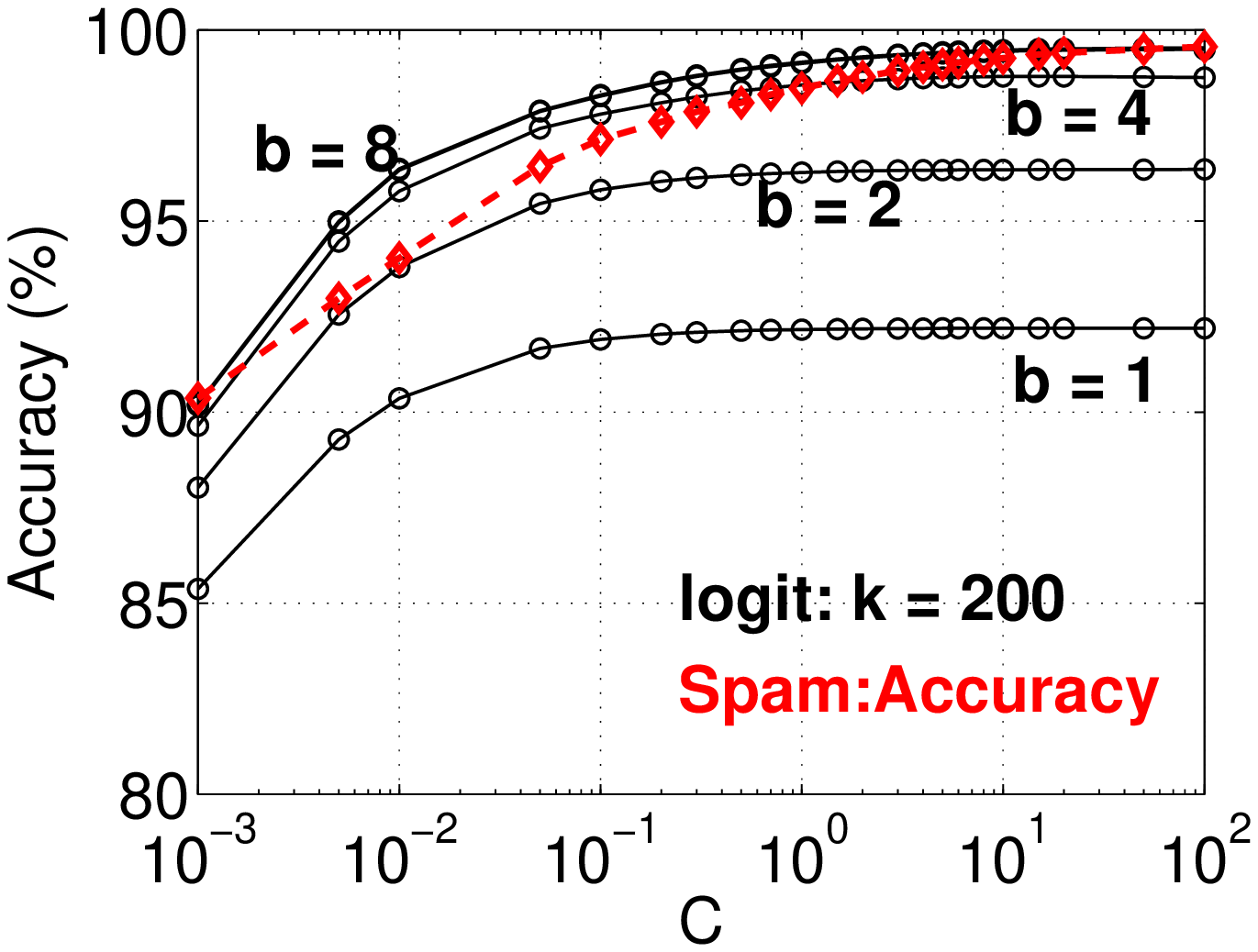}\hspace{-0.1in}
\includegraphics[width=1.7in]{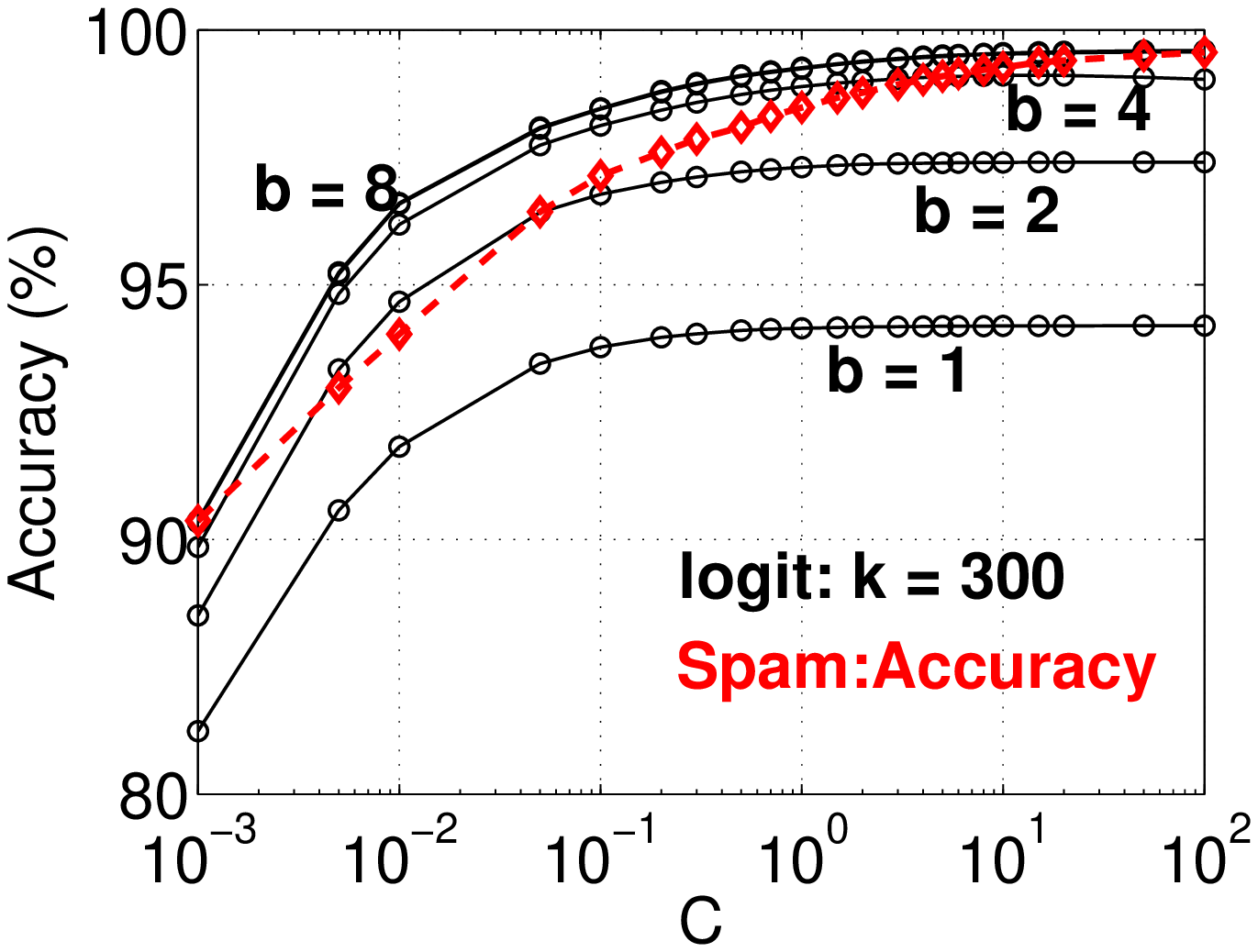}\hspace{-0.1in}
\includegraphics[width=1.7in]{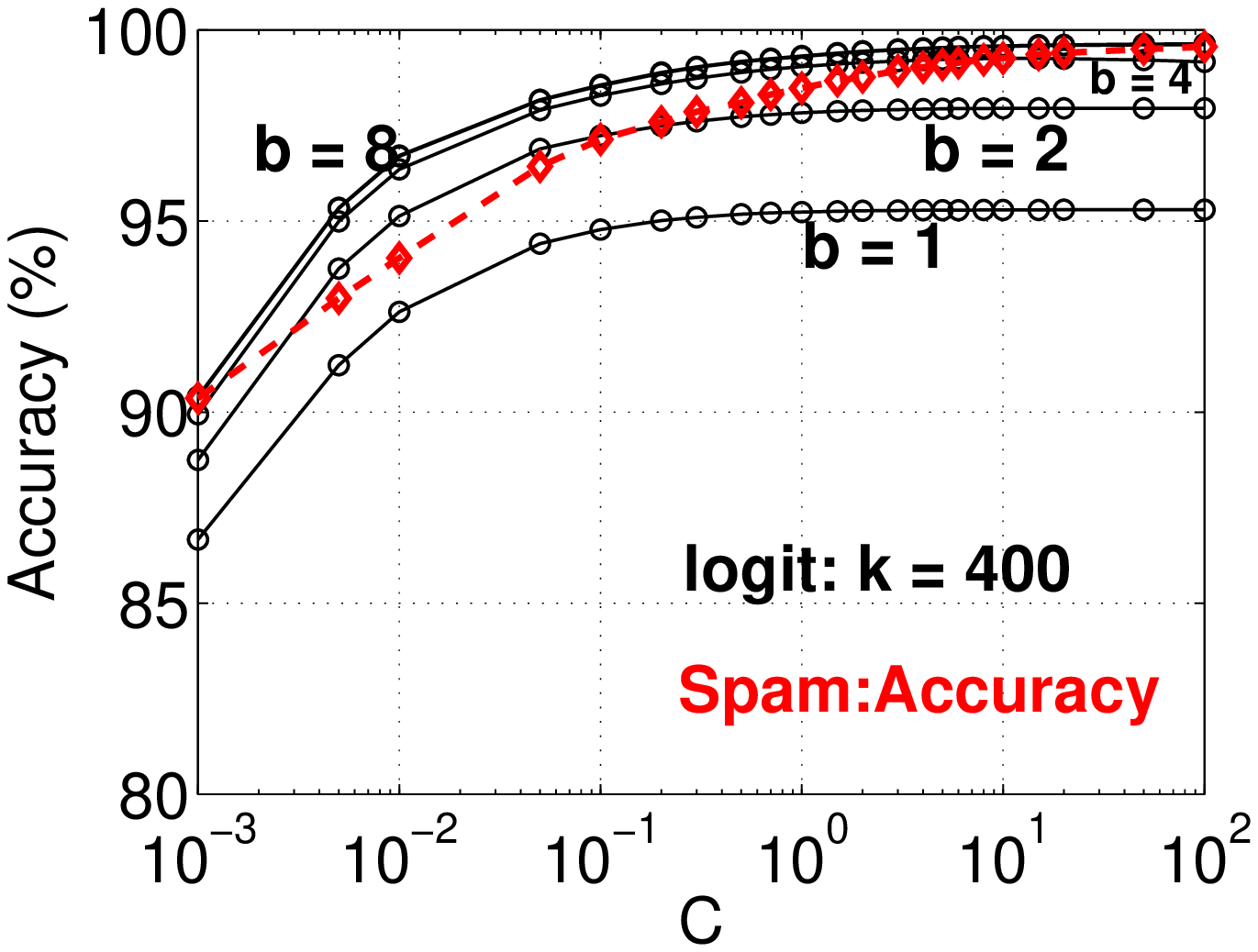}\hspace{-0.1in}
\includegraphics[width=1.7in]{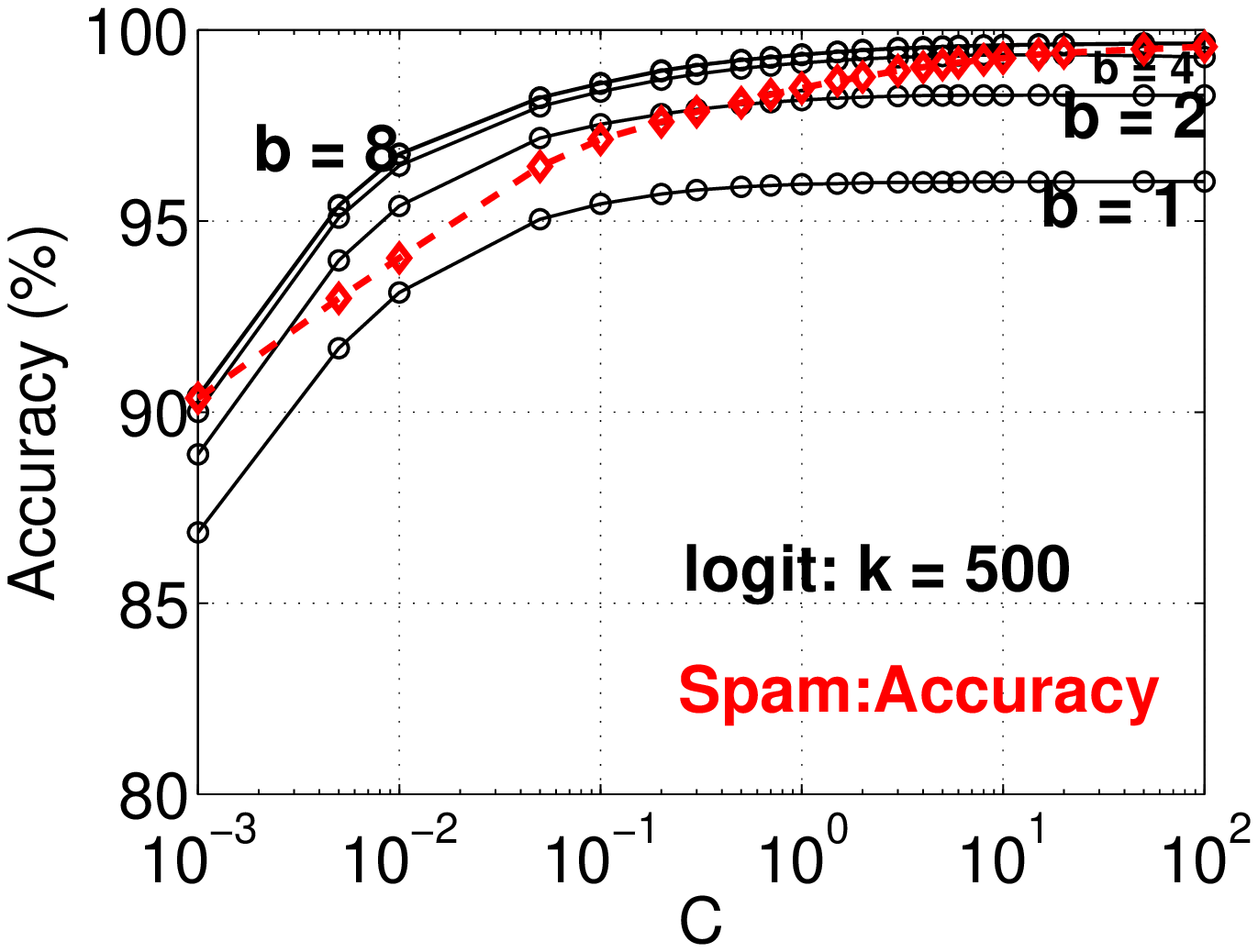}}

\vspace{-0.15in}
\caption{\textbf{Logistic regression test accuracy}. The dashed (red if color is available) curves represents the results using the original data.}\label{fig_acc_logit}
\end{figure}

\begin{figure}[h!]

\mbox{
\includegraphics[width=1.7in]{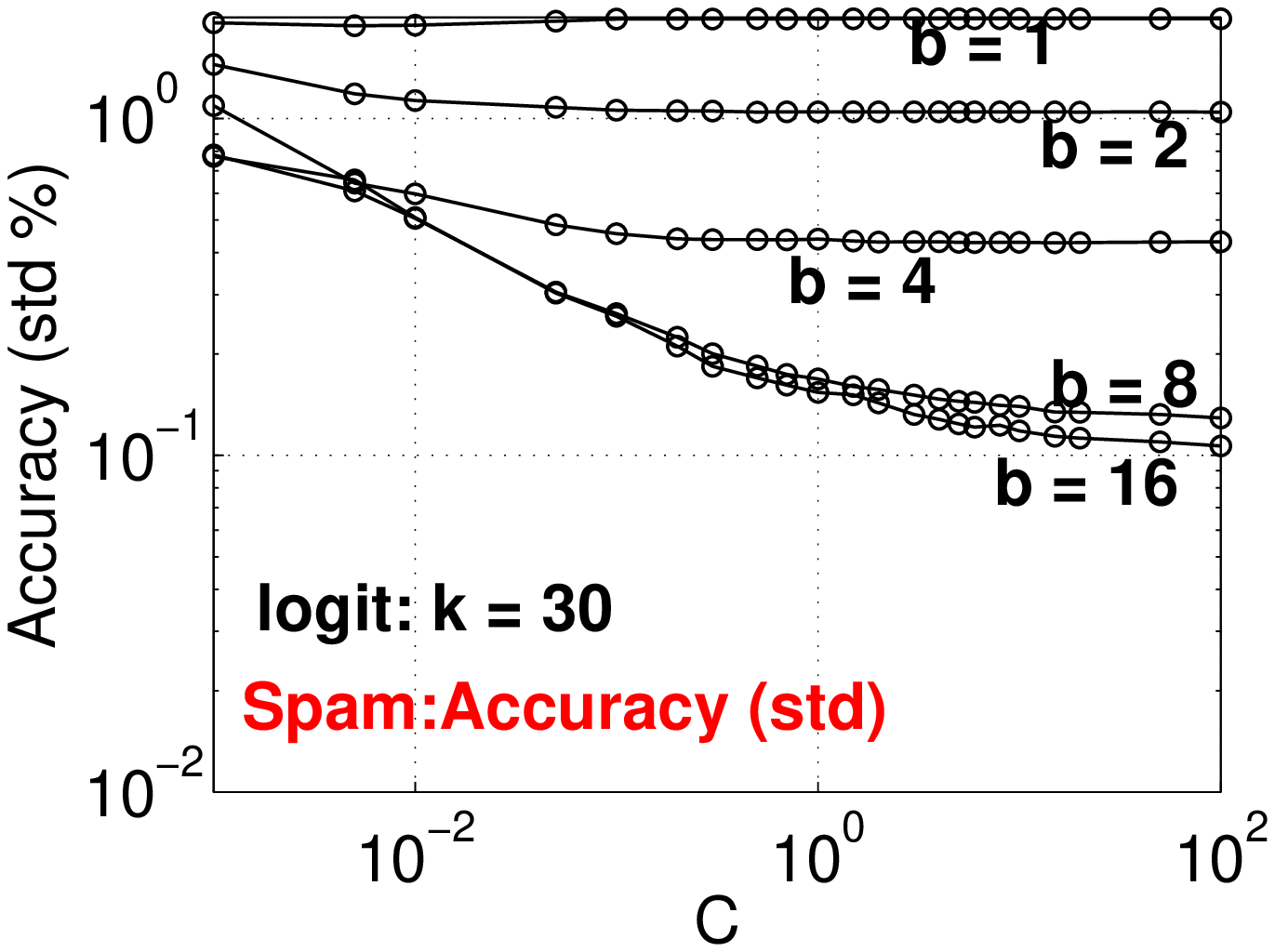}\hspace{-0.1in}
\includegraphics[width=1.7in]{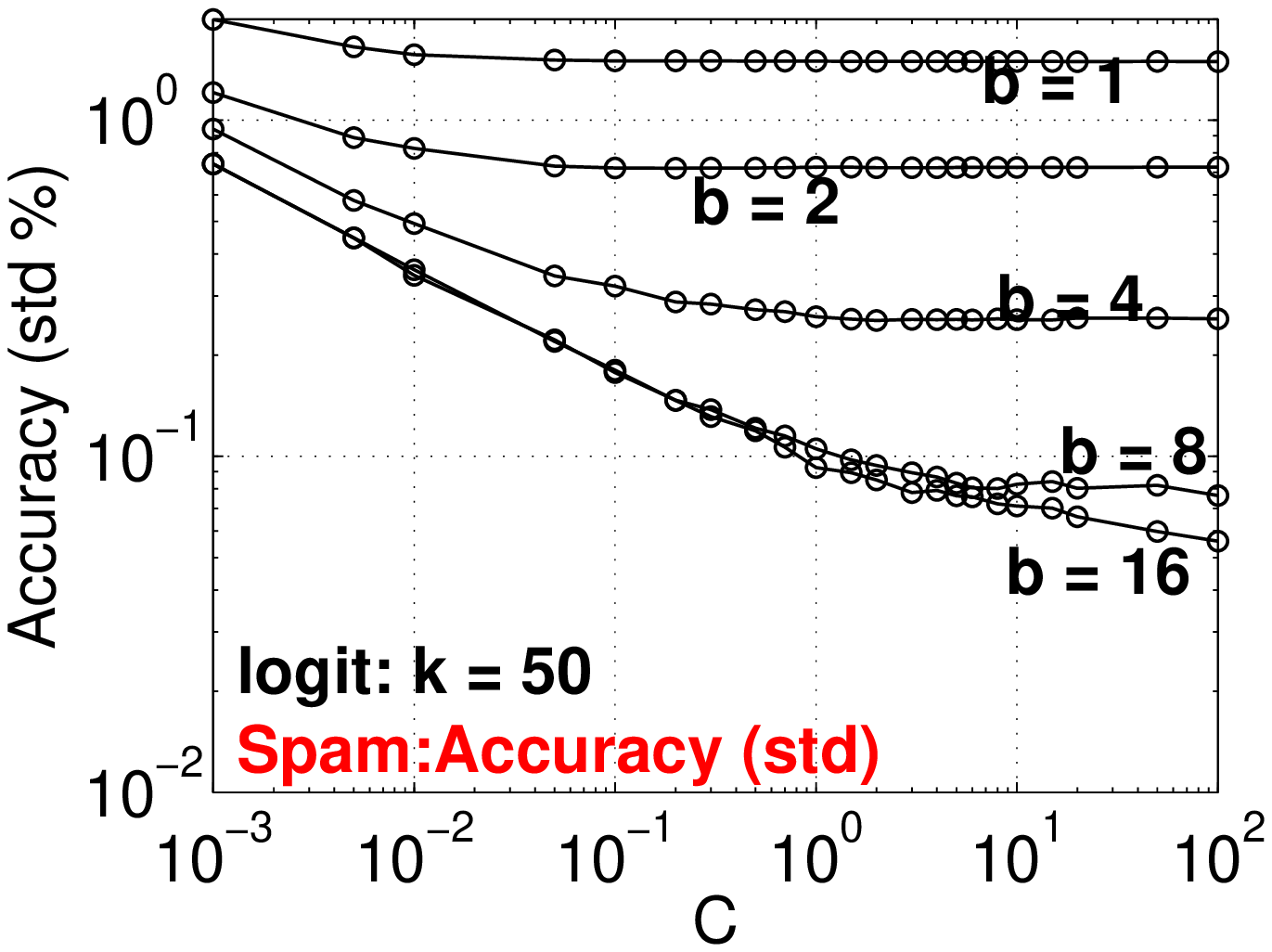}\hspace{-0.1in}
\includegraphics[width=1.7in]{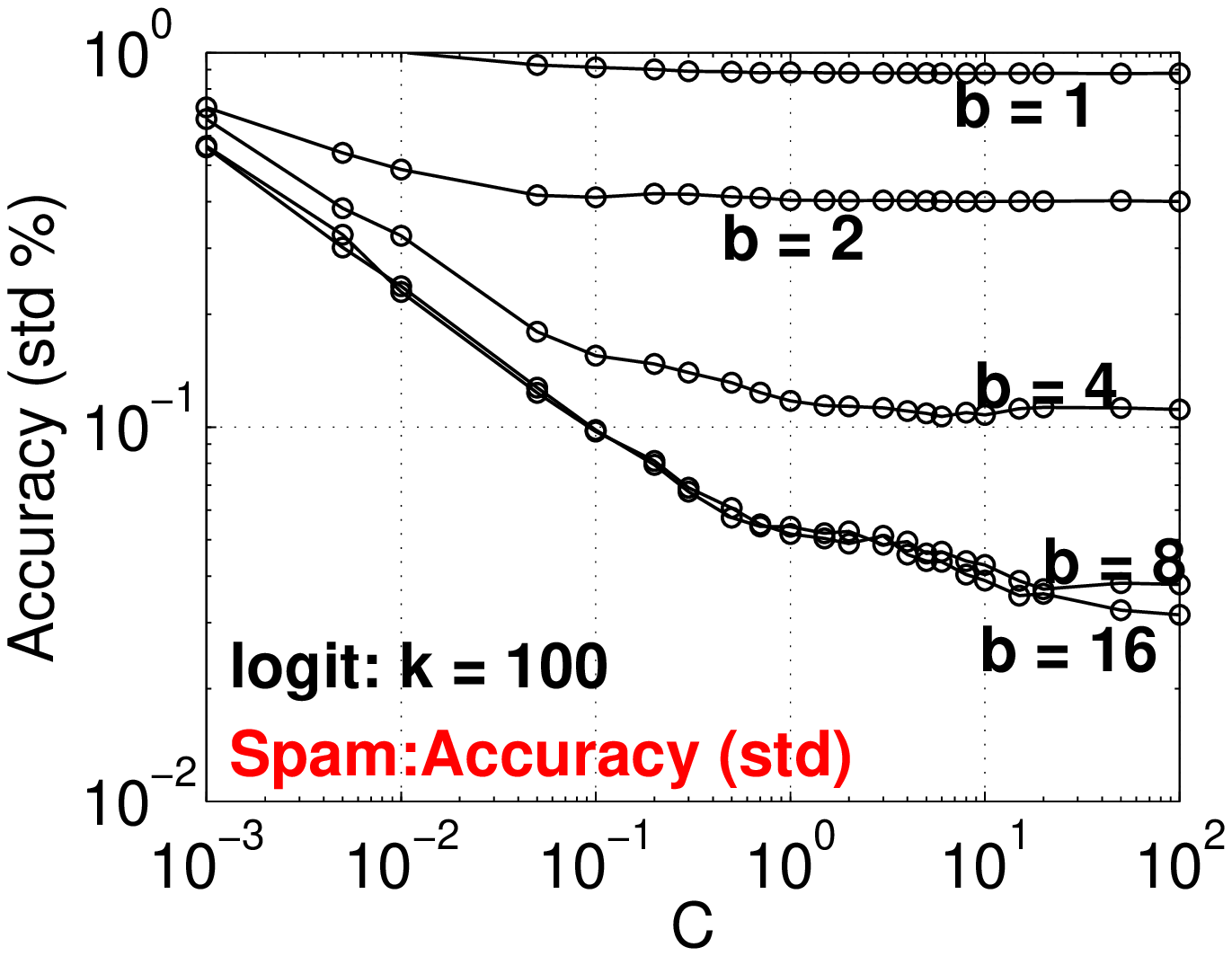}\hspace{-0.1in}
\includegraphics[width=1.7in]{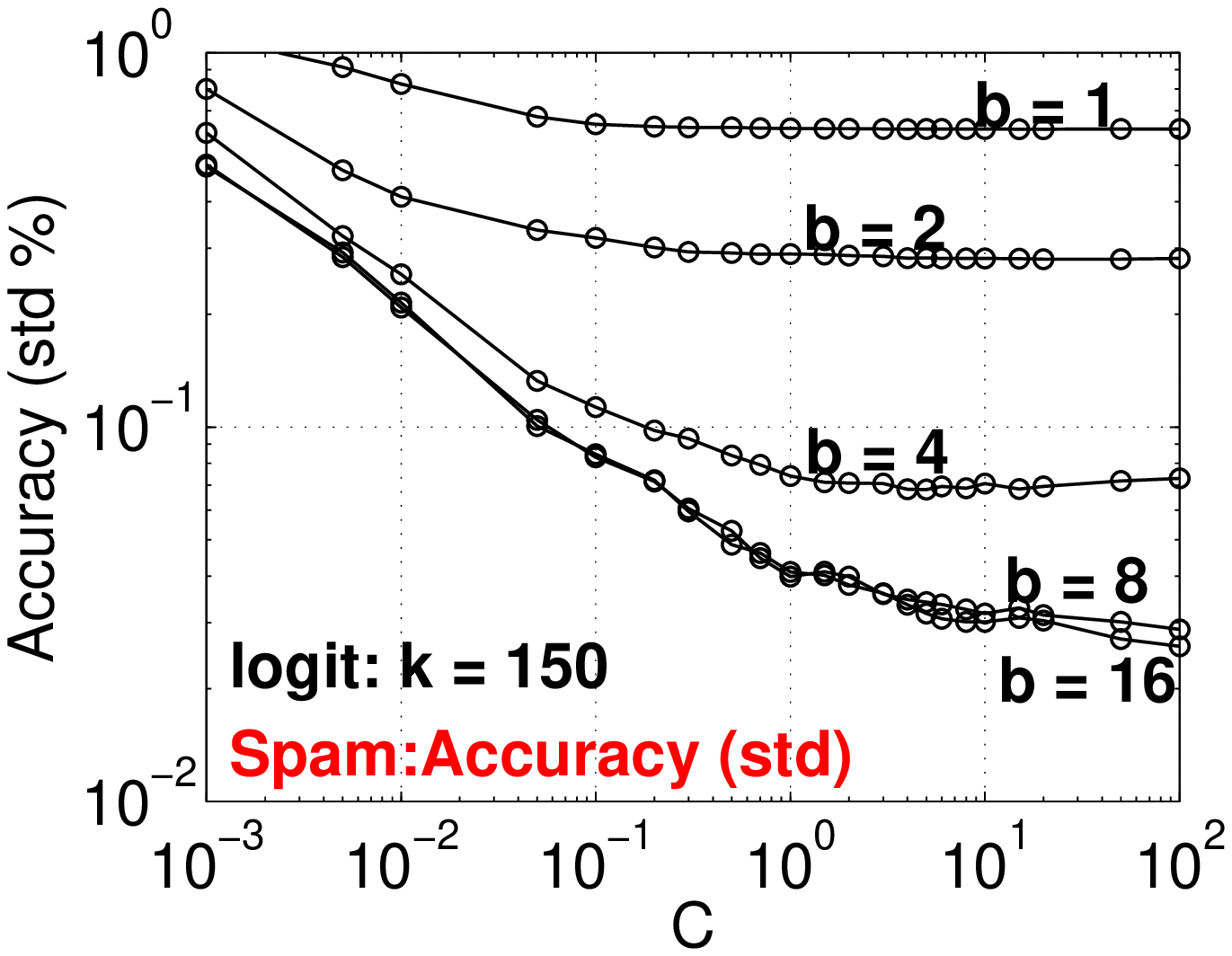}}

\mbox{
\includegraphics[width=1.7in]{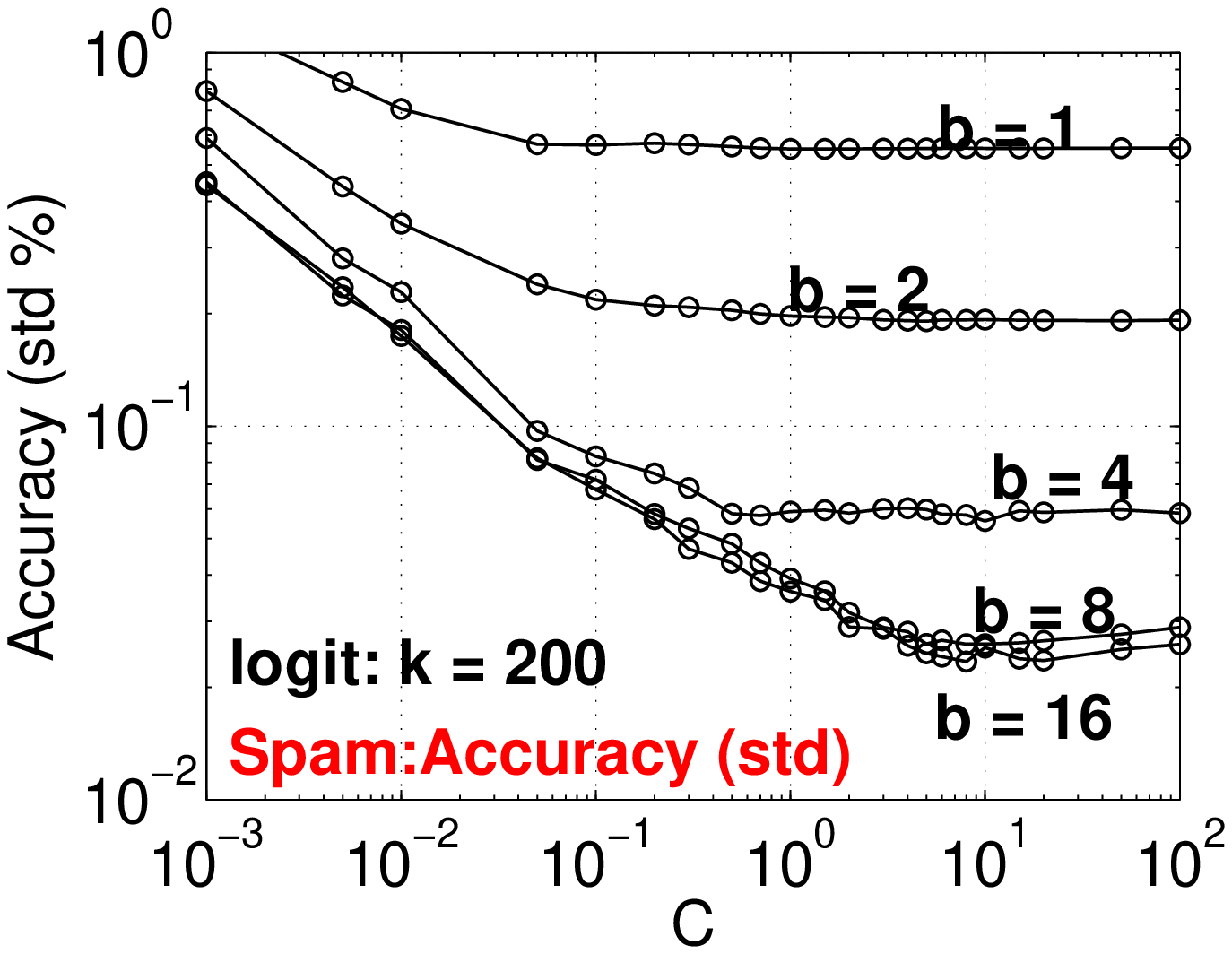}\hspace{-0.1in}
\includegraphics[width=1.7in]{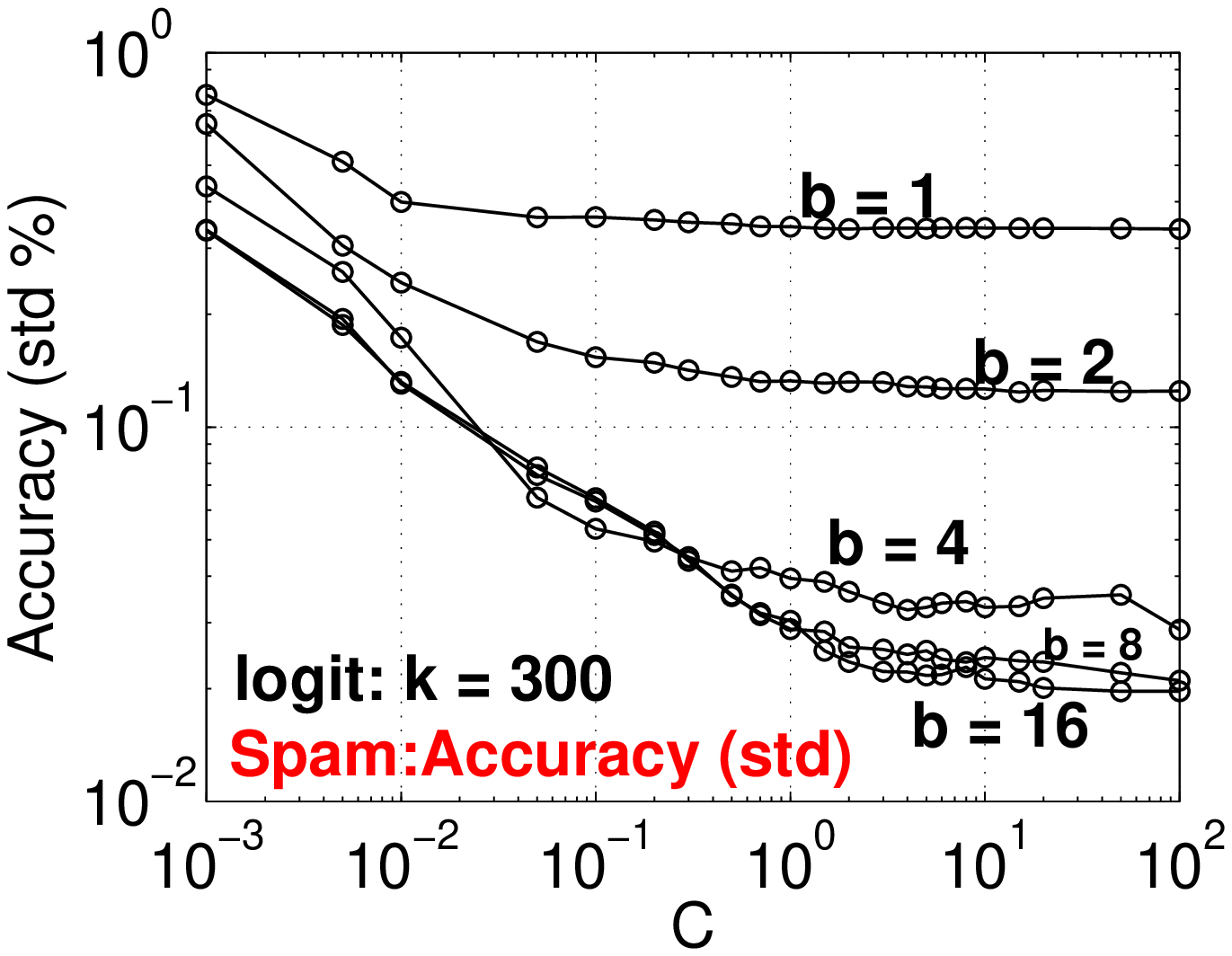}\hspace{-0.1in}
\includegraphics[width=1.7in]{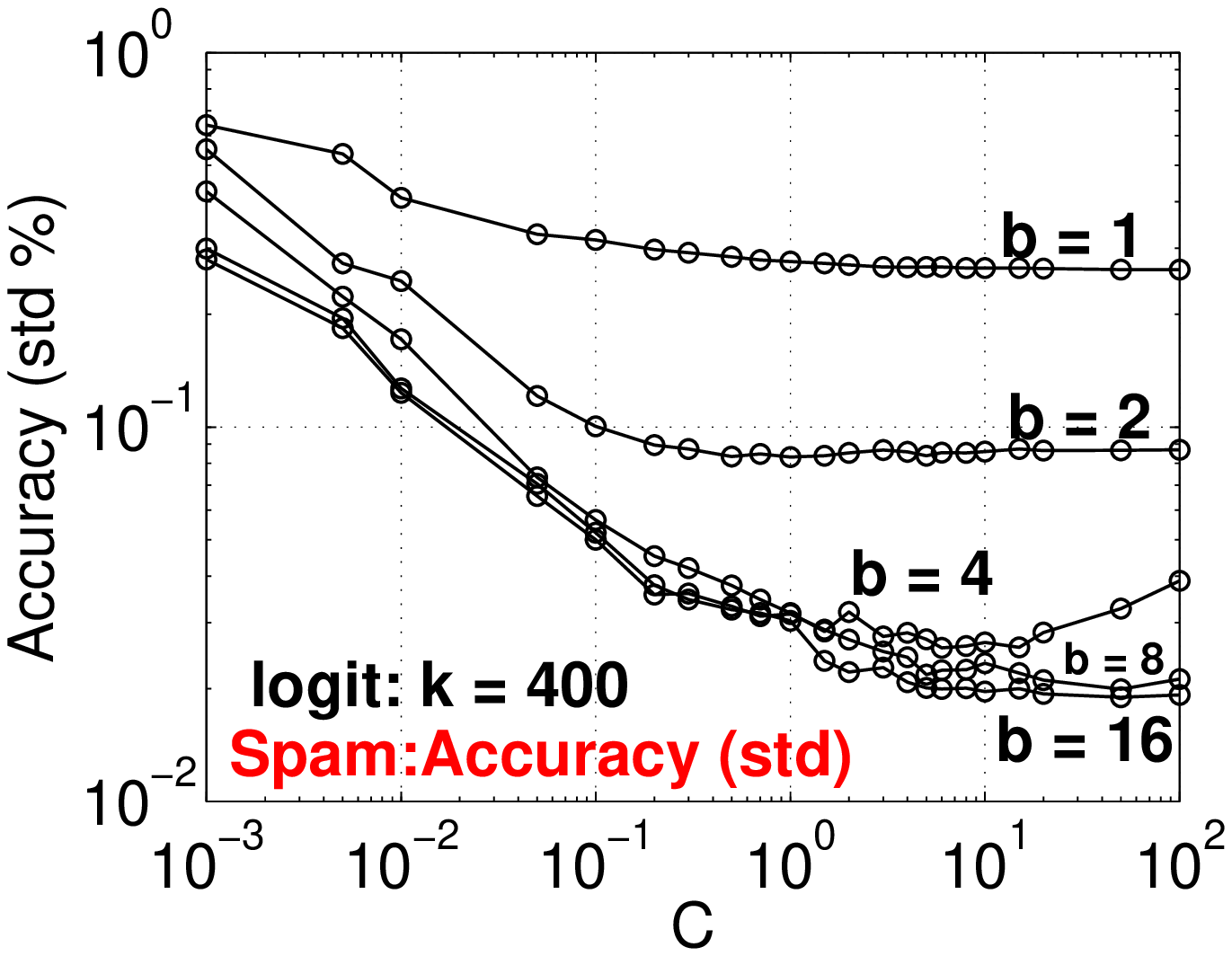}\hspace{-0.1in}
\includegraphics[width=1.7in]{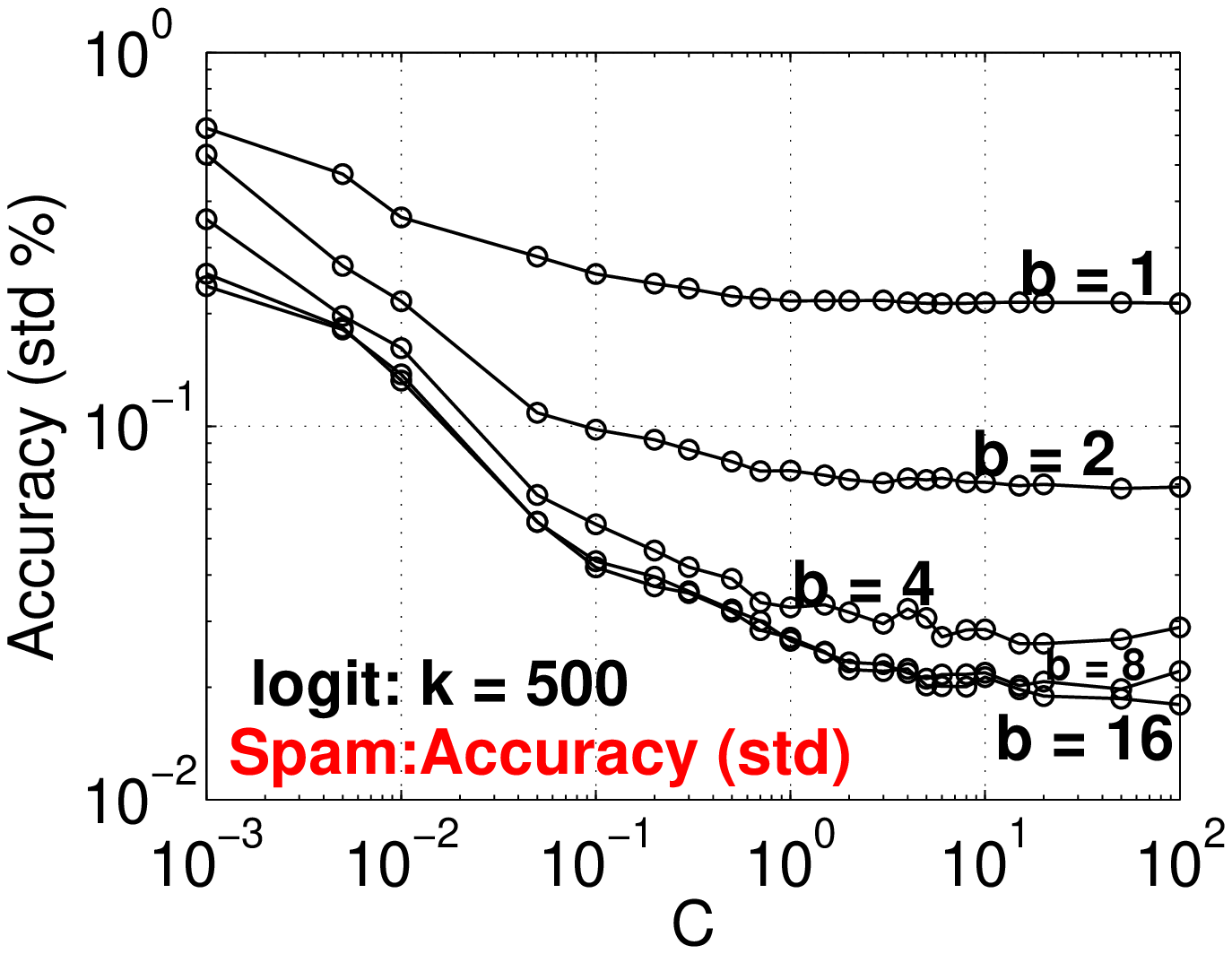}}

\vspace{-0.15in}

\caption{\textbf{Logistic regression test accuracy (std)}. The standard deviations are computed from 50 repetitions.  }\label{fig_acc_std_logit}
\end{figure}

\begin{figure}[h!]

\mbox{
\includegraphics[width=1.7in]{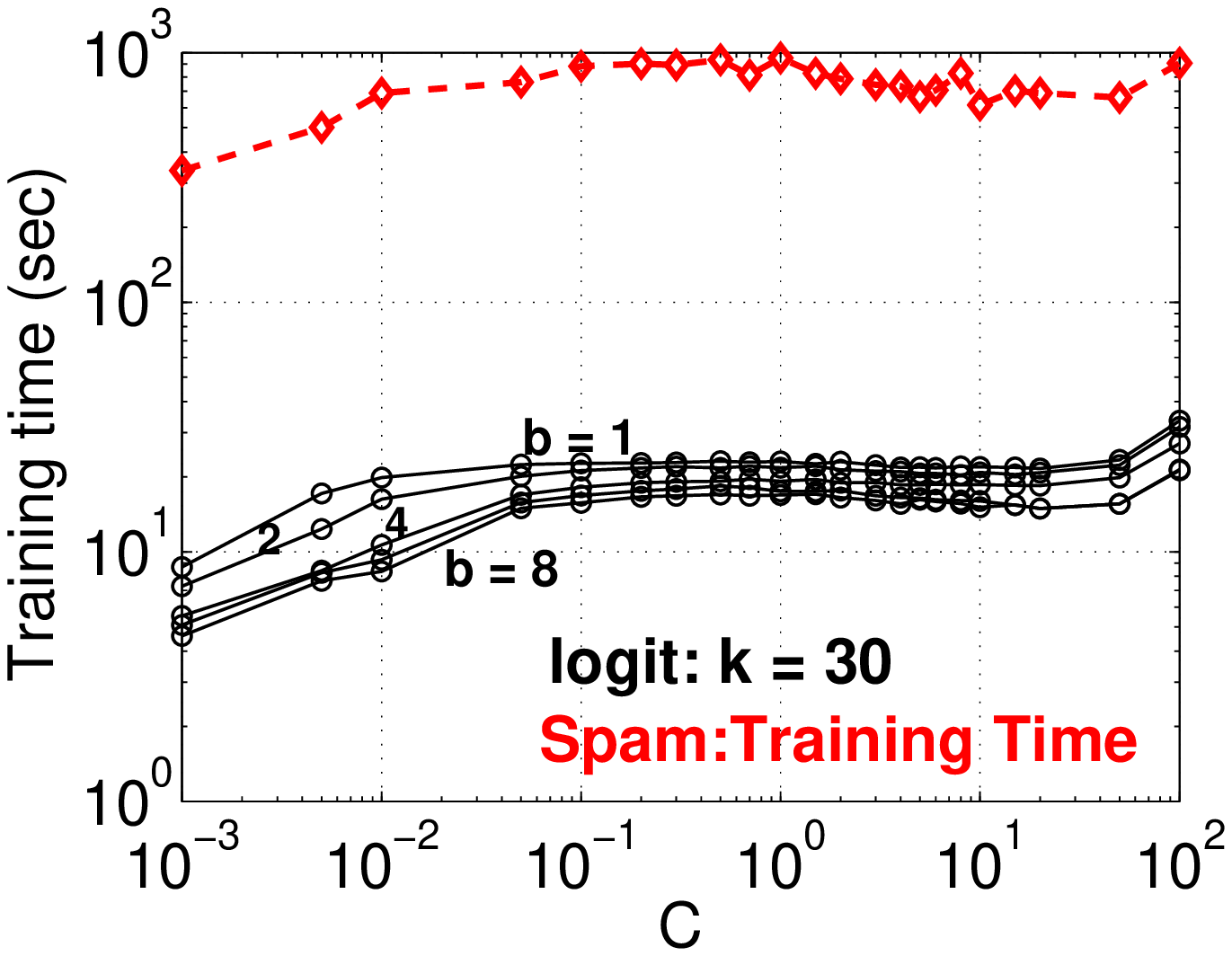}\hspace{-0.1in}
\includegraphics[width=1.7in]{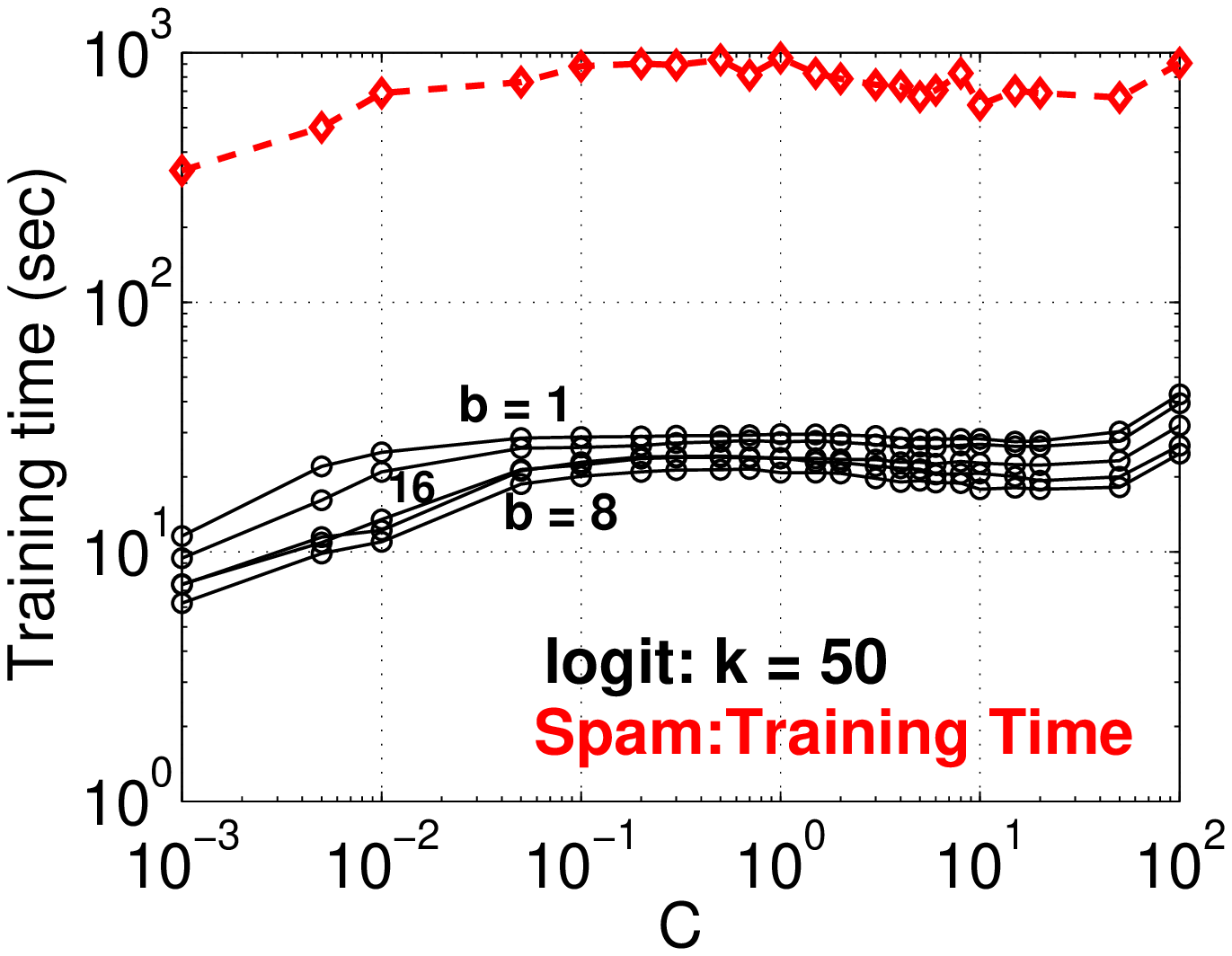}\hspace{-0.1in}
\includegraphics[width=1.7in]{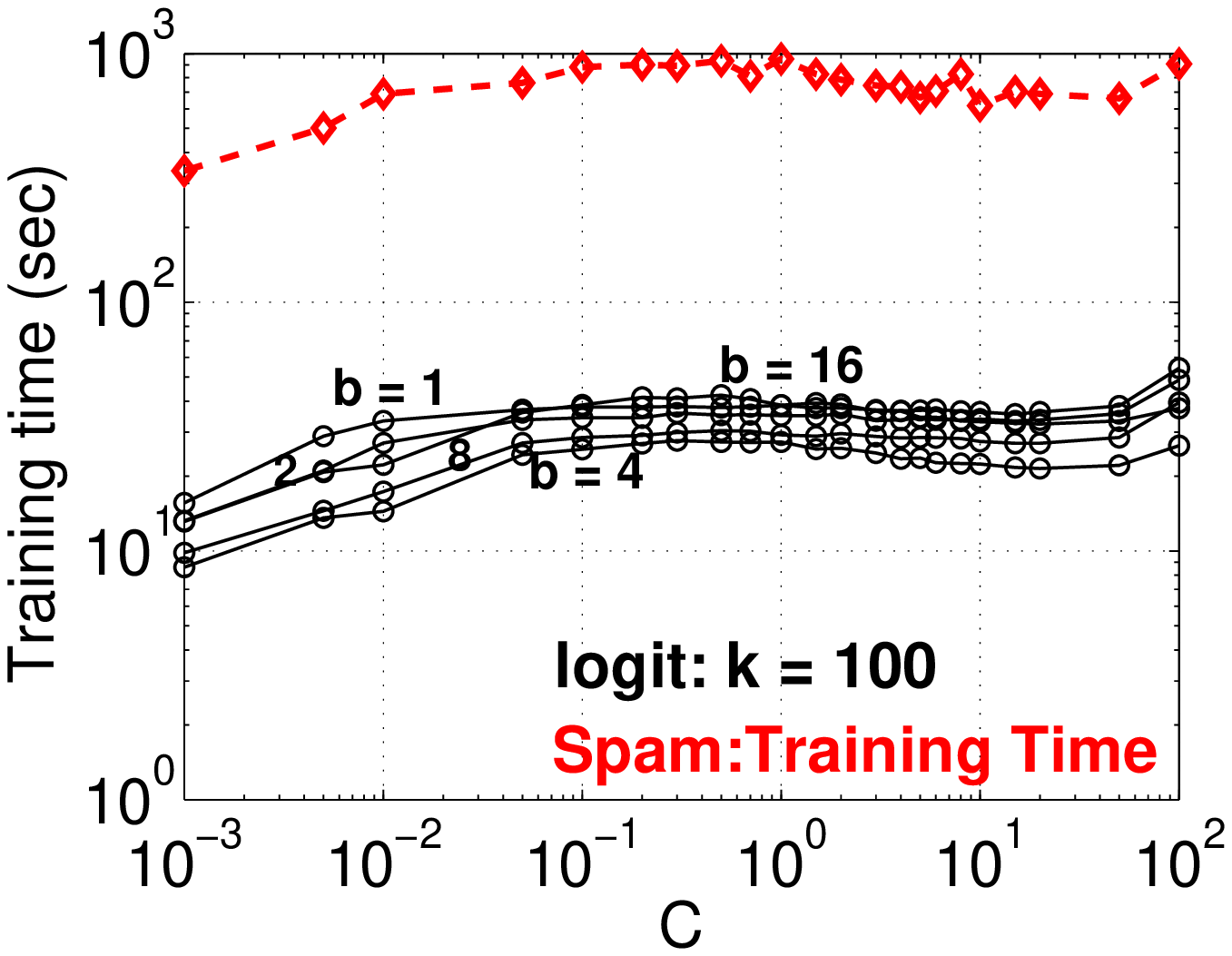}\hspace{-0.1in}
\includegraphics[width=1.7in]{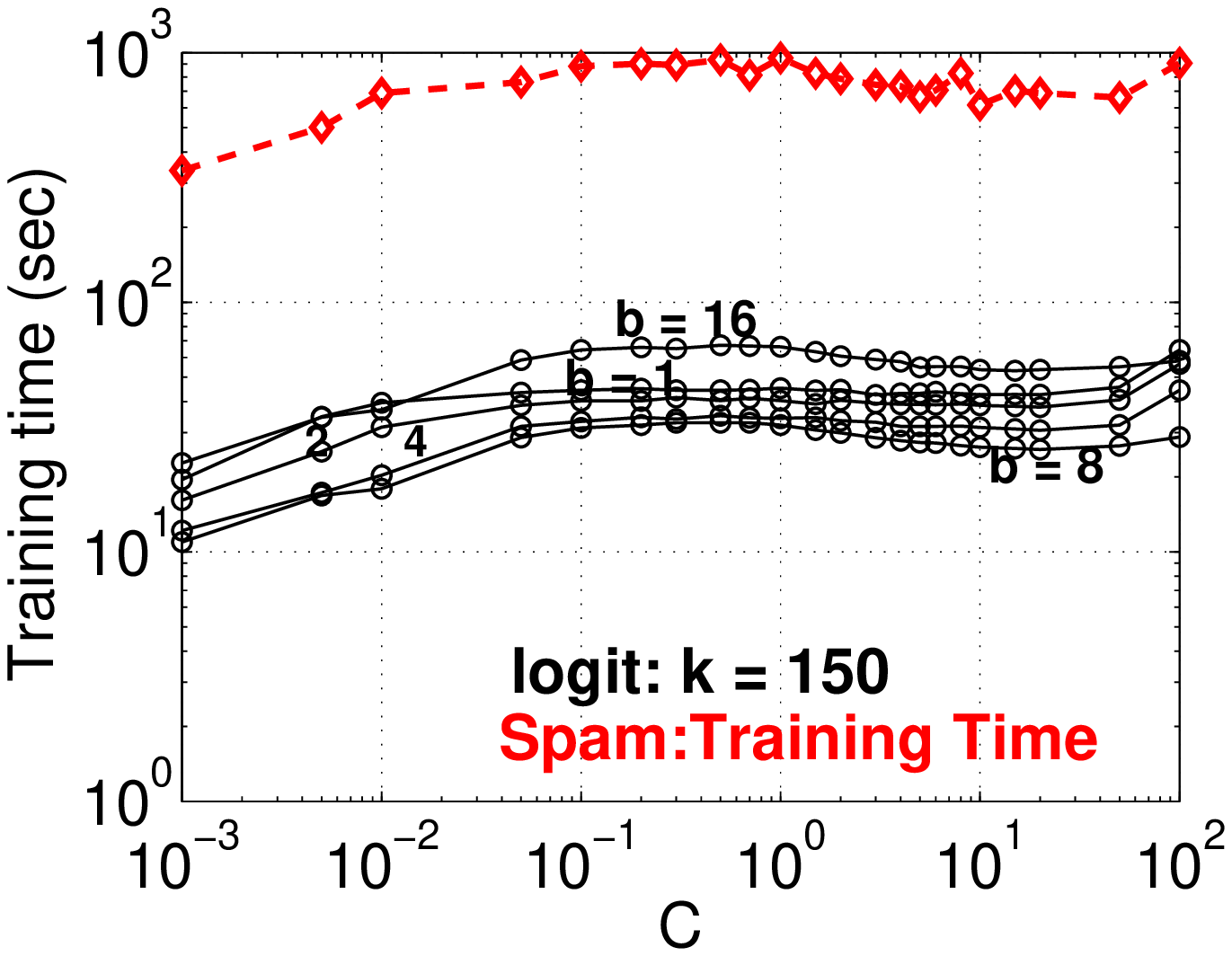}}

\mbox{
\includegraphics[width=1.7in]{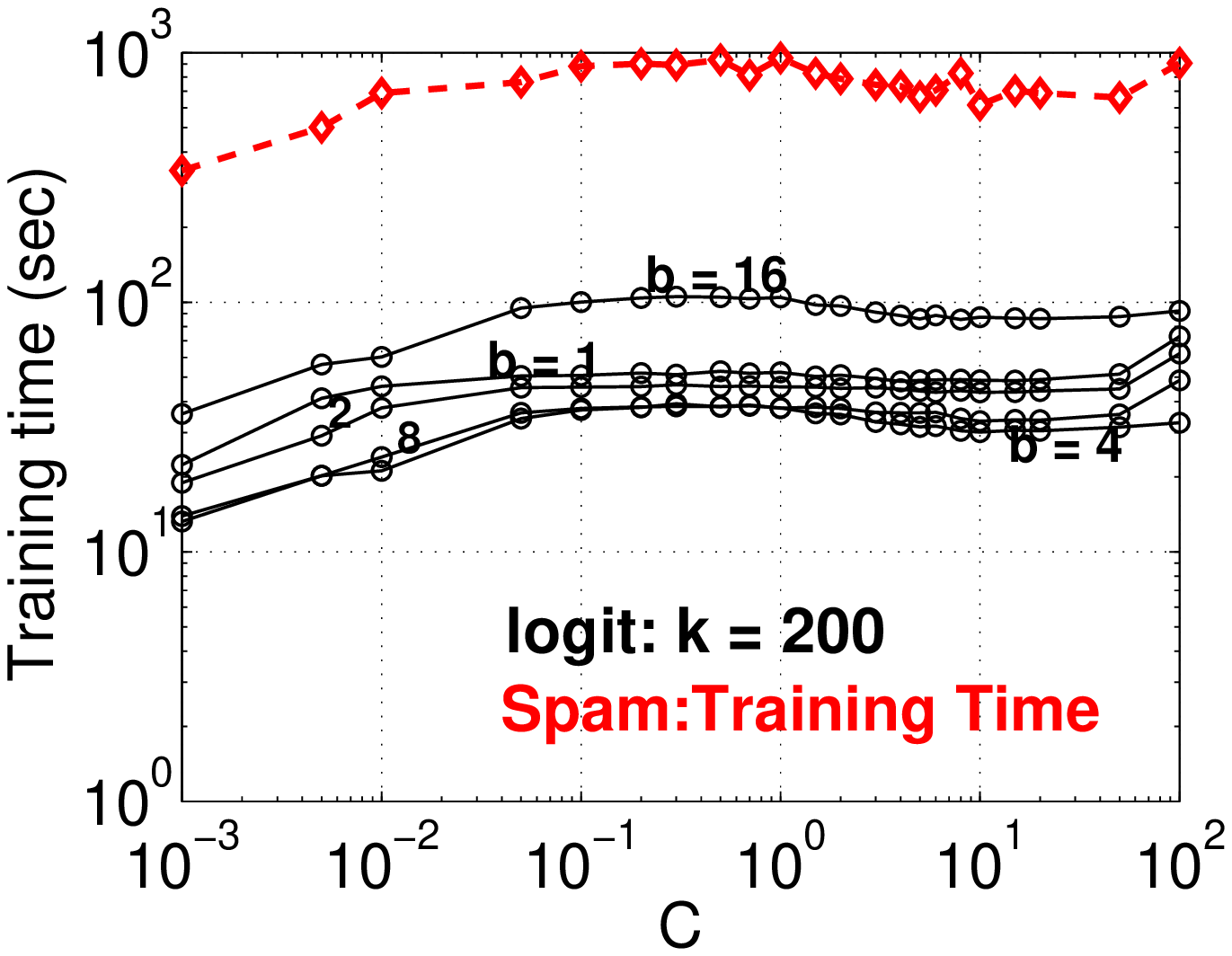}\hspace{-0.1in}
\includegraphics[width=1.7in]{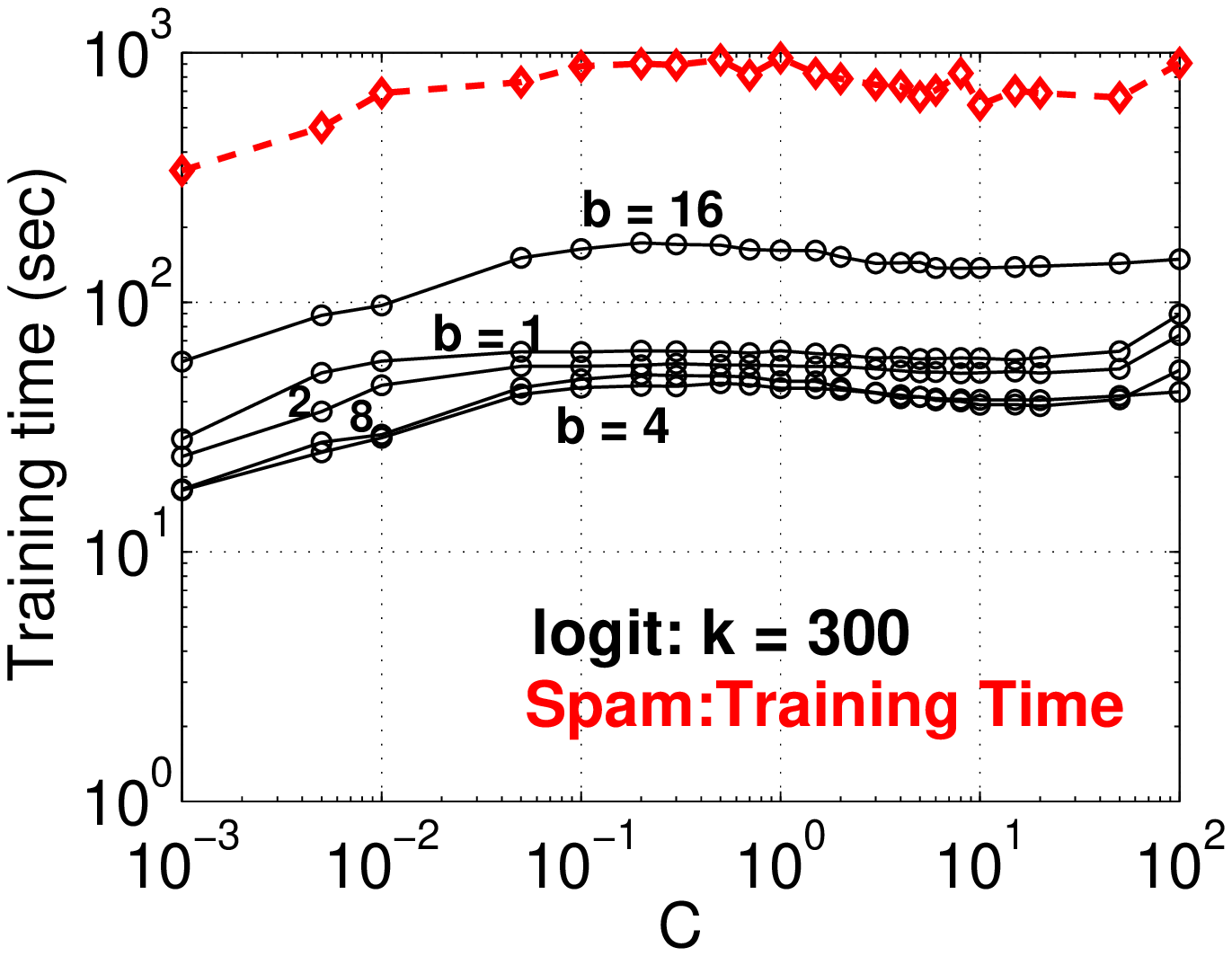}\hspace{-0.1in}
\includegraphics[width=1.7in]{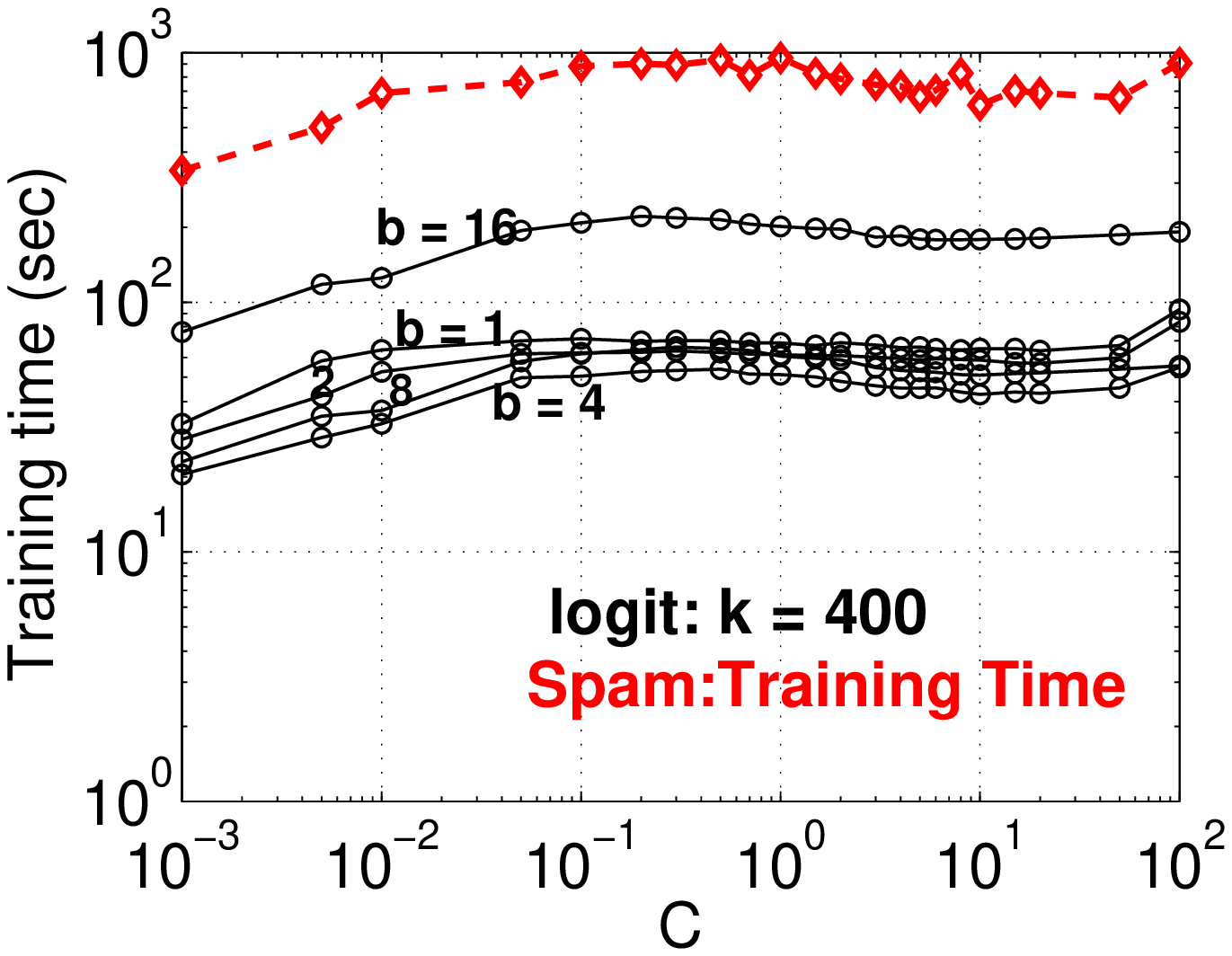}\hspace{-0.1in}
\includegraphics[width=1.7in]{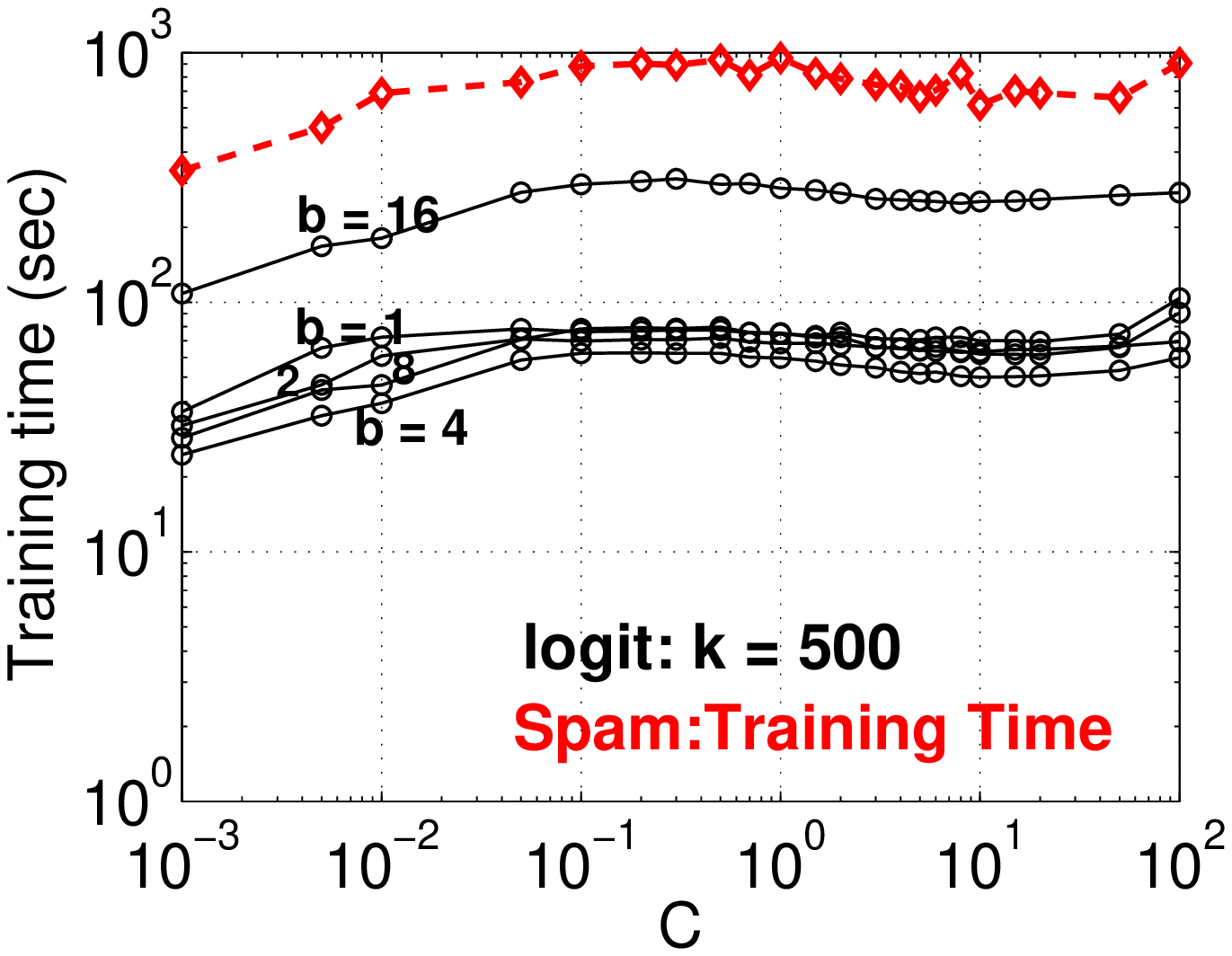}}

\vspace{-0.15in}

\caption{\textbf{Logistic Regression Training time}. The dashed (red if color is available) curves represents the results using the original data.}\label{fig_training_logit}
\end{figure}

Figure~\ref{fig_acc_logit} presents the test accuracies and Figure~\ref{fig_training_logit} presents the training time using logistic regression. Again, with $k\geq 150$ (or even $k\geq 100)$ and $b\geq 8$, $b$-bit minwise hashing can achieve the same test accuracies as using the original data. Figure~\ref{fig_acc_std_logit} presents the standard deviations, which again verify that our algorithm produces stable predictions for logistic regression.

From Figure~\ref{fig_training_logit}, we can see that the training time is substantially reduced, from about 1000 seconds to about $30\sim50$ seconds only (unless $b=16$ and $k$ is large).\\

In summary, it appears $b$-bit hashing is highly effective in reducing the data size and speeding up the training (and testing), for both (nonlinear and linear)  SVM and logistic regression. We notice that when using $b=16$, the training time can be much larger than  using $b\leq 8$. Interestingly, we find that $b$-bit hashing can be combined with  {\em Vowpal Wabbit (VW)}~\cite{Proc:Weinberger_ICML2009} to further reduce the training time, especially when $b$ is large.

\clearpage

\section{Random Projections and Vowpal Wabbit (VW)}

The two methods, random projections~\cite{Article:Achlioptas_JCSS03, Proc:Li_Hastie_Church_KDD06} and Vowpal Wabbit (VW)~\cite{Proc:Weinberger_ICML2009,Article:Shi_JMLR09} are not limited to binary data (although for ultra high-dimensional used in the context of search, the data are often binary). The VW algorithm is also related to the Count-Min sketch~\cite{Article:Cormode_05}. In this paper, we use ''VW`` particularly for the algorithm in~\cite{Proc:Weinberger_ICML2009}.

For convenience, we denote two $D$-dim data vectors by $u_1, u_2\in\mathbb{R}^{D}$. Again, the task is to estimate the inner product $a = \sum_{i=1}^D u_{1,i} u_{2,i}$.

\subsection{Random Projections}

The general idea is to multiply the data vectors, e.g., $u_1$ and $u_2$, by a random matrix $\{r_{ij}\}\in\mathbb{R}^{D\times k}$, where $r_{ij}$ is sampled i.i.d. from the following generic distribution with~\cite{Proc:Li_Hastie_Church_KDD06}
\begin{align}\label{eqn_r_ij}
E(r_{ij}) = 0, \ \ Var(r_{ij}) = 1,\ \ E(r_{ij}^3) = 0,\ \ E(r_{ij}^4) = s, \ \ s\geq 1.
\end{align}
We must have $s\geq 1$ because $Var(r_{ij}^2) = E(r_{ij}^4) - E^2(r_{ij}^2) = s-1 \geq 0$.

This generates two $k$-dim vectors, $v_1$ and $v_2$:
\begin{align}\notag
v_{1,j} = \sum_{i=1}^D u_{1,i}r_{ij},\hspace{0.2in} v_{2,j} = \sum_{i=1}^D u_{2,i}r_{ij}, \ \ \ j = 1, 2, ..., k
\end{align}

The general distributions which satisfy (\ref{eqn_r_ij}) includes the standard normal distribution (in this case, $s=3$) and the ``sparse projection'' distribution specified as
\begin{align}\label{eqn_sparse_r}
r_{ij} = \sqrt{s}\times\left\{\begin{array}{ll} 1 & \text{with prob.}\ \frac{1}{2s} \\ 0 & \text{with prob.}\ 1-\frac{1}{s}\\ -1 & \text{with prob.}\ \frac{1}{2s} \end{array}\right.
\end{align}

\cite{Proc:Li_Hastie_Church_KDD06} provided the following unbiased estimator $\hat{a}_{rp,s}$ of $a$ and the general variance formula:
\begin{align}
&\hat{a}_{rp,s} = \frac{1}{k}\sum_{j=1}^k v_{1,j}v_{2,j}, \hspace{0.5in} E(\hat{a}_{rp,s}) = a = \sum_{i=1}^D u_{1,i} u_{2,i},\\\label{eqn_var_rp}
&Var(\hat{a}_{rp,s}) = \frac{1}{k}\left[\sum_{i=1}^D u_{1,i}^2\sum_{i=1}^D u_{2,i}^2 + \left(\sum_{i=1}^D u_{1,i}u_{2,i}\right)^2 +(s-3)\sum_{i=1}^D u_{1,i}^2u_{2,i}^2\right]
\end{align}
which means $s=1$ achieves the smallest  variance. The only elementary distribution we know that satisfies (\ref{eqn_r_ij}) with $s=1$ is the two point distribution in $\{-1, 1\}$ with equal probabilities, i.e., (\ref{eqn_sparse_r}) with $s=1$.

\subsection{Vowpal Wabbit (VW)}

Again, in this paper, ``VW'' always refers to the particular algorithm in~\cite{Proc:Weinberger_ICML2009}. VW may be viewed as a ``bias-corrected'' version of the  Count-Min (CM) sketch algorithm~\cite{Article:Cormode_05}. In the original CM  algorithm,  the key step  is to independently and uniformly hash elements of the data vectors to buckets $\in\{1, 2, 3, ..., k\}$ and the hashed value is the sum of the elements in the bucket. That is $h(i) = j$ with probability $\frac{1}{k}$, where $j \in\{1, 2, ..., k\}$. For convenience, we introduce an indicator function:
\begin{align}\notag
I_{ij} =\left\{\begin{array}{ll}
1 &\text{if } h(i) = j\\
0 &\text{otherwise}
\end{array}
\right.\end{align}
which allow  us to write the hashed data as
\begin{align}\notag
w_{1,j} = \sum_{i=1}^D u_{1,i}I_{ij},\hspace{0.5in} w_{2,j} = \sum_{i=1}^D u_{2,i}I_{ij}
\end{align}

The estimate $\hat{a}_{cm} = \sum_{j=1}^k w_{1,j} w_{2,j}$ is (severely) biased for the task of estimating the inner products. The original paper~\cite{Article:Cormode_05} suggested a ``count-min'' step for positive data, by generating multiple independent estimates $\hat{a}_{cm}$ and taking the minimum as the final estimate. That step can not remove the bias and makes the analysis (such as variance) very difficult. Here we should mention that the bias of CM may not be a major issue in other tasks such as sparse recovery (or ``heavy-hitter'', or ``elephant detection'', by various communities).  \\

\cite{Proc:Weinberger_ICML2009} proposed a creative  method for bias-correction, which consists of pre-multiplying (element-wise) the original data vectors with a random vector whose entries are sampled i.i.d. from the  two-point distribution in $\{-1,1\}$ with equal probabilities, which corresponds to $s =1$ in (\ref{eqn_sparse_r}).

Here, we consider a more general situation by considering any  $s\geq 1$. After applying multiplication and hashing on $u_1$ and $u_2$ as in \cite{Proc:Weinberger_ICML2009}, the resultant vectors $g_1$ and $g_2$ are
{\begin{align}
g_{1,j} = \sum_{i=1}^D u_{1,i} r_i I_{ij}, \hspace{0.5in} g_{2,j} = \sum_{i=1}^D u_{2,i} r_i I_{ij}, \ \ \ j = 1, 2, ..., k
\end{align}}
where $r_i$ is defined as in (\ref{eqn_r_ij}), i.e.,  $E(r_i) = 0, \ E(r_i^2) = 1, \ E(r_i^3) = 0, \ E(r_i^4) = s$. We have the following Lemma.

\begin{lemma}\label{lem_vw}
{\begin{align}\vspace{-0.2in}
&\hat{a}_{vw,s} = \sum_{j=1}^k g_{1,j} g_{2,j},\hspace{0.2in}  E(\hat{a}_{vw,s}) = \sum_{i=1}^D u_{1,i}u_{2,i} =a, \\\label{eqn_var_vw}
&Var(\hat{a}_{vw,s}) = (s-1)\sum_{i=1}^Du_{1,i}^2u_{2,i}^2 + \frac{1}{k}\left[\sum_{i=1}^D u_{1,i}^2\sum_{i=1}^D u_{2,i}^2 + \left(\sum_{i=1}^D u_{1,i}u_{2,i}\right)^2-2\sum_{i=1}^D u_{1,i}^2u_{2,i}^2\right]
\end{align}}
\noindent\textbf{Proof:}\ See Appendix~\ref{proof_lem_vw}.$\Box$
\end{lemma}

Interestingly, the variance (\ref{eqn_var_vw}) says we do need $s=1$, otherwise the additional term $(s-1)\sum_{i=1}^Du_{1,i}^2u_{2,i}^2$ will not vanish even as the sample size $k\rightarrow\infty$. In other words, the choice of random distribution in VW is essentially the only option if we want to remove the bias by pre-multiplying the data vectors (element-wise) with a vector of random variables. Of course, once we let $s=1$, the variance (\ref{eqn_var_vw}) becomes identical to the variance of random projections (\ref{eqn_var_rp}).

\section{Comparing $b$-Bit Minwise Hashing with VW}

\begin{figure}[h!]
\mbox{
\includegraphics[width=1.7in]{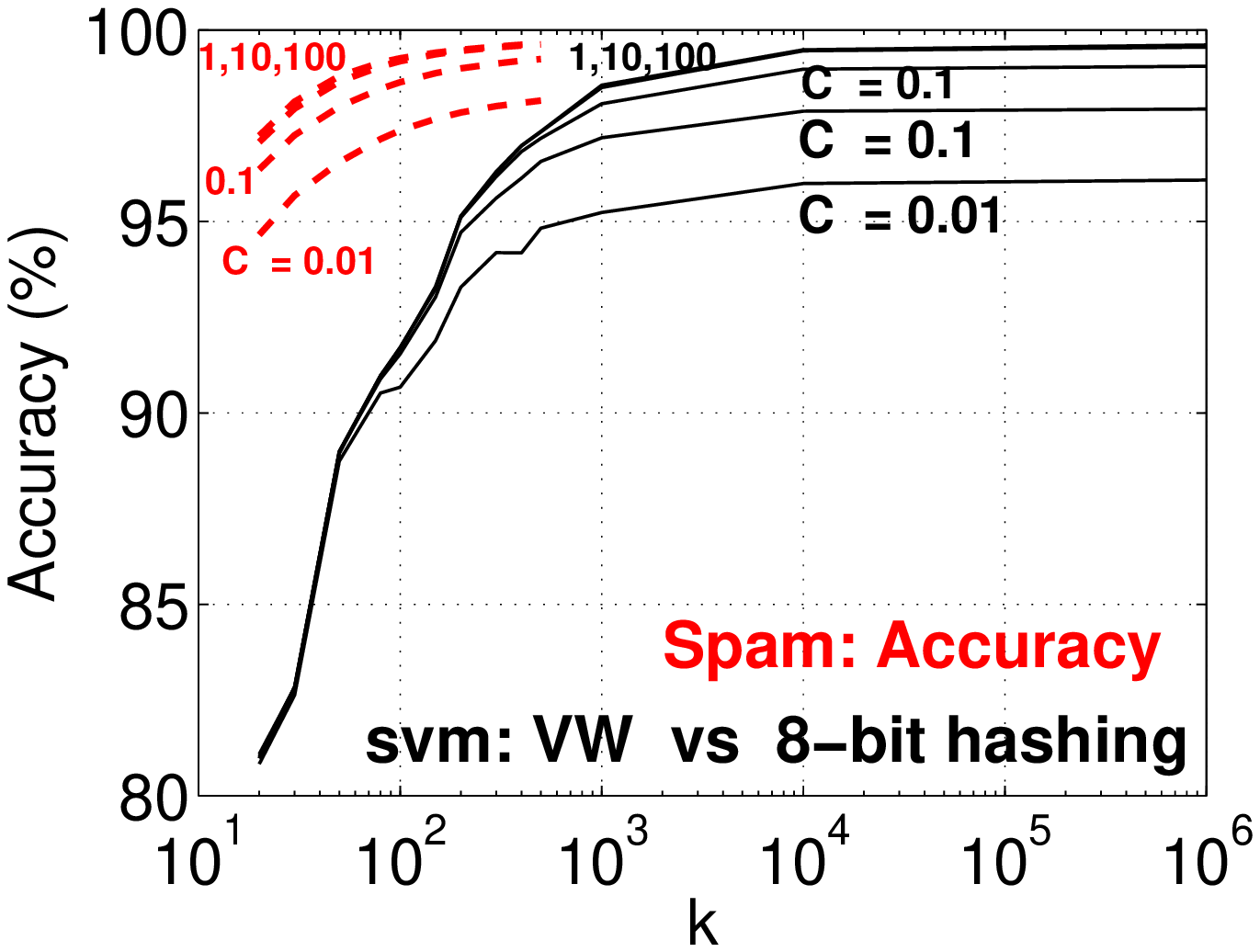}\hspace{-0.1in}
\includegraphics[width=1.7in]{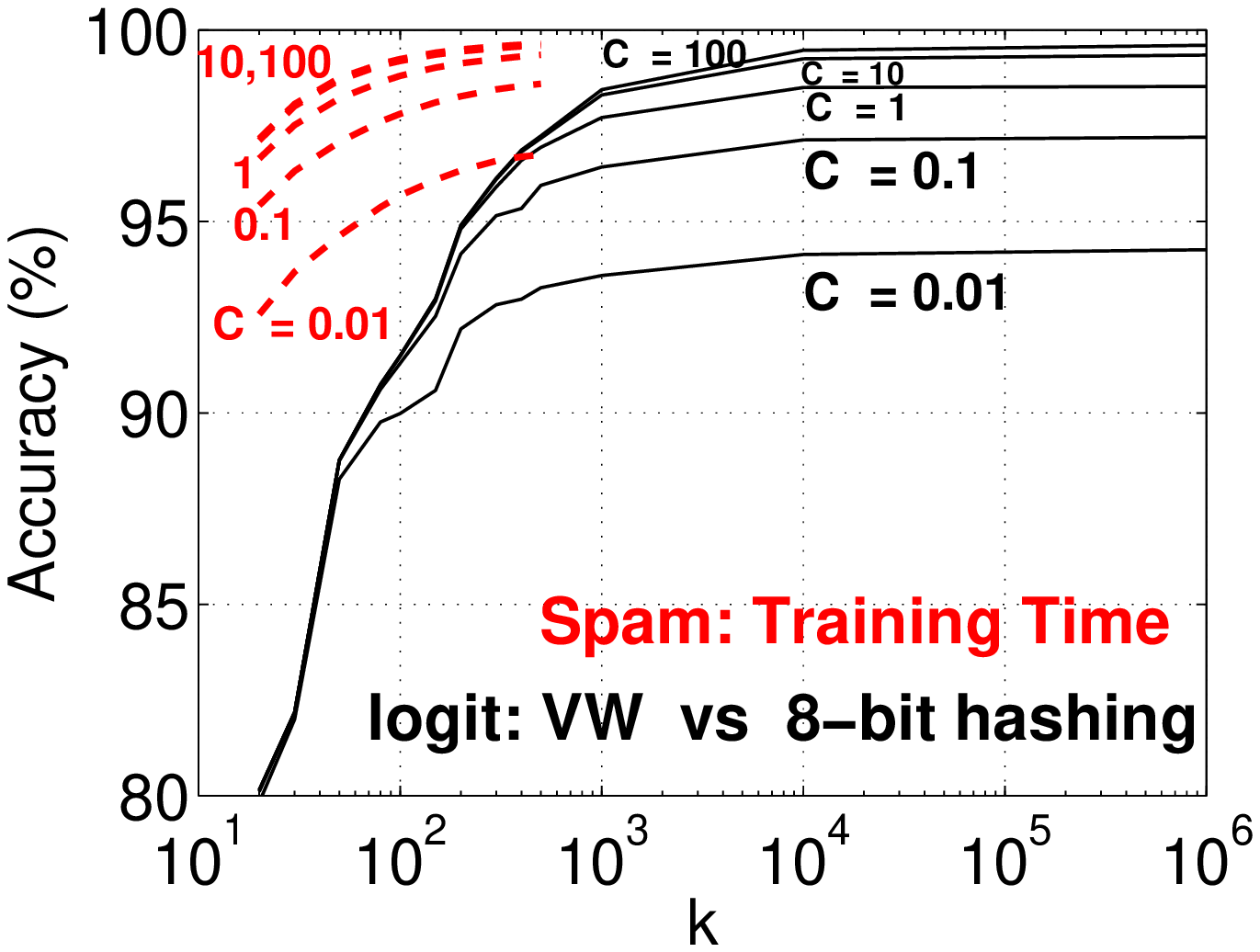}\hspace{-0.1in}
\includegraphics[width=1.7in]{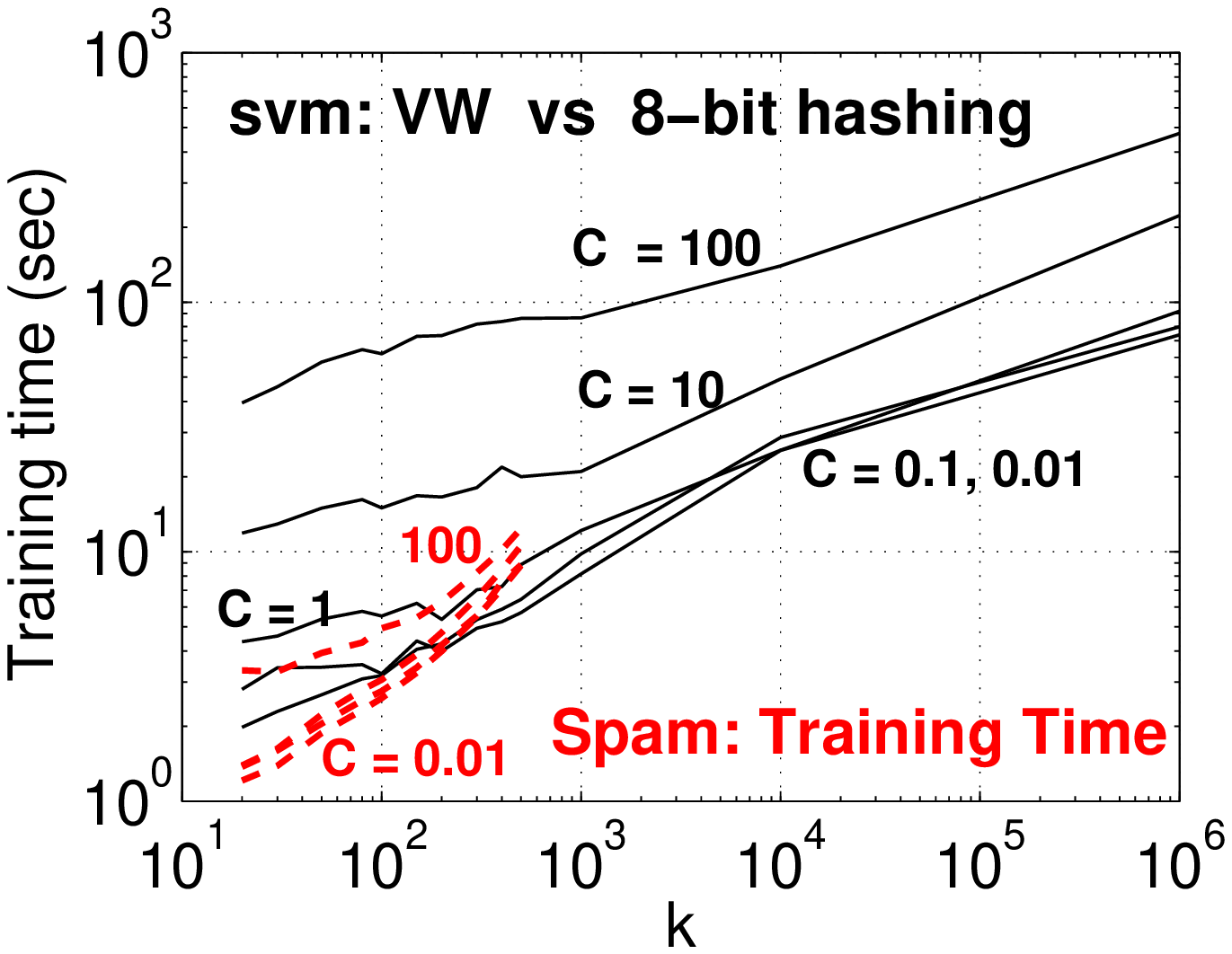}\hspace{-0.1in}
\includegraphics[width=1.7in]{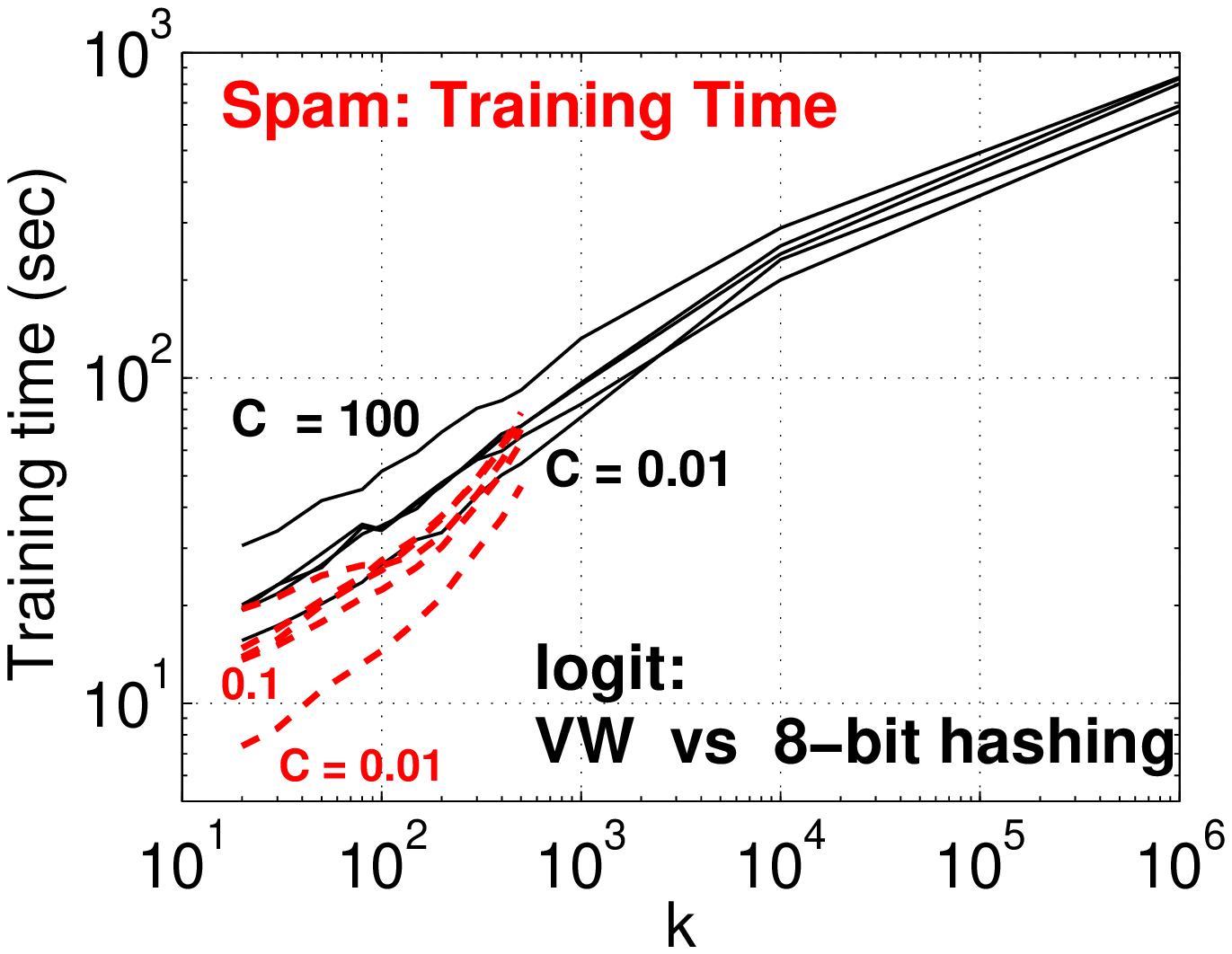}}

\vspace{-0.15in}

\caption{The dashed (red if color is available) curves represent $b$-bit minwise hashing results (only for $k\leq 500$) while solid curves represent VW.  We display results for $C=0.01, 0.1, 1, 10, 100$. }\label{fig_VW}
\end{figure}

We implemented VW (which, in this paper, always refers to the  algorithm developed in~\cite{Proc:Weinberger_ICML2009}) and tested it on the same webspam dataset. Figure~\ref{fig_VW} shows that $b$-bit minwise hashing is substantially more accurate (at the same sample size $k$) and requires significantly less training time (to achieve the same accuracy). For example, $8$-bit minwise hashing with $k=200$ achieves about the same test accuracy as VW with $k=10^6$. Note that we only stored the non-zeros of the hashed data generated by VW.

This empirical finding is not surprising, because the variance of $b$-bit hashing is usually substantially smaller than the variance of VW (and random projections). In Appendix~\ref{app_compare_VW}, we show that, at the same storage cost, $b$-bit hashing usually improves VW by 10- to 100-fold, by assuming each sample of VW requires 32 bits storage. Of course, even if VW only stores each sample using 16 bits, an improvement of 5- to 50-fold would still be very substantial.  Note that this comparison makes sense for the purpose of {\em data reduction}, i.e., the sample size $k$ is substantially smaller than the number of non-zeros in the original (massive) data.\\

There is one interesting issue here. Unlike random projections (and minwise hashing), VW is a {\em sparsity-preserving} algorithm, meaning that in the resultant sample vector of length $k$, the number of non-zeros will not exceed the  number of non-zeros in the original  vector. In fact, it is easy to see that the fraction of zeros in the resultant vector would be (at least) $\left(1-\frac{1}{k}\right)^c\approx \exp\left(-\frac{c}{k}\right)$, where $c$ is the number of non-zeros in the original data vector. When $\frac{c}{k} \geq 5$, then $\exp\left(-\frac{c}{k}\right) \approx 0$. In other words, if our goal is {\em data reduction} (i.e., $k\ll c$), then the hashed data by VW are dense. \\

In this paper, we mainly focus on {\em data reduction}. As discussed in the introduction, for many industry applications, the relatively sparse datasets are often massive in the absolute scale and we assume we can not store all the non-zeros. In fact, this is the also one of the basic motivations for developing minwise hashing.

However, the case of $ c\ll k$ can also be interesting and useful in our work. This is because VW is  an excellent tool for achieving {\em compact indexing} due to the {\em sparsity-preserving} property. Basically, we can let $k$ be very large (like $2^{26}$ in~\cite{Proc:Weinberger_ICML2009}). As the original dictionary size $D$ is extremely large (e.g., $2^{64}$), even $k= 2^{26}$ will be a meaningful reduction of the indexing. Of course, using a very large $k$ will not be useful for the purpose of {\em data reduction}.\\

\section{Combining b-Bit Minwise Hashing with VW}

In our algorithm, we reduce the original massive data to $n b k$ bits only, where $n$ is the number of data points. With (e.g.,) $k=200$ and $b=8$,  our technique achieves a huge {\em data reduction}. In the run-time, we need to expand each data point into a binary vector of length $2^b k$ with exactly $k$ 1's. If $b$ is large like $16$, the new binary vectors will be highly sparse.  In fact, in Figure~\ref{fig_training} and Figure~\ref{fig_training_logit}, we can see that when using $b=16$, the training time becomes substantially larger than using $b\leq 8$ (especially when $k$ is large).

On the other hand, once we have expanded the vectors, the task is merely computing inner products, for which we can actually use VW. Therefore, in the run-time, after we have generated the sparse binary vectors of length $2^bk$, we hash them using VW with sample size $m$ (to differentiate from $k$). How large should $m$ be? Lemma~\ref{lem_b-bit+vw} may provide some  insights. \\

Recall Section~\ref{sec_minwise} provides the estimator, denoted by $\hat{R}_b$, of the resemblance $R$, using $b$-bit minwise hashing. Now, suppose we first apply VW hashing with size $m$ on the binary vector of length $2^bk$ before estimating $R$, which will introduce some additional randomness (on top of $b$-bit hashing). We denote the new estimator by $\hat{R}_{b,vw}$. Lemma~\ref{lem_b-bit+vw} provides its theoretical variance.

\begin{lemma}\label{lem_b-bit+vw}
\begin{align}
&E\left(\hat{R}_{b,vw}\right) = R\\\label{eqn_var_b-bit+vw}
&\text{Var}\left(\hat{R}_{b,vw}\right) = {Var}\left(\hat{R}_b\right) + \frac{1}{m}\frac{1}{\left[1-C_{2,b}\right]^2}\left(1+P_b^2-\frac{P_b(1+P_b)}{k}\right),\\\notag
&\hspace{0.73in} = \frac{1}{k}\frac{P_b(1-P_b)}{\left[1-C_{2,b}\right]^2} + \frac{1}{m}\frac{1+P_b^2}{\left[1-C_{2,b}\right]^2} - \frac{1}{mk}\frac{P_b(1+P_b)}{\left[1-C_{2,b}\right]^2}
\end{align}
where ${Var}\left(\hat{R}_b\right) = \frac{1}{k}\frac{P_b(1-P_b)}{\left[1-C_{2,b}\right]^2}$ is given by (\ref{eqn_Var_b}) and $C_{2,b}$ is the constant defined in Theorem~\ref{The_basic}.\\

\textbf{Proof}:\ The proof is quite straightforward, by following the conditional expectation formula: $E(X) = E(E(X|Y))$, and the conditional variance formula $Var(X) = E(Var(X|Y))+Var(E(X|Y)$.

Recall, originally we estimate the resemblance by $\hat{R}_b = \frac{\hat{P}_b - C_{1,b}}{\left[1-C_{2,b}\right]}$, where $\hat{P}_b = \frac{T}{k}$ and $T$ is the number of matches in the two hashed data vectors of length $k$ generated by $b$-bit hashing. $E(T)=kP_b$ and $Var(T)=kP_b(1-P_b)$. Now, we apply VW (of size $m$ and $s=1$) on the hashed data vectors to estimate $T$ (instead of counting it exactly). We denote this estimates by $\hat{T}$ and $\hat{P}_{b,vw} = \frac{\hat{T}}{k}$.

Because we know the VW estimate is unbiased, we have
\begin{align}\notag
E\left(\hat{P}_{b,vw}\right) = E\left(\frac{\hat{T}}{k}\right) = \frac{E(T)}{k} = \frac{kP_b}{k} = P_b.
\end{align}
Using the conditional variance formula and the variance of VW (\ref{eqn_var_vw}) (with $s=1$), we obtain
\begin{align}\notag
Var\left(\hat{P}_{b,vw}\right) =&\frac{1}{k^2}\left[ E\left(\frac{1}{m}\left[k^2 + T^2 - 2T\right]\right) + Var\left(T\right)\right]\\\notag
=&\frac{1}{k^2}\left[\frac{1}{m}\left(k^2+kP_b(1-P_b)+k^2P_b^2 - 2kP_b\right) + kP_b(1-P_b)\right]\\\notag
=&\frac{P_b(1-P_b)}{k} + \frac{1}{m}\left(1+P_b^2-\frac{P_b(1+P_b)}{k}\right)
\end{align}
This completes the proof. $\Box$\\
\end{lemma}

Compared to the original variance ${Var}\left(\hat{R}_b\right)  = \frac{1}{k}\frac{P_b(1-P_b)}{\left[1-C_{2,b}\right]^2}$, the additional term $\frac{1}{m}\frac{1+P_b^2}{\left[1-C_{2,b}\right]^2}$  in (\ref{eqn_var_b-bit+vw}) can be relatively large if $m$ is not large enough. Therefore, we should choose $m\gg k$ (to reduce the additional variance) and $m\ll 2^bk$ (otherwise there is no need to apply this VW step).  If $b=16$, then $m=2^8k$ may be a good trade-off, because $k\ll 2^8k \ll 2^{16}k$.\\

Figure~\ref{fig_16-bit+vw} provides an empirical study to verify this intuition. Basically, as $m =2^8k$, using VW on top of 16-bit hashing achieves the same accuracies at using 16-bit hashing directly and reduces the training time quite noticeably.

\begin{figure}[h!]

\mbox{
\includegraphics[width=1.7in]{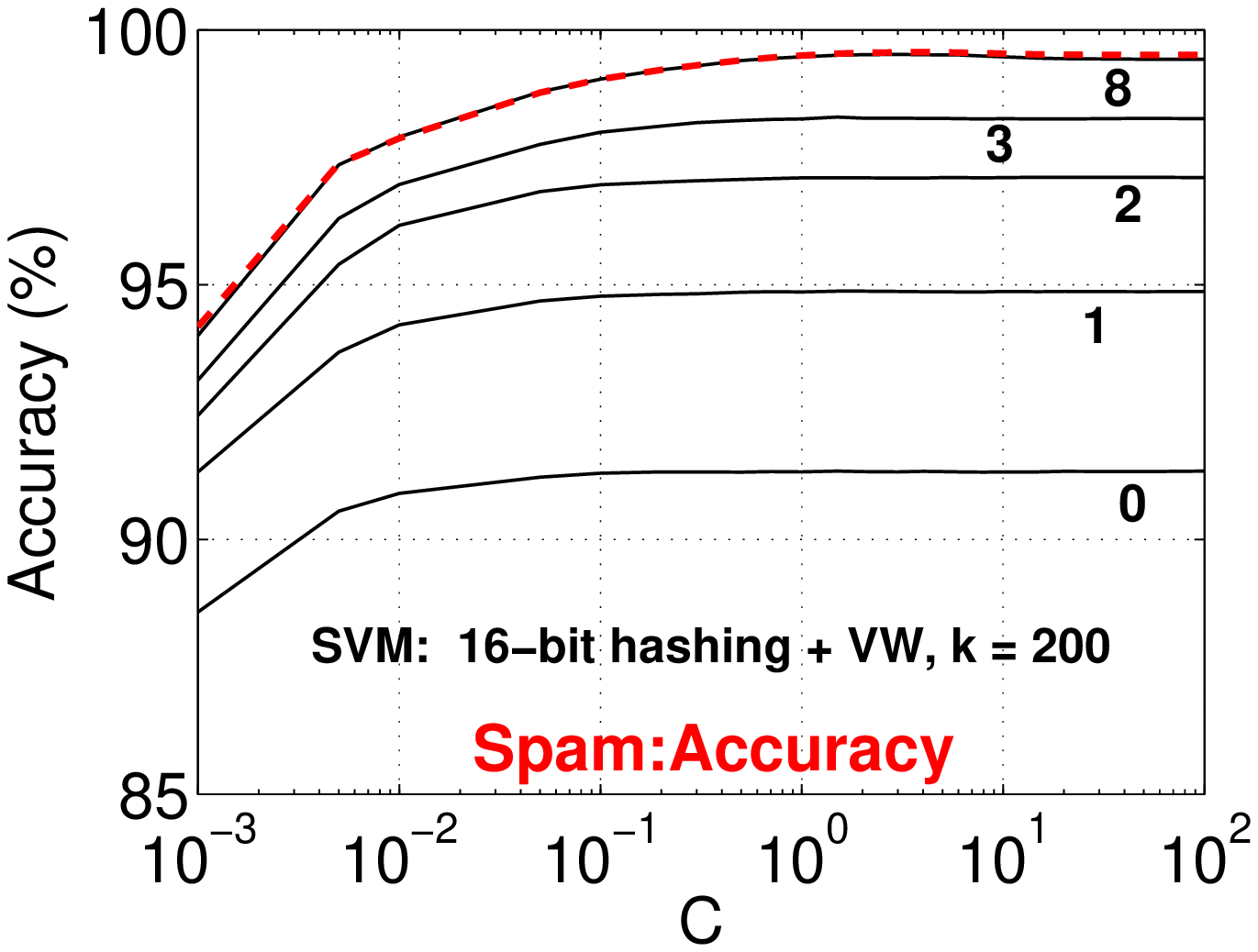}\hspace{-0.1in}
\includegraphics[width=1.7in]{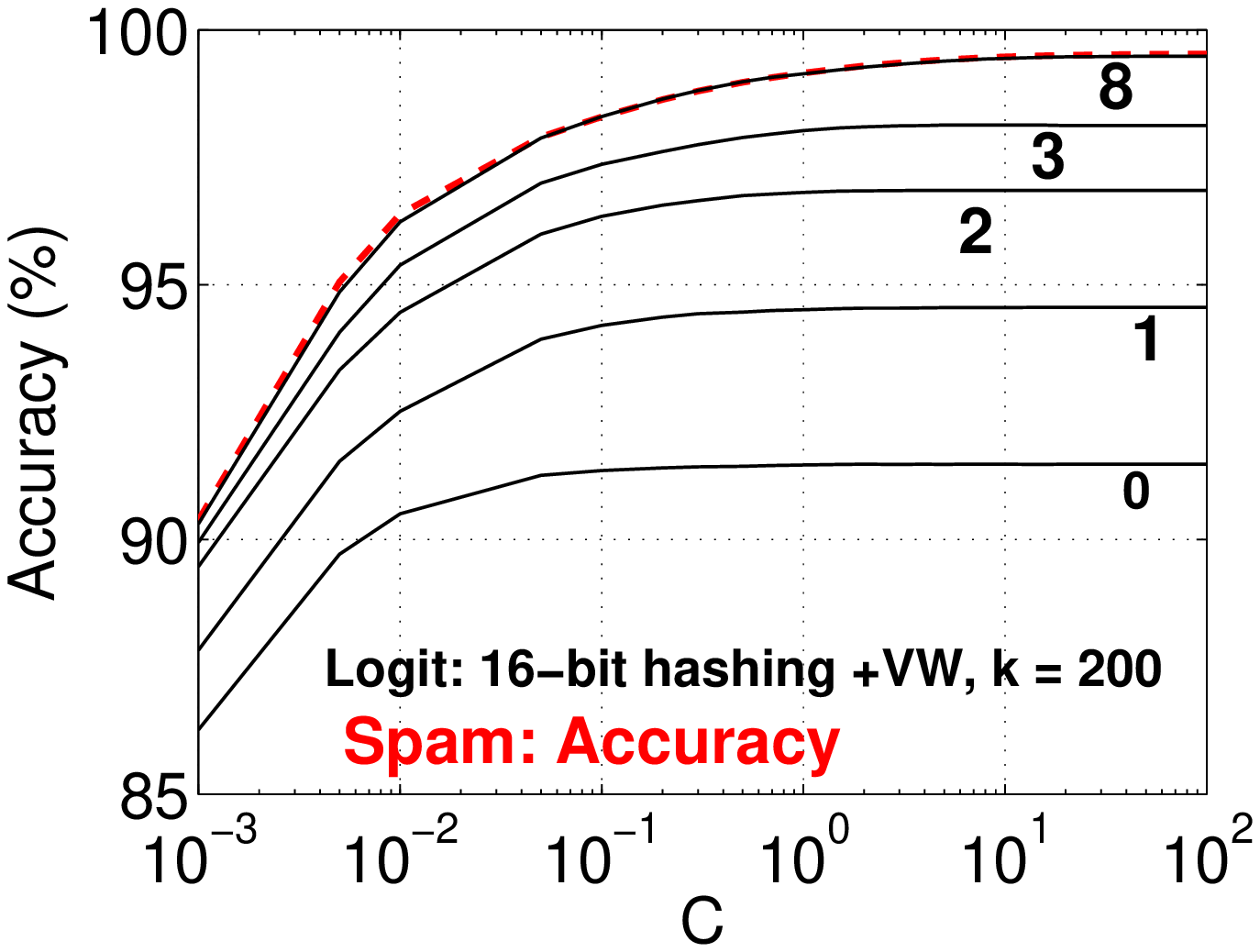}\hspace{-0.1in}
\includegraphics[width=1.7in]{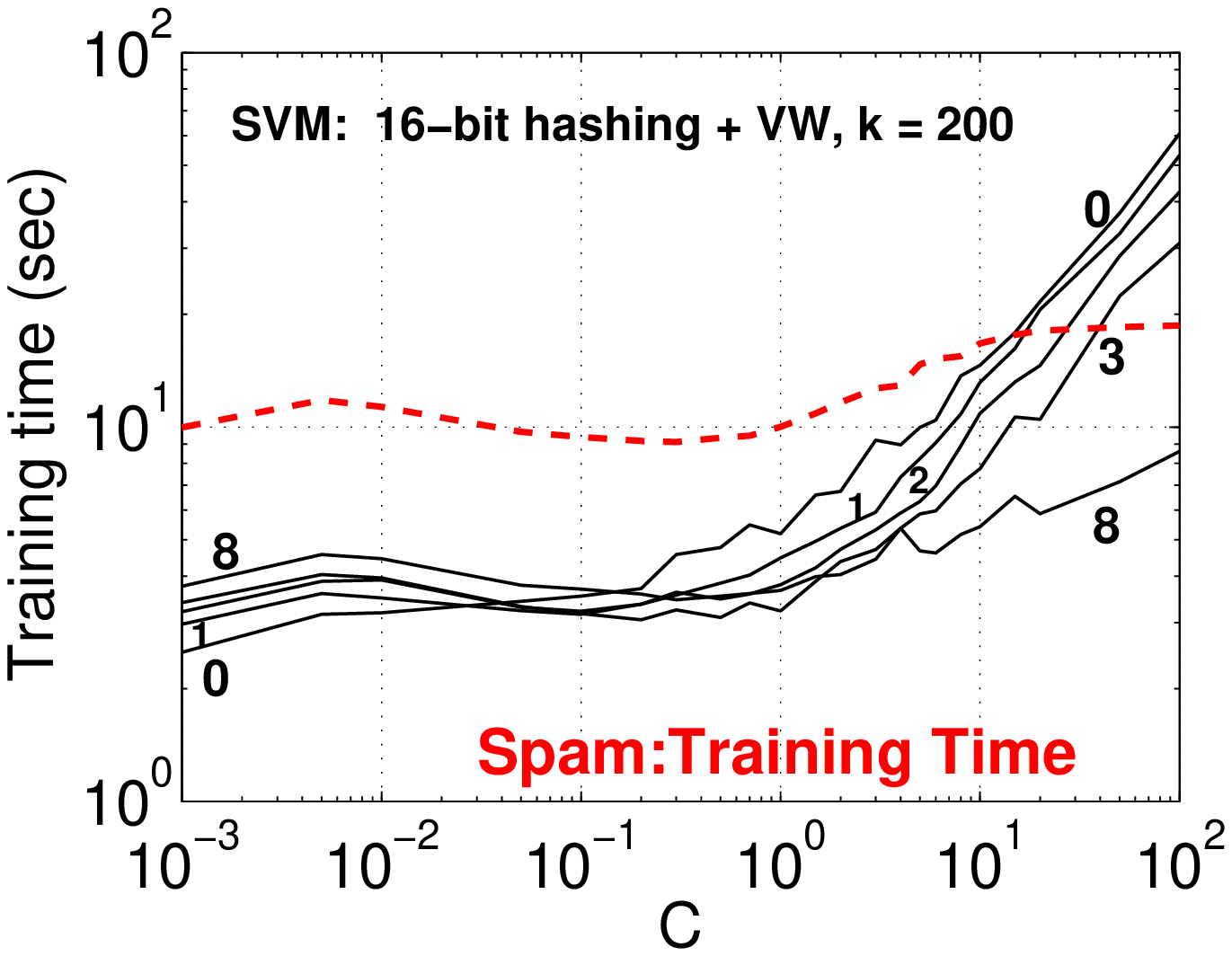}\hspace{-0.1in}
\includegraphics[width=1.7in]{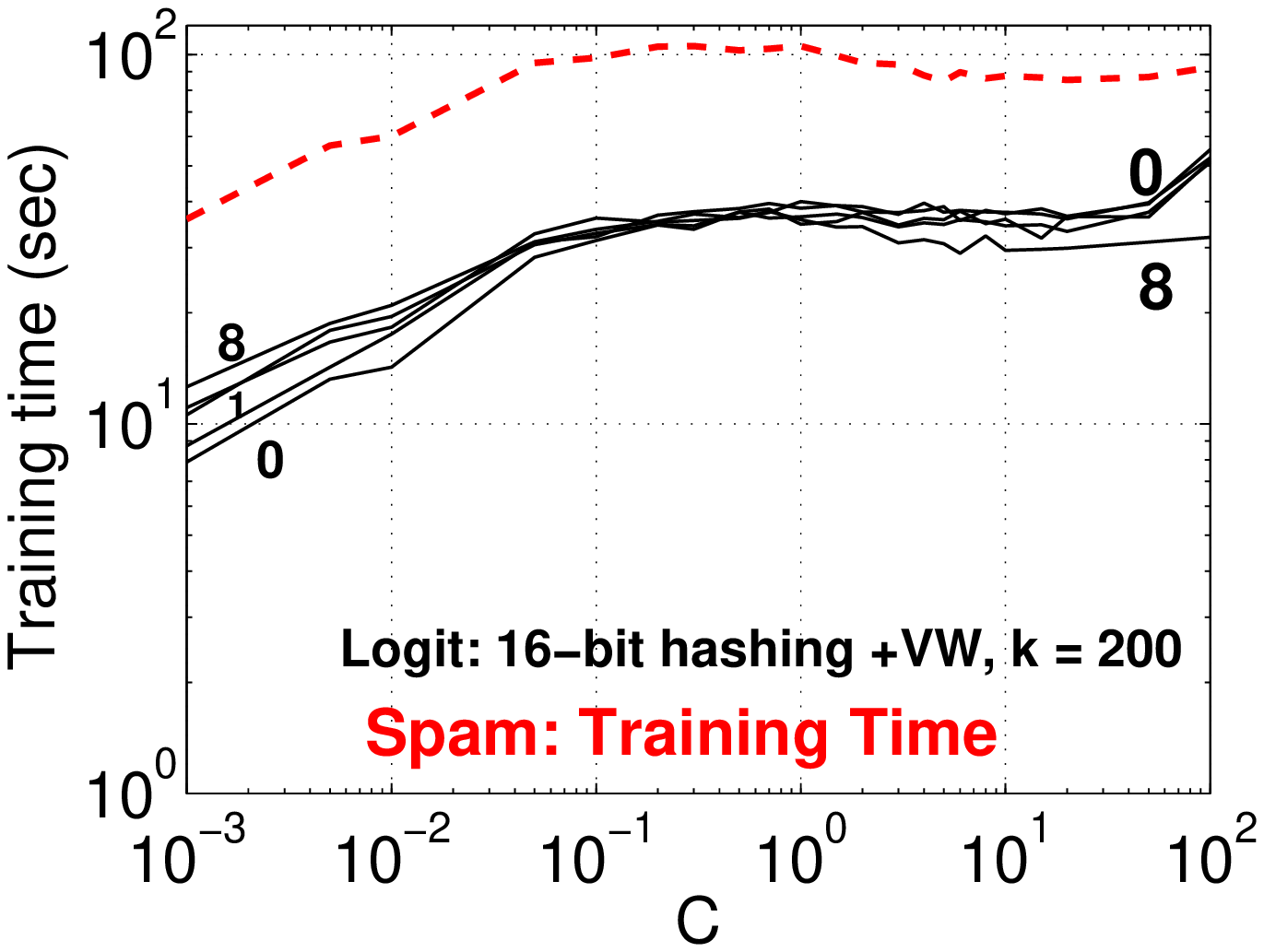}}

\vspace{-0.15in}

\caption{We apply VW hashing on top of the binary vectors (of length $2^bk$) generated by $b$-bit hashing, with size $m = 2^0k, 2^1k, 2^2k, 2^3k, 2^8k$, for $k=200$ and $b=16$. The numbers on the solid curves (0, 1, 2, 3, 8) are the exponents. The dashed (red if color if available) curves are the results from only using $b$-bit hashing. When $m=2^8k$, this method achieves the same test accuracies (left two panels) while considerably reducing the training time (right two panels), if we focus on $C\approx 1$.}
\end{figure}\label{fig_16-bit+vw}

We also experimented with combining 8-bit hashing with VW. We found that we need $m=2^8k$ to achieve similar accuracies, i.e., the additional VW step did not bring more improvement (without hurting accuracies) in terms of  training speed when $b=8$. This is understandable from the analysis of the variance in Lemma~\ref{lem_b-bit+vw}.

\section{Practical Considerations}

Minwise hashing has been widely used in (search) industry and $b$-bit minwise hashing requires only very minimal (if any) modifications (by doing less work). Thus, we expect $b$-bit minwise hashing will be adopted in practice. It is also well-understood in practice that we can use (good) hashing functions to very efficiently simulate permutations.

In many real-world scenarios, the preprocessing step is not critical because it requires only one scan of the data, which can be conducted off-line (or on the data-collection stage, or at the same time as n-grams are generated), and it is trivially parallelizable. In fact, because $b$-bit minwise hashing can substantially reduce the memory consumption, it may be now affordable to store considerably more examples in the memory (after $b$-bit hashing) than before,  to avoid (or minimize) disk IOs. Once the hashed data have been generated, they can be used and re-used for many tasks such as supervised  learning, clustering, duplicate detections, near-neighbor search, etc. For example, a learning task may need to re-use the same (hashed) dataset to perform many cross-validations and parameter tuning (e.g., for experimenting with many $C$ values in SVM).   \\

Nevertheless, there might be situations in which the preprocessing time can be an issue. For example, when a new unprocessed document (i.e. n-grams are not yet available)  arrives and a particular application requires an immediate response from the learning algorithm, then the preprocessing cost might (or might not) be an issue. Firstly, generating n-grams will take some time. Secondly, if during the session a disk IO occurs, then the IO cost will typically mask the cost of preprocessing for $b$-bit minwise hashing.

Note that the preprocessing cost for the VW  algorithm can be substantially lower. Thus, if the time for pre-processing is indeed a concern (while the storage cost or test accuracies are not as much), one may want to consider using VW (or {\em very sparse random projections}~\cite{Proc:Li_Hastie_Church_KDD06}) for those applications.

\section{Conclusion} \label{Conclusion}

As data sizes continue to grow faster than the memory and computational power, machine-learning tasks in industrial practice are increasingly faced with training datasets that exceed the resources on a single server. A number of approaches have been proposed that address this by either scaling out the training process or partitioning the data, but both solutions can be expensive.

In this paper, we propose a compact representation of sparse, binary datasets based on $b$-bit minwise hashing. We show that the $b$-bit minwise hashing estimators are  positive definite kernels and can be naturally integrated with learning algorithms such as SVM and logistic regression, leading to dramatic improvements in training time and/or resource requirements. We also compare $b$-bit minwise hashing with the  Vowpal Wabbit (VW) algorithm, which has the same variances as random projections. Our theoretical and empirical comparisons illustrate that usually $b$-bit minwise hashing is significantly more accurate (at the same storage) than VW for binary data.  Interestingly, $b$-bit minwise hashing can be combined with VW to achieve further improvements in terms of training speed when $b$ is large (e.g., $b\geq 16$).

\newpage

\appendix
\section{Approximation Errors of the Basic Probability Formula}\label{app_basic_error}

\begin{figure}[h!]
\begin{center}
\mbox{
{\includegraphics[width = 1.7  in]{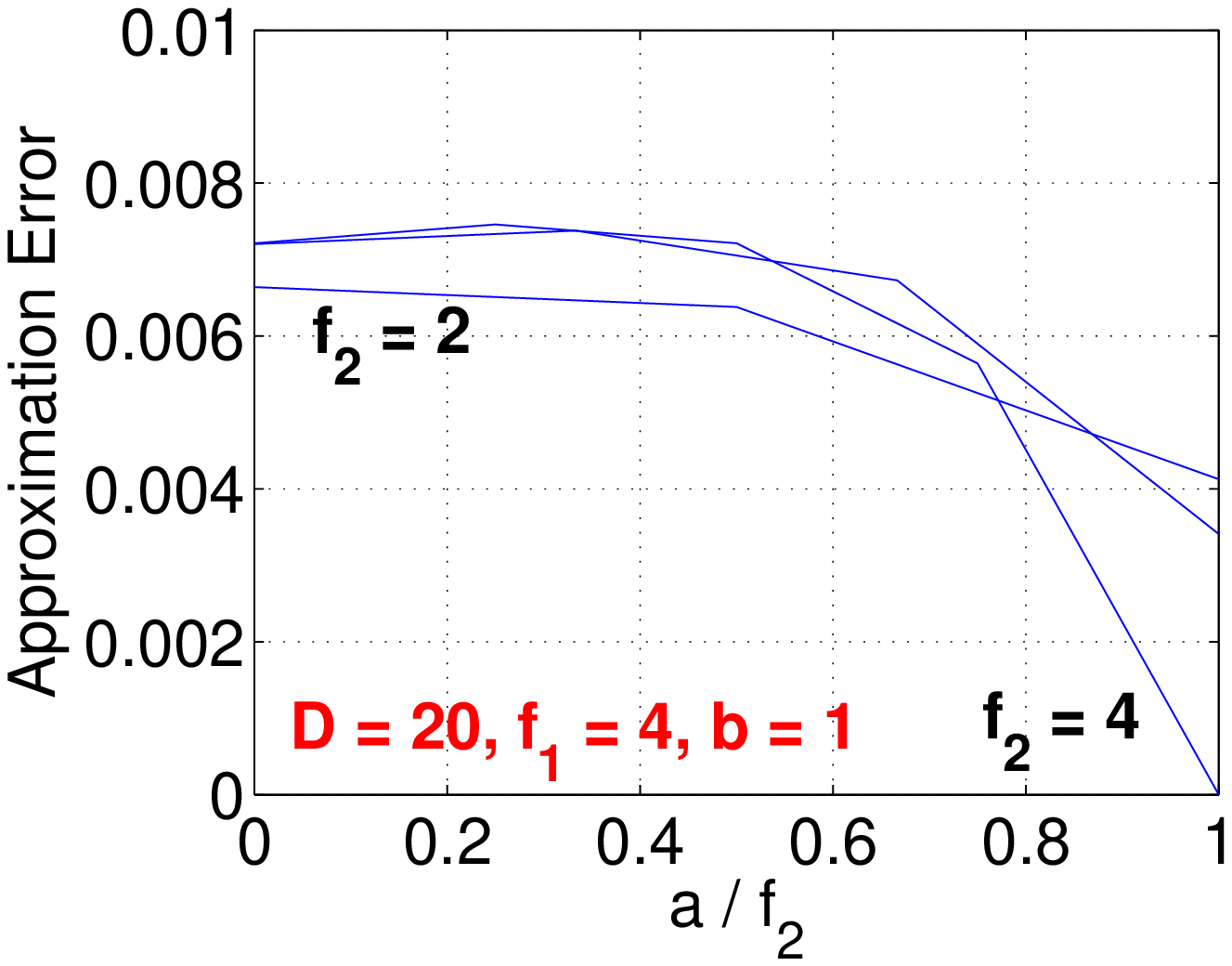}}
{\includegraphics[width = 1.7  in]{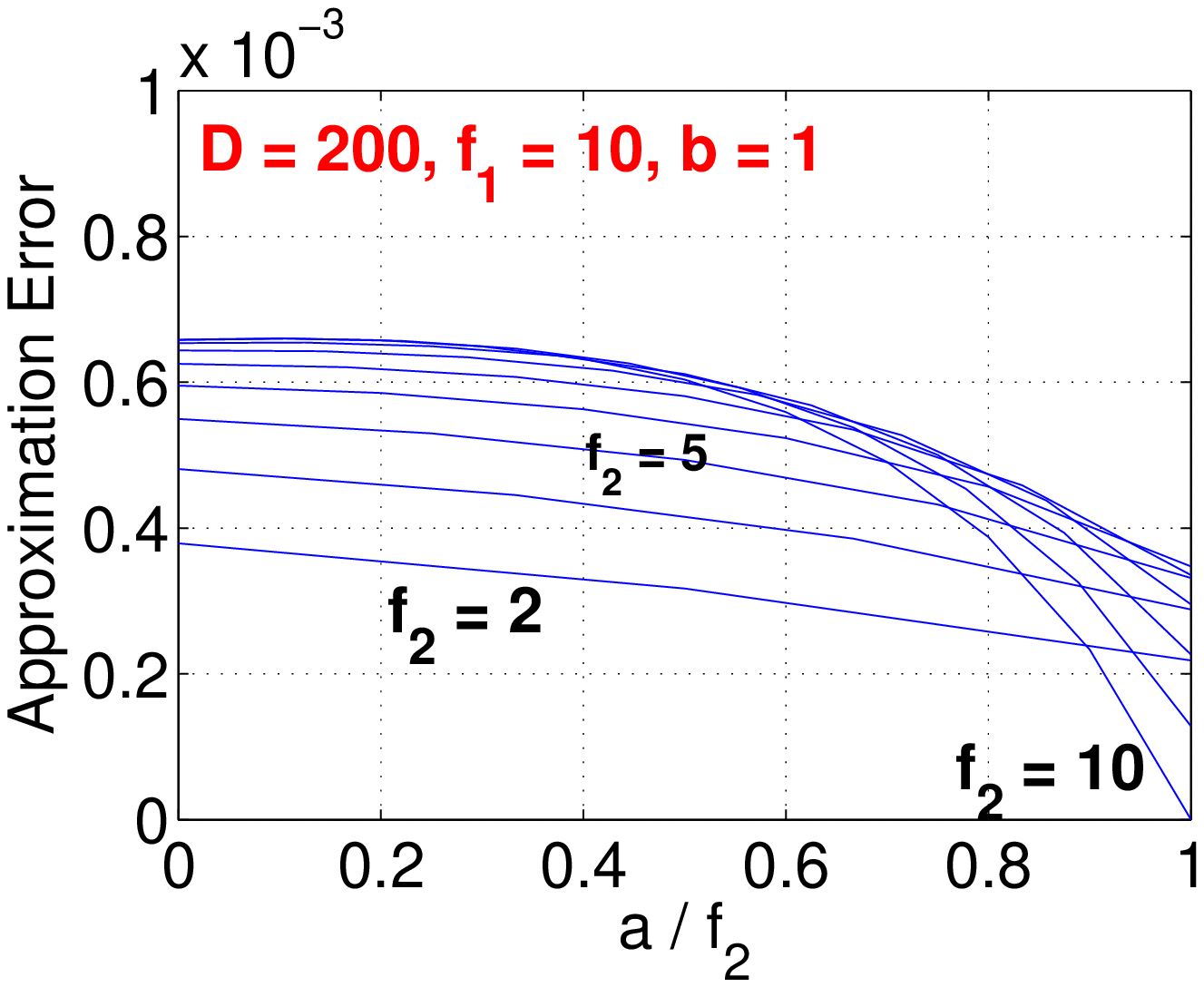}}
{\includegraphics[width = 1.7  in]{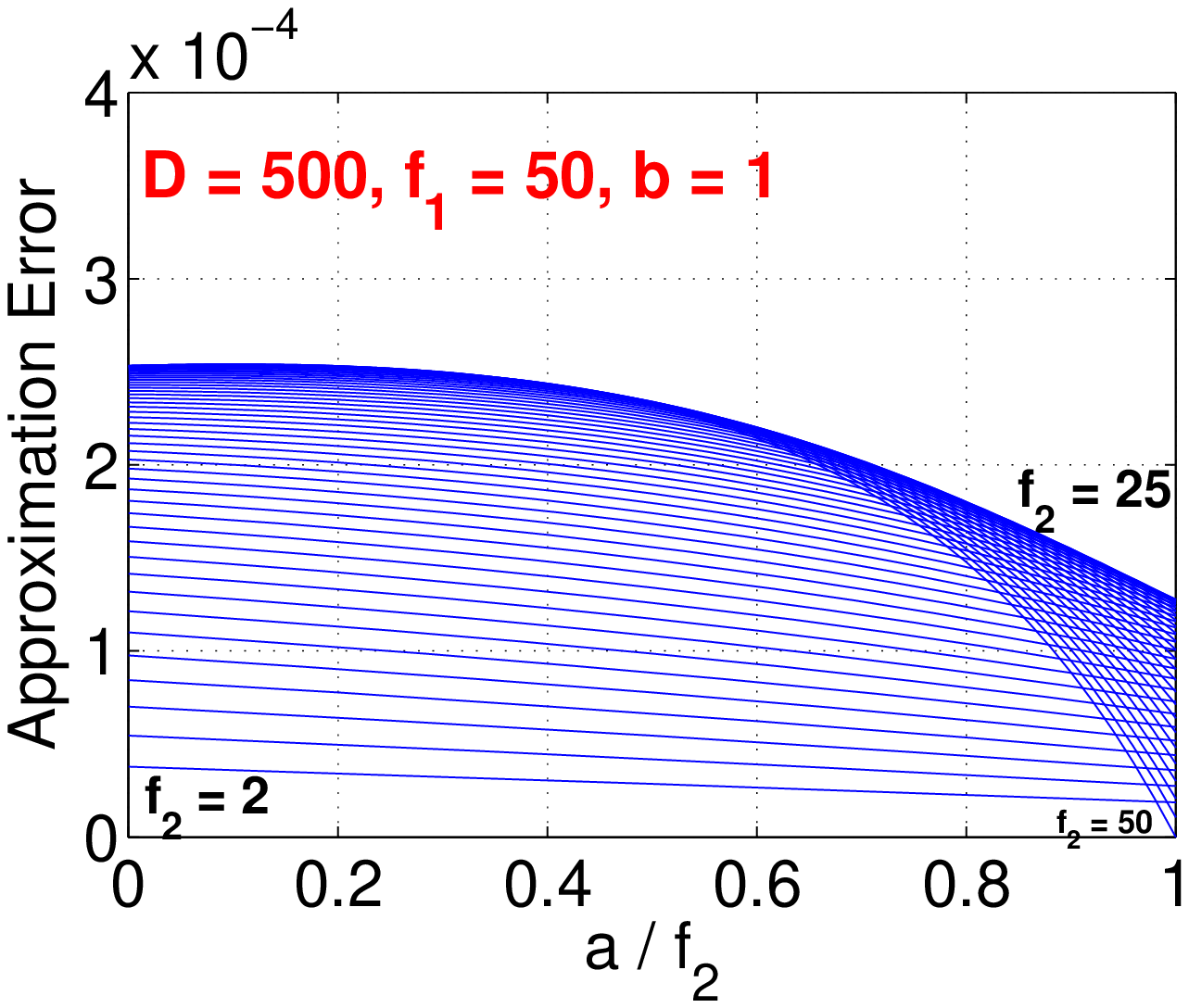}}}

\mbox{
{\includegraphics[width = 1.7  in]{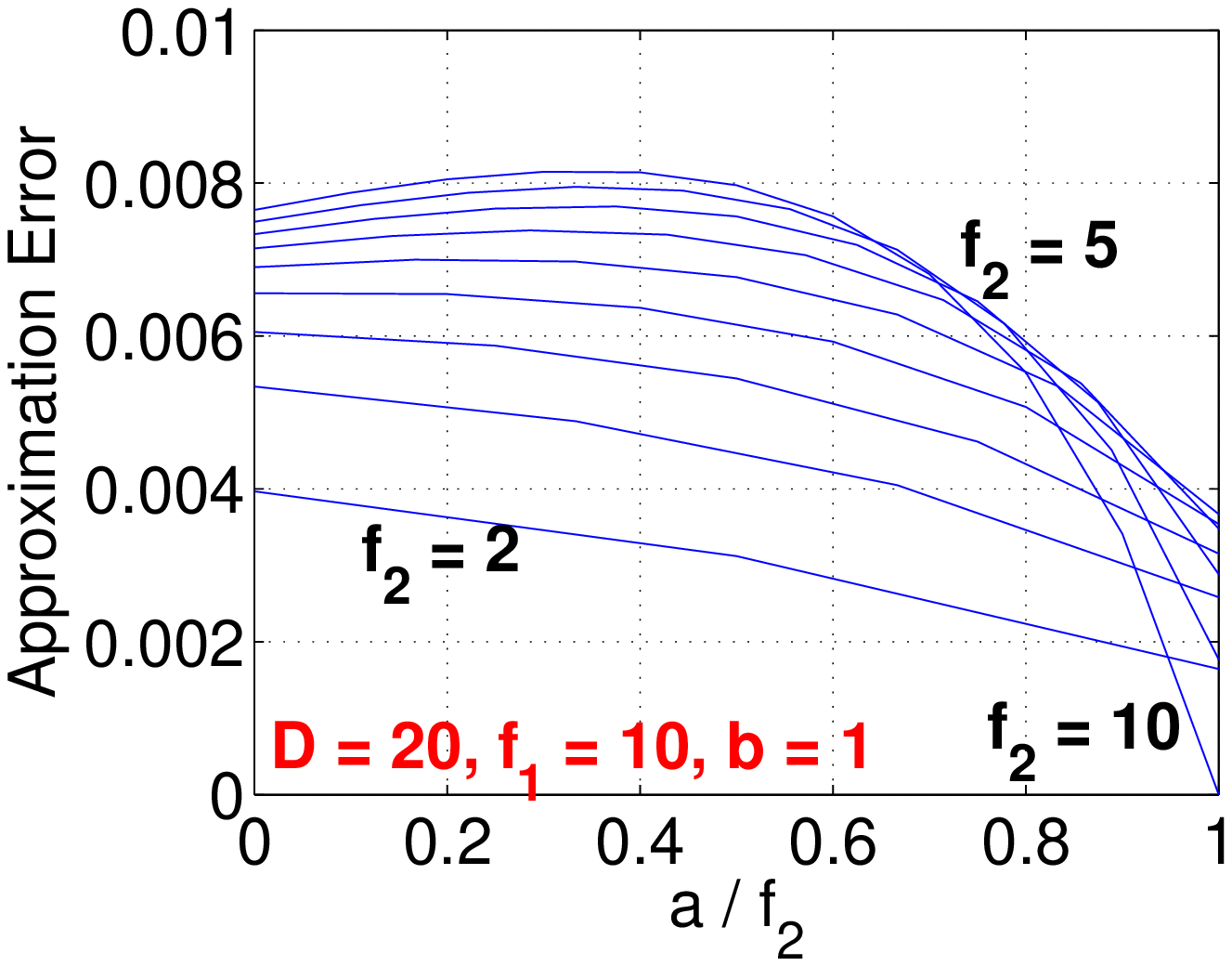}}
{\includegraphics[width = 1.7  in]{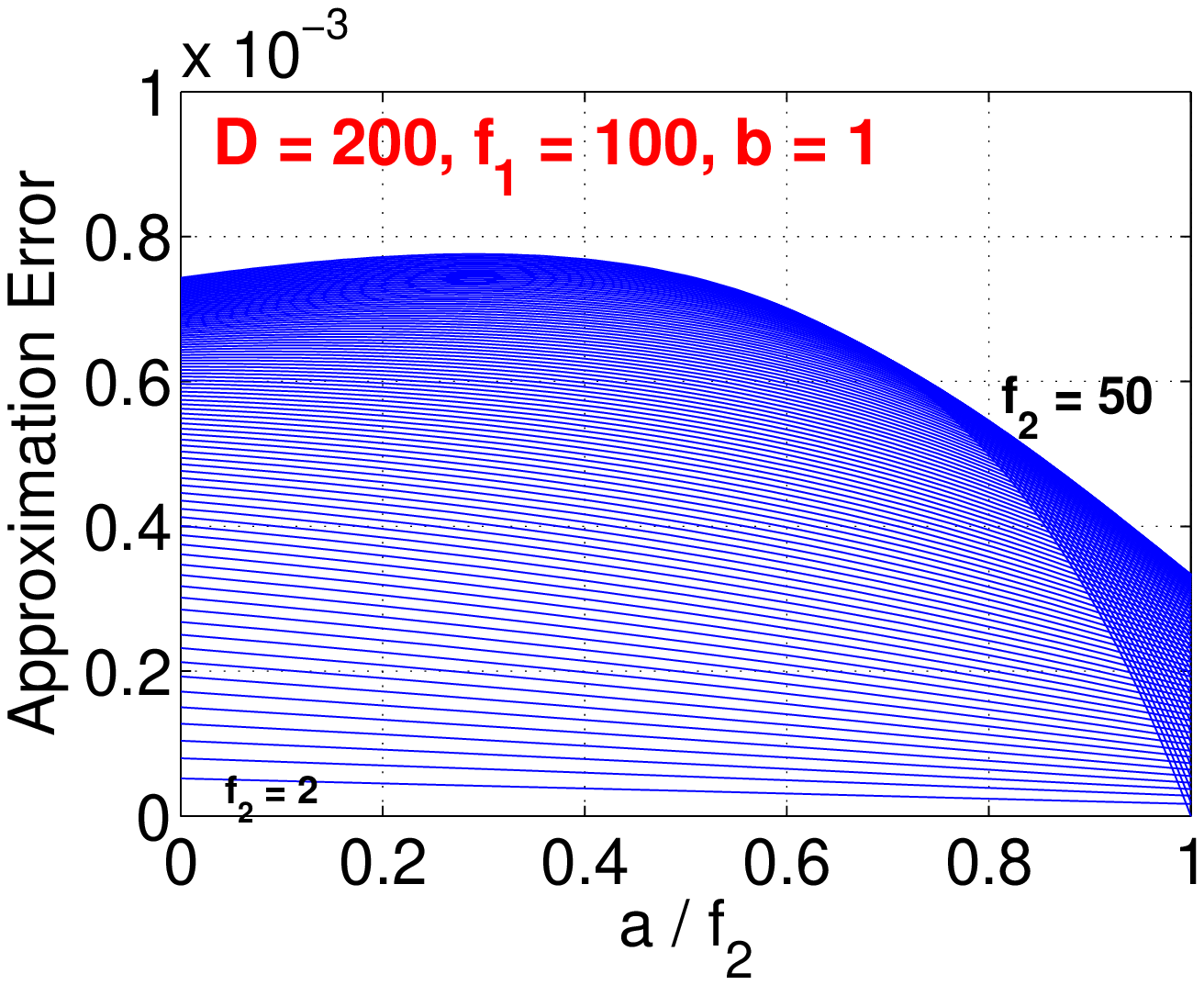}}
{\includegraphics[width = 1.7  in]{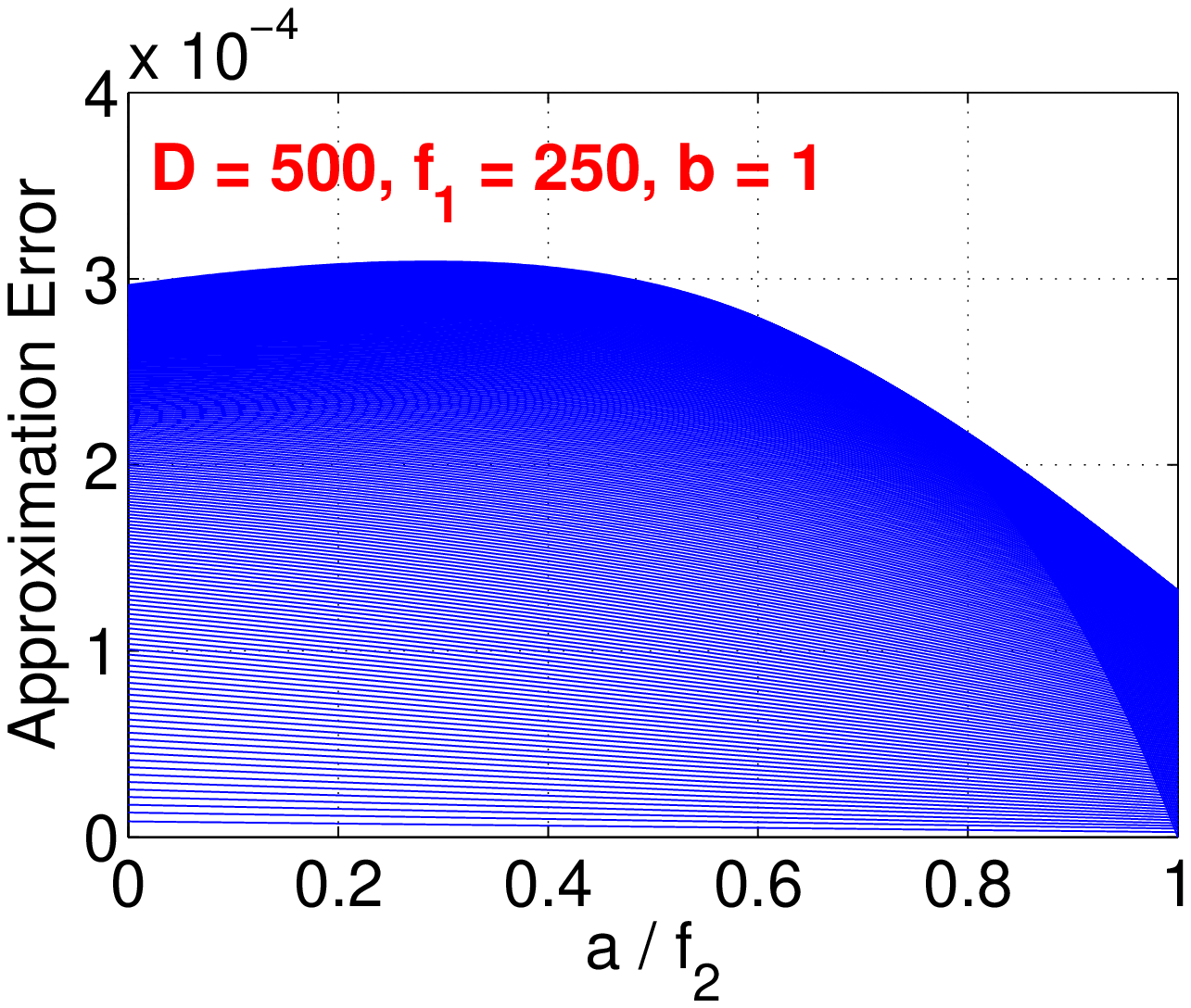}}}

\mbox{
{\includegraphics[width = 1.7  in]{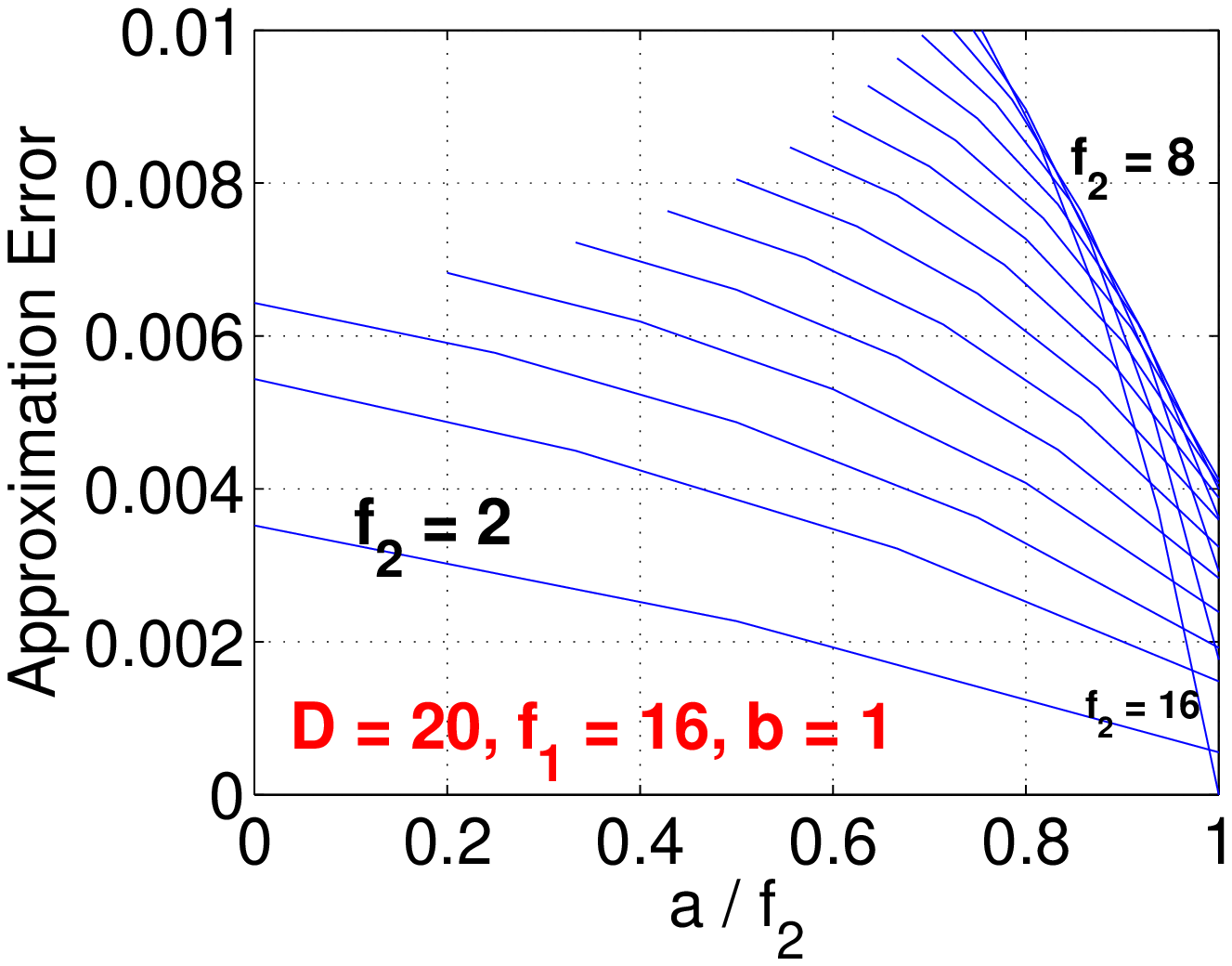}}
{\includegraphics[width = 1.7  in]{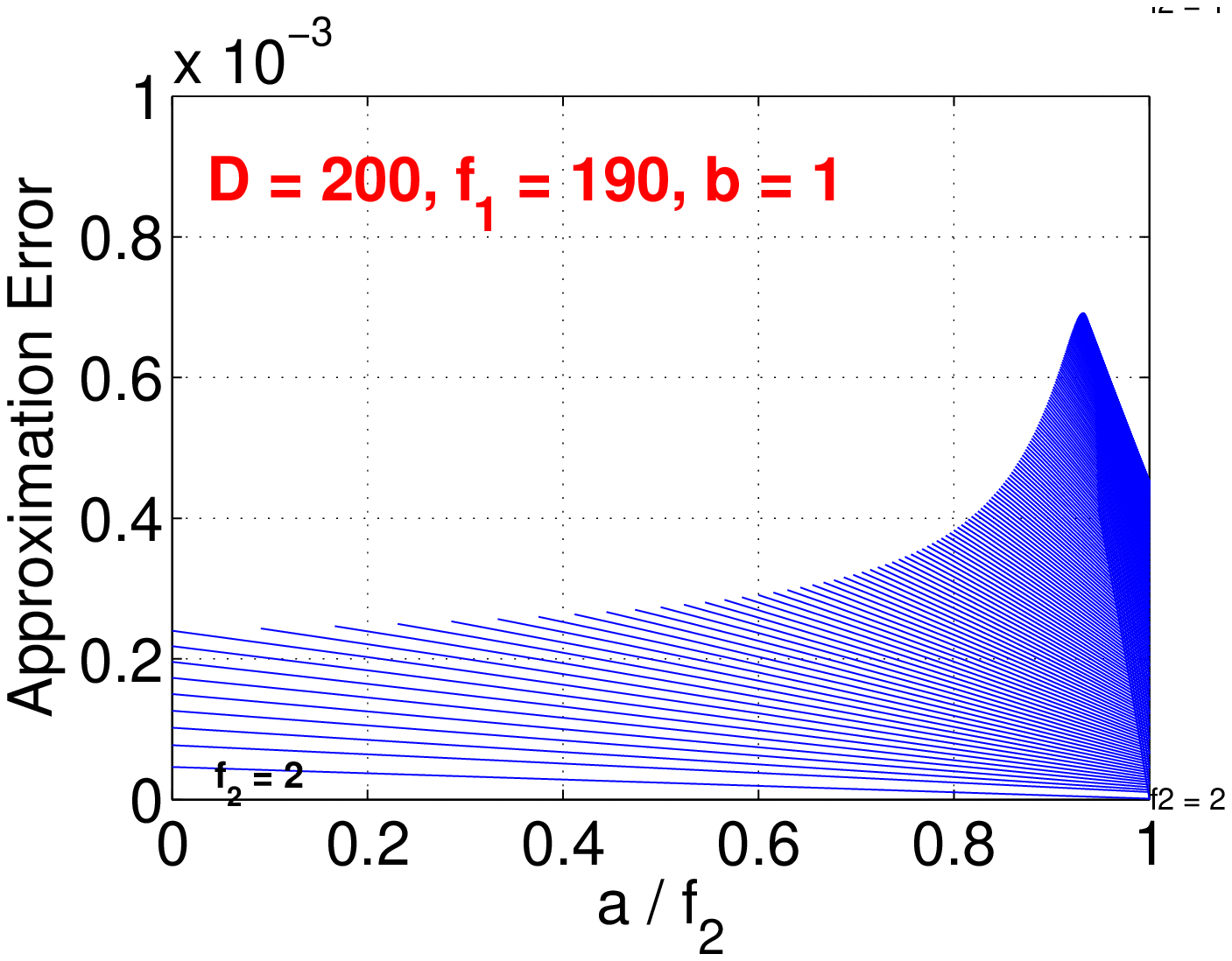}}
{\includegraphics[width = 1.7  in]{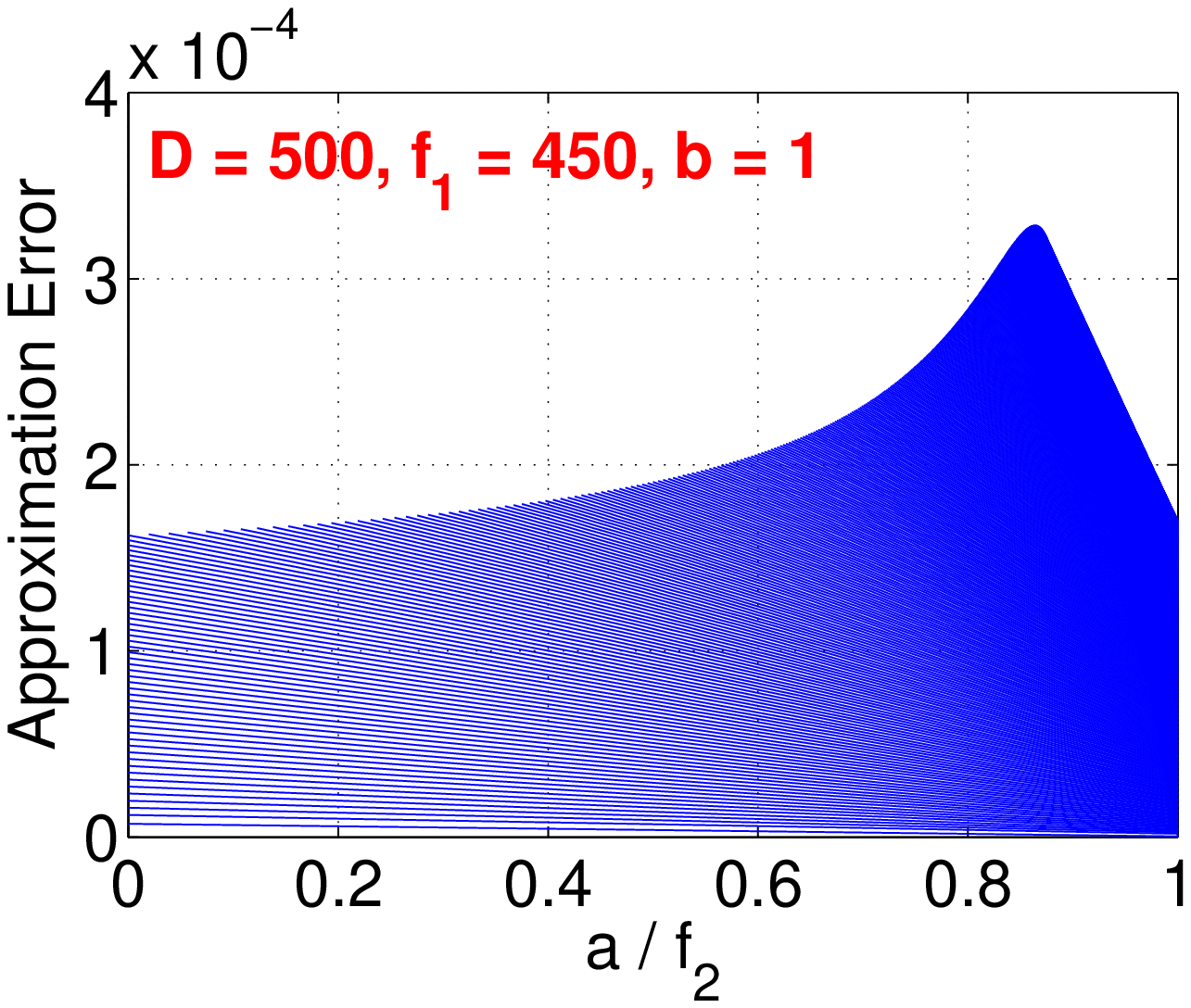}}}

\mbox{
{\includegraphics[width = 1.7  in]{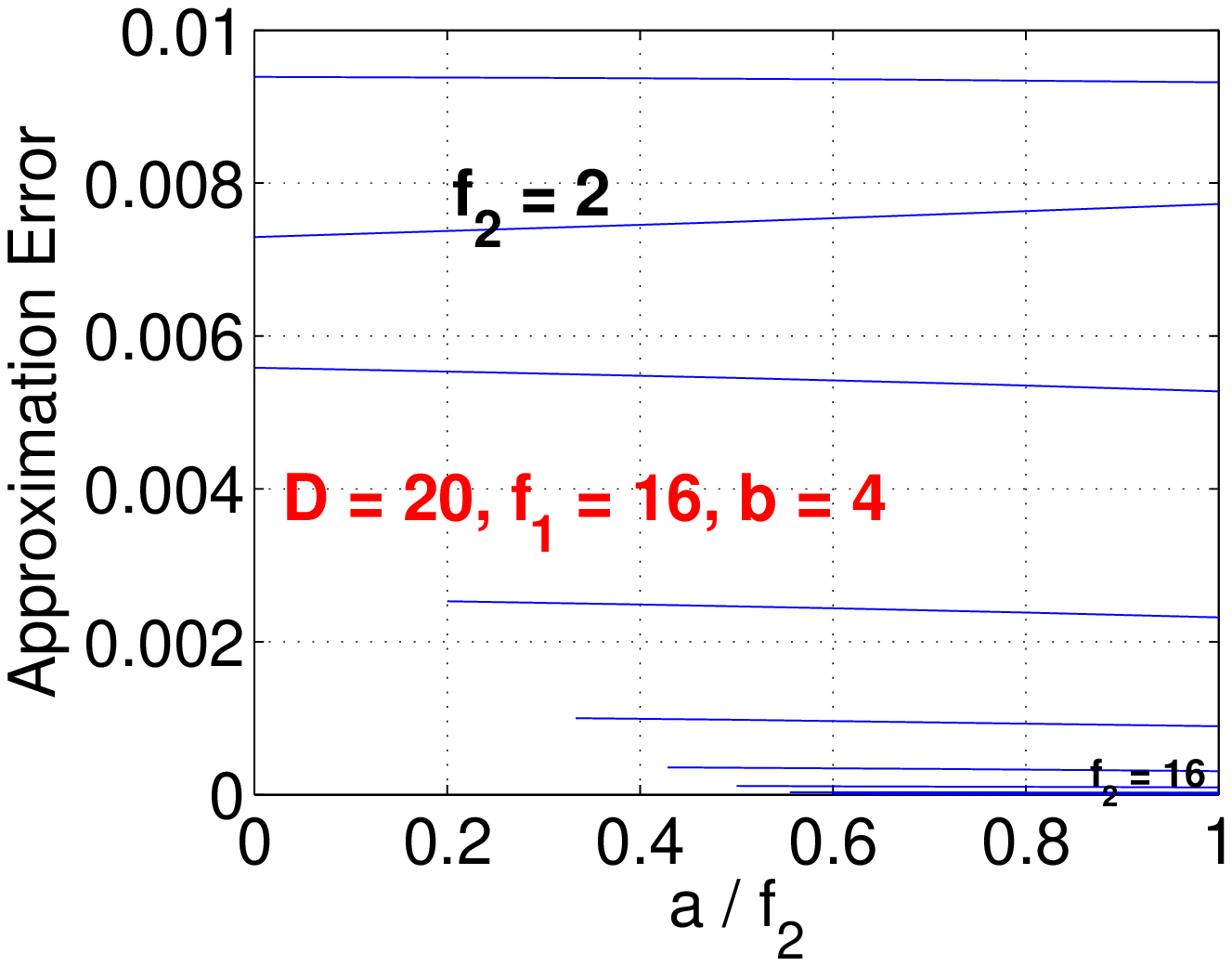}}
{\includegraphics[width = 1.7  in]{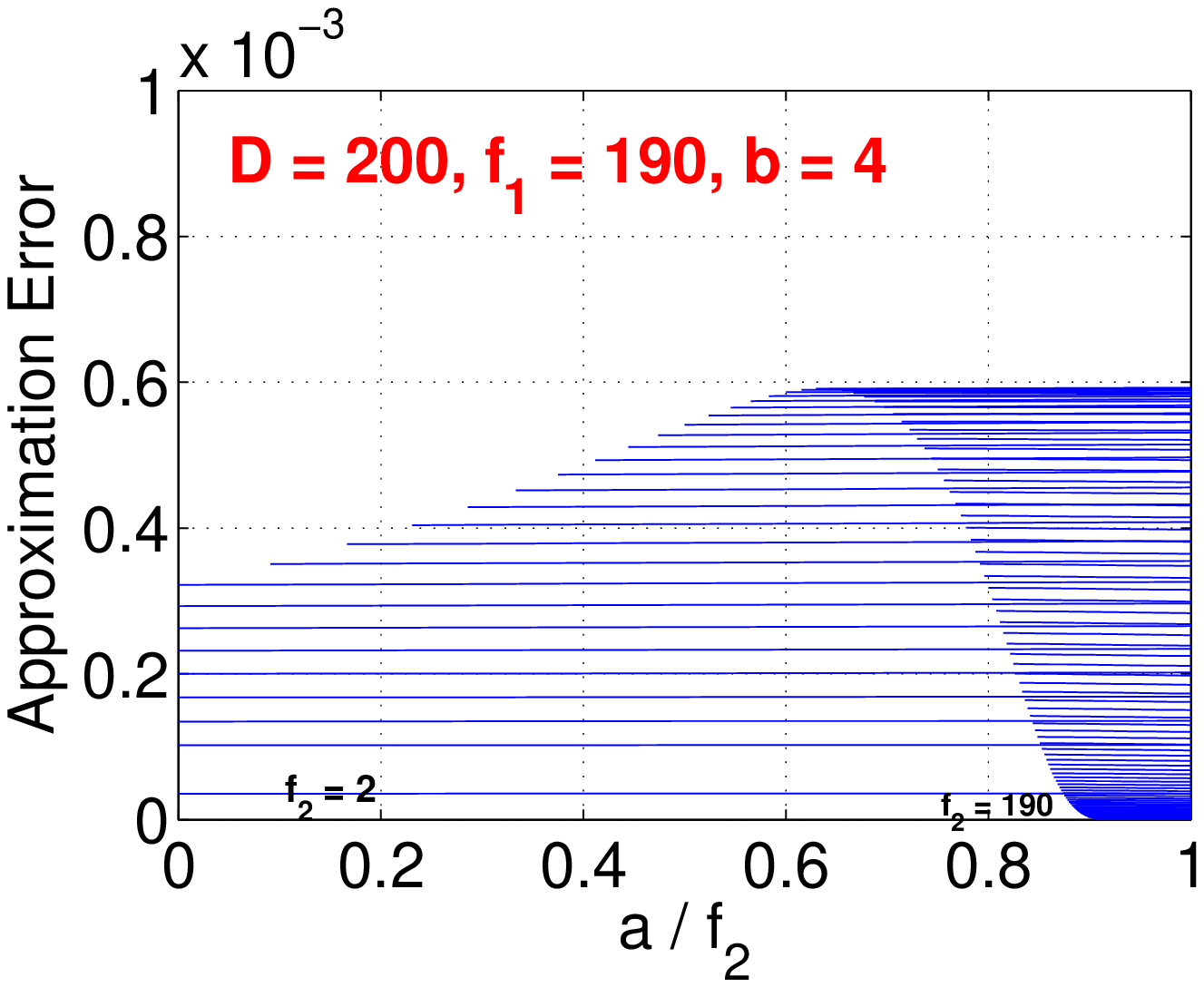}}
{\includegraphics[width = 1.7  in]{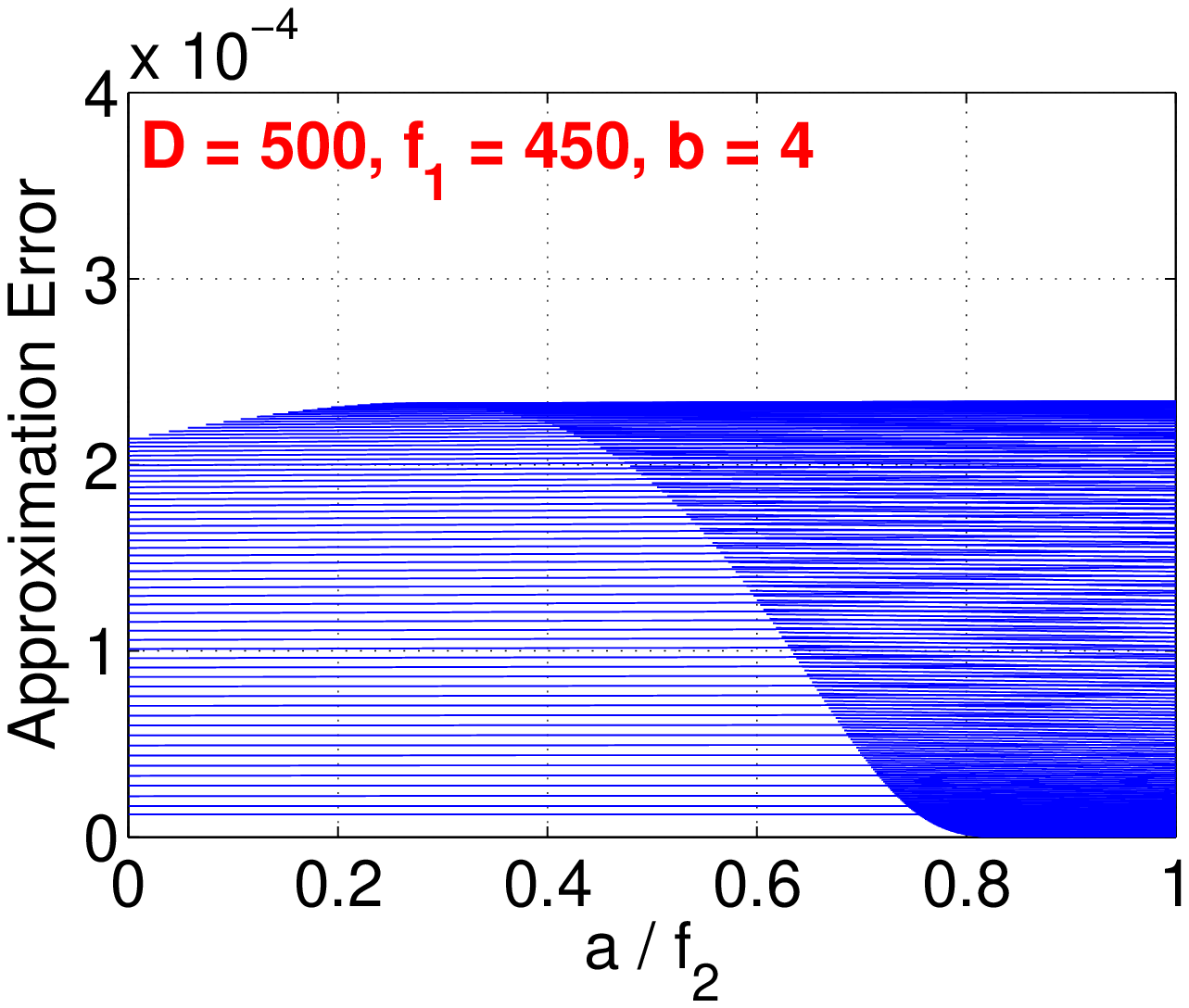}}}
\end{center}
\caption{The absolute errors (approximate - exact) by using (\ref{eqn_basic}) are very small even for $D=20$ (left panels), $D=200$ (middle panels), and $D=500$ (right panels). The exact probability can be numerically computed for small $D$ (from a probability matrix of size $D\times D$). For each $D$, we selected three $f_1$ values. We always let $f_2 = 2, 3, ..., f_1$ and $a = 0, 1, 2, ..., f_2$. }\label{fig_ApproximateError}
\end{figure}

Note that the only assumption needed in the proof of Theorem~\ref{The_basic} is that $D$ is large, which is virtually always satisfied in practice. Interestingly, (\ref{eqn_basic}) is remarkably accurate even for very small $D$. Figure~\ref{fig_ApproximateError} shows that when $D=20/200/500$, the absolute error caused by using (\ref{eqn_basic}) is $<0.01/0.001/0.0004$.  The exact probability, which has no closed-form, can be computed by exhaustive enumerations for small $D$.

\section{Proof of Lemma~\ref{lem_vw}}\label{proof_lem_vw}

The VW algorithm~\cite{Proc:Weinberger_ICML2009} provides a bias-corrected version of the Count-Min (CM) sketch algorithm~\cite{Article:Cormode_05}.

\subsection{The Analysis of the CM Algorithm}

   The key step in CM is to independently and uniformly hash elements of the data vectors to $\{1, 2, 3, ..., k\}$. That is $h(i) = q$ with equal probabilities, where $q \in\{1, 2, ..., k\}$. For convenience, we introduce the following indicator function:
\begin{align}\notag
I_{iq} =\left\{\begin{array}{ll}
1 &\text{if } h(i) = q\\
0 &\text{otherwise}
\end{array}
\right.
\end{align}

Thus, we can denote the CM ``samples''  after the hashing step by
\begin{align}\notag
w_{1,q} = \sum_{i=1}^D u_{1,i}I_{iq}, \hspace{0.5in} w_{2,q} = \sum_{i=1}^D u_{2,i}I_{iq}
\end{align}
and estimate the inner product by $\hat{a}_{cm} = \sum_{q=1}^k w_{1,q} w_{2,q}$, whose expectation and variance can be shown to be
\begin{align}\label{eqn_mean_cm}
&E(\hat{a}_{cm}) = \sum_{i=1}^D u_{1,i}u_{2,i} + \frac{1}{k}\sum_{i\neq j} u_{1,i}u_{2,j}  = a + \frac{1}{k}\sum_{i=1}^Du_{1,i}\sum_{i=1}^Du_{2,i} - \frac{1}{k}a\\\label{eqn_var_cm}
&Var(\hat{a}_{cm}) = \frac{1}{k}\left(1-\frac{1}{k}\right)\left[\sum_{i=1}^D u_{1,i}^2\sum_{i=1}^D u_{2,i}^2 + \left(\sum_{i=1}^D u_{1,i}u_{2,i}\right)^2 - 2\sum_{i=1}^D u_{1,i}^2u_{2,i}^2\right]
\end{align}

From the definition of $I_{iq}$, we can easily infer its moments, for example,
\begin{align}\notag
I_{iq} =\left\{\begin{array}{ll}
1 &\text{if } h(i) = q\\
0 &\text{otherwise}
\end{array}
\right.,\hspace{0.3in} E(I_{iq}^n) = \frac{1}{k},\hspace{0.3in} E(I_{iq}I_{iq^\prime}) = 0\ \text{ if } q\neq q^\prime, \hspace{0.3in}E(I_{iq}I_{i^\prime q^\prime}) = \frac{1}{k^2}\ \text{ if } i\neq i^\prime
\end{align}

The proof of the mean (\ref{eqn_mean_cm}) is simple:
\begin{align}\notag
E(\hat{a}_{cm}) = \sum_{q=1}^k \sum_{i=1}^D u_{1,i}I_{iq} \sum_{i=1}^D u_{2,i}I_{iq} = \sum_{q=1}^k \left[\sum_{i=1}^D u_{1,i}u_{2,i}
E\left(I_{iq}^2\right)+\sum_{i\neq j} u_{1,i}u_{2,j}E\left(I_{iq}I_{jq}\right)\right]= \sum_{i=1}^D u_{1,i}u_{2,i} + \frac{1}{k}\sum_{i\neq j} u_{1,i}u_{2,j}.
\end{align}

The variance (\ref{eqn_var_cm} is more complicated:
\begin{align}\notag
Var(\hat{a}_{cm}) = E(\hat{a}_{cm}^2) - E^2(\hat{a}_{cm})= \sum_{q=1}^k E(w_{1,q}^2w_{2,q}^2)+\sum_{q\neq q^\prime} E(w_{1,q} w_{2,q} w_{1,q^\prime}w_{2,q^\prime}) -
 \left(\sum_{i=1}^D u_{1,i}u_{2,i} + \frac{1}{k}\sum_{i\neq j} u_{1,i}u_{2,j}\right)^2
\end{align}
The following expansions are helpful:
\begin{align}\notag
\left[\sum_{i=1}^D a_i \sum_{i=1}^D b_i\right]^2 =& \sum_{i=1}^D a_i^2b_i^2
+ \sum_{i\neq j} a_i^2b_j^2+2a_i^2b_ib_j+2b_i^2a_ia_j+2a_ib_ia_jb_j\\\notag
+&\sum_{i\neq j\neq c} a_i^2b_jb_c+b_i^2a_ja_c + 4a_ib_ia_jb_c+\sum_{i\neq j\neq c\neq t}a_ib_ja_cb_t\\\notag
\left[\sum_{i\neq j} a_i b_j\right]^2 =&  \sum_{i\neq j} a_i^2b_j^2+a_ib_ia_jb_j
+\sum_{i\neq j\neq c} a_i^2b_jb_c+b_i^2a_ja_c + 2a_ib_ia_jb_c+\sum_{i\neq j\neq c\neq t}a_ib_ja_cb_t\\\notag
\sum_{i=1}^Da_ib_i\sum_{i\neq j} a_i b_j =& \sum_{i\neq j} a_i^2b_ib_j+b_i^2a_ia_j+\sum_{i\neq j\neq c} a_i^2b_jb_c
\end{align}

which, combined with the moments of $I_{iq}$, yield
\begin{align}\notag
\sum_{q=1}^k E(w_{1,q}^2w_{2,q}^2) =& \sum_{i=1}^D u_{1,i}^2u_{2,i}^2
+ \frac{1}{k}\sum_{i\neq j} u_{1,i}^2u_{2,j}^2+2u_{1,i}^2u_{2,i}u_{2,j}+2u_{2,i}^2u_{1,i}u_{1,j}+2u_{1,i}u_{2,i}u_{1,j}u_{2,j}\\\notag
+&\frac{1}{k^2}\sum_{i\neq j\neq c} u_{1,i}^2u_{2,j}u_{2,c}+u_{2,i}^2u_{1,j}u_{1,c} + 4u_{1,i}u_{2,i}u_{1,j}u_{2,c}+\frac{1}{k^3}\sum_{i\neq j\neq c\neq t}u_{1,i}u_{2,j}u_{1,c}u_{2,t}
\end{align}
\begin{align}\notag
&\sum_{q\neq q^\prime} E(w_{1,q} w_{2,q} w_{1,q^\prime}w_{2,q^\prime})\\\notag
=& k(k-1)\left[\frac{1}{k^2}\sum_{i\neq j} u_{1,i}u_{2,i}u_{1,j}u_{2,j}+\frac{2}{k^3}\sum_{i\neq j \neq c} u_{1,i}u_{2,i}u_{1,j}u_{2,c}+\frac{1}{k^4}\sum_{i\neq j\neq c\neq t} u_{1,i}u_{2,j}u_{1,c}u_{2,t}\right]\\\notag
=& (k-1)\left[\frac{1}{k}\sum_{i\neq j} u_{1,i}u_{2,i}u_{1,j}u_{2,j}+\frac{2}{k^2}\sum_{i\neq j \neq c} u_{1,i}u_{2,i}u_{1,j}u_{2,c}+\frac{1}{k^3}\sum_{i\neq j\neq c\neq t} u_{1,i}u_{2,j}u_{1,c}u_{2,t}\right]
\end{align}

\begin{align}\notag
&E(\hat{a}_{cm}^2) = \sum_{q=1}^k E(w_{1,q}^2w_{2,q}^2) +\sum_{q\neq q^\prime} E(w_{1,q} w_{2,q} w_{1,q^\prime}w_{2,q^\prime})\\\notag
=& \sum_{i=1}^D u_{1,i}^2u_{2,i}^2 + \sum_{i\neq j} u_{1,i}u_{2,i}u_{1,j}u_{2,j}
+ \frac{1}{k}\sum_{i\neq j} u_{1,i}^2u_{2,j}^2+2u_{1,i}^2u_{2,i}u_{2,j}+2u_{2,i}^2u_{1,i}u_{1,j}+u_{1,i}u_{2,i}u_{1,j}u_{2,j}+\frac{2}{k}\sum_{i\neq j \neq c} u_{1,i}u_{2,i}u_{1,j}u_{2,c}\\\notag
+&\frac{1}{k^2}\sum_{i\neq j\neq c} u_{1,i}^2u_{2,j}u_{2,c}+u_{2,i}^2u_{1,j}u_{1,c} + 2u_{1,i}u_{2,i}u_{1,j}u_{2,c}+\frac{1}{k^2}\sum_{i\neq j\neq c\neq t}u_{1,i}u_{2,j}u_{1,c}u_{2,t}\\\notag
=&\left(\sum_{i=1}^Du_{1,i}u_{2,i}\right)^2+\left(\frac{1}{k}-\frac{1}{k^2}\right)\sum_{i\neq j} u_{1,i}^2u_{2,j}^2+u_{1,i}u_{2,i}u_{1,j}u_{2,j} + \frac{2}{k}\sum_{i=1}^Du_{1,i}u_{2,i}\sum_{i\neq j}u_{1,i}u_{2,j}+\frac{1}{k^2}\left(\sum_{i\neq j}u_{1,i}u_{2,j}\right)^2
\end{align}
Therefore,

\begin{align}\notag
&Var(\hat{a}_{cm}) = E(\hat{a}_{cm}^2) - \left(\sum_{i=1}^D u_{1,i}u_{2,i} + \frac{1}{k}\sum_{i\neq j} u_{1,i}u_{2,j}\right)^2\\\notag
=&\left(\frac{1}{k}-\frac{1}{k^2}\right)\sum_{i\neq j} u_{1,i}^2u_{2,j}^2+u_{1,i}u_{2,i}u_{1,j}u_{2,j}\\\notag
=&\frac{1}{k}\left(1-\frac{1}{k}\right)\left[\sum_{i=1}^Du_{1,i}^2 \sum_{i=1}^Du_{2,i}^2 + \left(\sum_{i=1}^Du_{1,i}u_{2,i}\right)^2 - 2\sum_{i=1}^Du_{1,i}^2u_{2,i}^2\right]
\end{align}

\subsection{The Analysis of the More General Version of the VW Algorithm}

The nice approach proposed in the VW paper~\cite{Proc:Weinberger_ICML2009} is to pre-element-wise-multiply the data vectors with a random vector $r$ before taking the hashing operation. We denote the two resultant vectors (samples) by $g_1$ and $g_2$ respectively:
\begin{align}\notag
g_{1,q} = \sum_{i=1}^D u_{1,i} r_i I_{iq}, \hspace{0.5in} g_{2,q} = \sum_{i=1}^D u_{2,i} r_i I_{iq}
\end{align}
where $r_i \in \{-1, 1\}$ with equal probabilities. Here, we provide a more general scheme by sampling $r_i$ from a sub-Gaussian distribution with parameter $s$ and
\begin{align}\notag
E(r_i) = 0, \ E(r_i^2) = 1, \ E(r_i^3) = 0, \ E(r_i^4) = s
\end{align}
which include normal (i.e., $s=3$) and the distribution on $\{-1,\ 1\}$ with equal probabilities (i.e., $s=1$) as special cases.

Let $\hat{a}_{vw,s} = \sum_{q=1}^k g_{1,q} g_{2,q}$. The goal is to show
\begin{align}\notag
&E(\hat{a}_{vw,s}) = \sum_{i=1}^D u_{1,i}u_{2,i}\\\notag
&Var(\hat{a}_{vw,s}) = (s-1)\sum_{i=1}^Du_{1,i}^2u_{2,i}^2+\frac{1}{k}\left[\sum_{i=1}^D u_{1,i}^2\sum_{i=1}^D u_{2,i}^2 + \left(\sum_{i=1}^D u_{1,i}u_{2,i}\right)^2 - 2\sum_{i=1}^D u_{1,i}^2u_{2,i}^2\right]
\end{align}

We can use the previous results and the conditional expectation and variance formulas:
\begin{align}\notag
E(X) = E(E(X|Y)), \hspace{0.5in} Var(X) = E(Var(X|Y)) + Var(E(X|Y))
\end{align}

$E(r_i) = 0$, $E(r_i^2) =1$, $E(r_i^3) = 0$, $E(r_i^4)=s$.

\begin{align}\notag
&E(\hat{a}_{vw,s}) =E(E(\hat{a}_{vw,s}|r)) = E\left(\sum_{i=1}^D u_{1,i} u_{2,i} r_i^2 + \frac{1}{k}\sum_{i\neq j} u_{1,i} u_{2,j} r_i r_j\right) = \sum_{i=1}^D u_{1,i} u_{2,i}\\\notag
&E(Var(\hat{a}_{vw,s}|r)) = \left(\frac{1}{k}-\frac{1}{k^2}\right) E\left(\sum_{i\neq j} u_{1,i}^2u_{2,j}^2r_i^2r_j^2+u_{1,i}u_{2,i}u_{1,j}u_{2,j}r_i^2r_j^2\right)
=\left(\frac{1}{k}-\frac{1}{k^2}\right)\sum_{i\neq j} u_{1,i}^2u_{2,j}^2+u_{1,i}u_{2,i}u_{1,j}u_{2,j}
\end{align}
As $Var(E(\hat{a}_{vw,s}|r)) = E(E^2(\hat{a}_{vw,s}|r))-E^2(E(\hat{a}_{vw,s}|r)) $, we need to compute
\begin{align}\notag
&E(E^2(\hat{a}_{vw,s}|r)) = E\left(\sum_{i=1}^D u_{1,i} u_{2,i} r_i^2 + \frac{1}{k}\sum_{i\neq j} u_{1,i} u_{2,j} r_i r_j\right)^2\\\notag
=&s\sum_{i=1}^D u_{1,i}^2u_{2,i}^2 + \sum_{i\neq j} u_{1,i}u_{2,i}u_{1,j}u_{2,j}+\frac{1}{k^2}\sum_{i\neq j}u_{1,i}^2u_{2,j}^2+u_{1,i}u_{2,i}u_{1,j}u_{2,j}
\end{align}
Thus,
\begin{align}\notag
&Var(E(\hat{a}_{vw,s}|r)) = (s-1)\sum_{i=1}^D u_{1,i}^2u_{2,i}^2 +\frac{1}{k^2}\sum_{i\neq j}u_{1,i}^2u_{2,j}^2+u_{1,i}u_{2,i}u_{1,j}u_{2,j}\\\notag
&Var(\hat{a}_{vw,s}) = (s-1)\sum_{i=1}^D u_{1,i}^2u_{2,i}^2+\frac{1}{k}\sum_{i\neq j} u_{1,i}^2u_{2,j}^2+u_{1,i}u_{2,i}u_{1,j}u_{2,j}
\end{align}

\subsection{Another Simple Scheme for Bias-Correction}

By examining the expectation (\ref{eqn_mean_cm}), the bias of CM can be easily removed, because
\begin{align}\label{eqn_est_cm_nb}\vspace{-0.15in}
\hat{a}_{cm,nb} =\frac{k}{k-1} \left[\hat{a}_{cm} - \frac{1}{k}\sum_{i=1}^Du_{1,i}\sum_{i=1}^Du_{2,i}\right]
\end{align}
 is unbiased with variance
\begin{align}\label{eqn_var_cm_nb}
Var(\hat{a}_{cm,nb}) = \frac{1}{k-1}\left[\sum_{i=1}^D u_{1,i}^2\sum_{i=1}^D u_{2,i}^2 + \left(\sum_{i=1}^D u_{1,i}u_{2,i}\right)^2-2\sum_{i=1}^D u_{1,i}^2u_{2,i}^2\right],
\end{align}
which is essentially the same as the variance of VW.

\section{Comparing  $b$-Bit Minwise Hashing with VW Random Projections}\label{app_compare_VW}

We compare VW (and random projections) with $b$-bit minwise hashing for the task of estimating inner products on binary data.  With binary data, i.e., $u_{1,i}, u_{2,i}\in \{0,1\}$, we have $f_1 = \sum_{i=1}^D u_{1,i}$, $f_2 = \sum_{i=1}^D u_{2,i}$, $a = \sum_{i=1}^D u_{1,i}u_{2,i}$. The variance (\ref{eqn_var_vw}) (by using $s=1$) becomes
\begin{align}\notag
Var\left(\hat{a}_{vw,s=1}\right) = \frac{f_1f_2+a^2-2a}{k}
\end{align}

We can compare this variance with the variance of $b$-bit minwise hashing. Because the variance (\ref{eqn_Var_b}) is for estimating the resemblance, we need to convert it into the variance for estimating the inner product $a$ using the relation:
 \begin{align}\notag
 a = \frac{R}{1+R}(f_1+f_2)
 \end{align}

We can estimate $a$ from the estimated $R$,
\begin{align}\notag
\hat{a}_{b} =& \frac{\hat{R}_b}{1+\hat{R}_b}(f_1+f_2)\\\notag
\text{Var}\left(\hat{a}_{b}\right) =& \left[\frac{1}{(1+R)^2}(f_1+f_2)\right]^2\text{Var}(\hat{R}_b)
\end{align}

For $b$-bit minwise hashing, each sample is stored using only $b$ bits. For VW (and random projections), we assume  each sample is stored using 32 bits (instead of 64 bits) for two reasons: (i) for binary data, it would be very unlikely for the hashed value to be close to $2^{64}$, even when $D=2^{64}$; (ii) unlike $b$-bit minwise hashing, which requires exact bit-matching in the estimation stage, random projections only need to compute the inner products for which it would suffice to store  hashed values as (double precision) real numbers.

Thus, we define the following  ratio to compare the two methods.
\begin{align}\label{eqn_G_vw}
G_{vw} =& \frac{Var\left(\hat{a}_{vw,s=1}\right)\times 32}{Var\left(\hat{a}_{b}\right)\times b}
\end{align}

If $G_{vw}>1$, then $b$-bit minwise hashing is more accurate than binary random projections. Equivalently, when $G_{vw}>1$, in order to achieve the same level of accuracy (variance), $b$-bit minwise hashing needs smaller storage space than random projections. There are two issues we need to elaborate on:
\begin{enumerate}\notag
\item Here, we assume the purpose of using VW is for {\em data reduction}. That is, $k$ is small compared to the number of non-zeros (i.e., $f_1$, $f_2$). We do not consider the case when $k$ is taken to be extremely large for the benefits of {\em compact indexing} without achieving {\em data reduction}.
\item Because we assume $k$ is small, we need to represent the sample with enough precision. That is why we assume each sample of VW is stored using 32 bits. In fact, since the ratio $G_{vw}$ is usually very large (e.g., $10\sim 100$) by using 32 bits for each VW sample, it will remain to be very large (e.g., $5\sim 50$) even if we only need to store each VW sample using 16 bits.
\end{enumerate}

\begin{figure}[h!]
\begin{center}
\mbox{
{\includegraphics[width = 2  in]{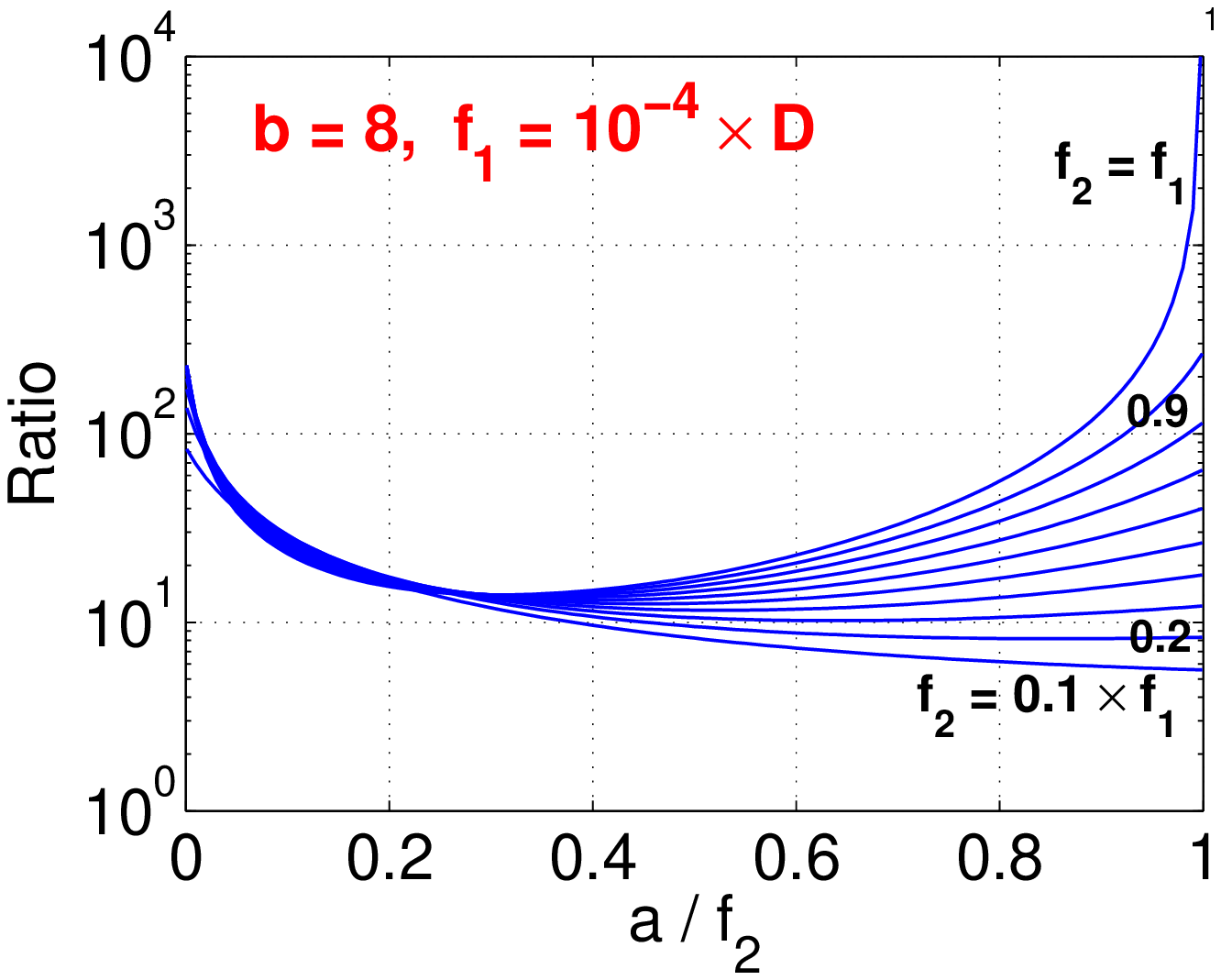}}
{\includegraphics[width = 2  in]{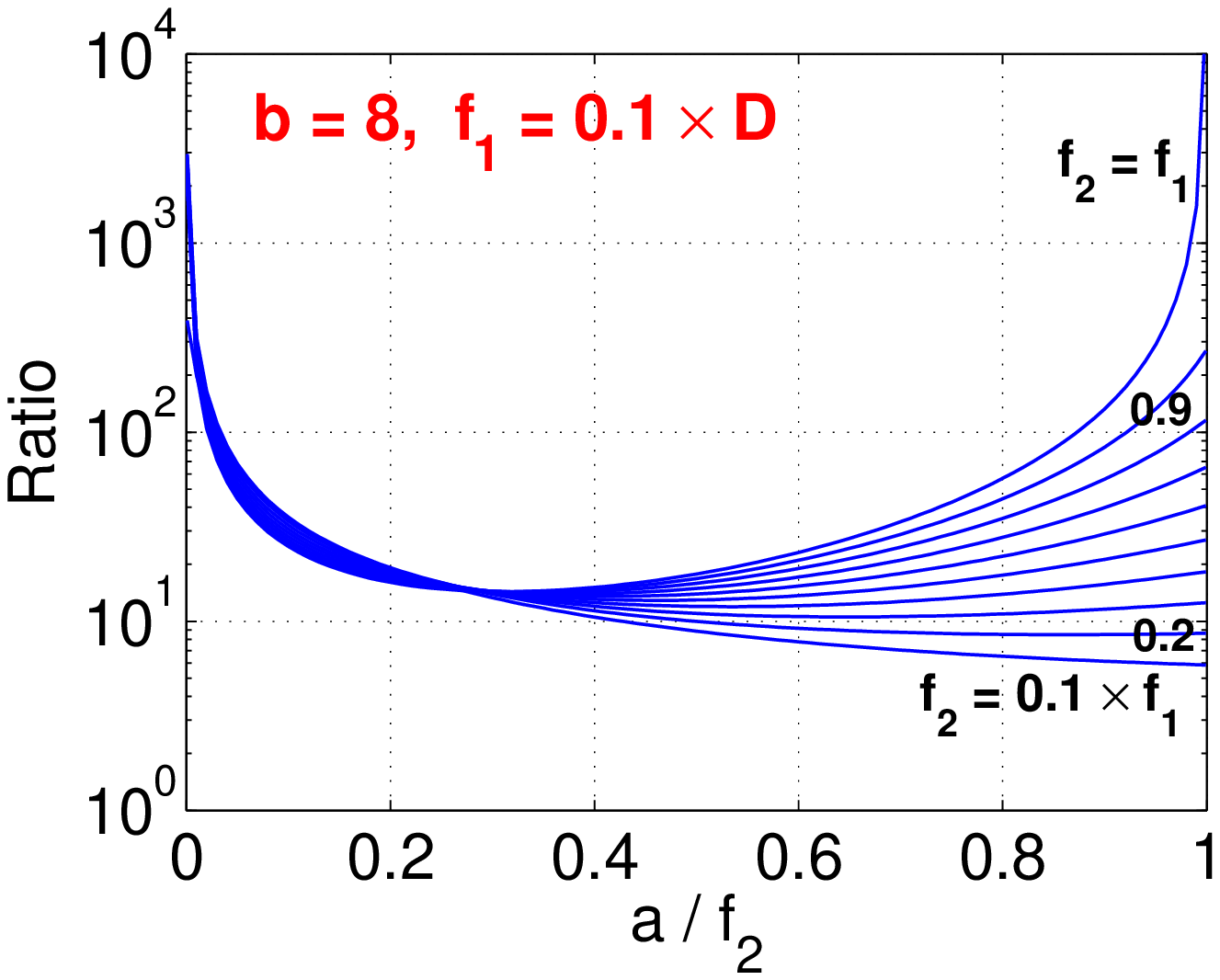}}}

\mbox{
{\includegraphics[width = 2  in]{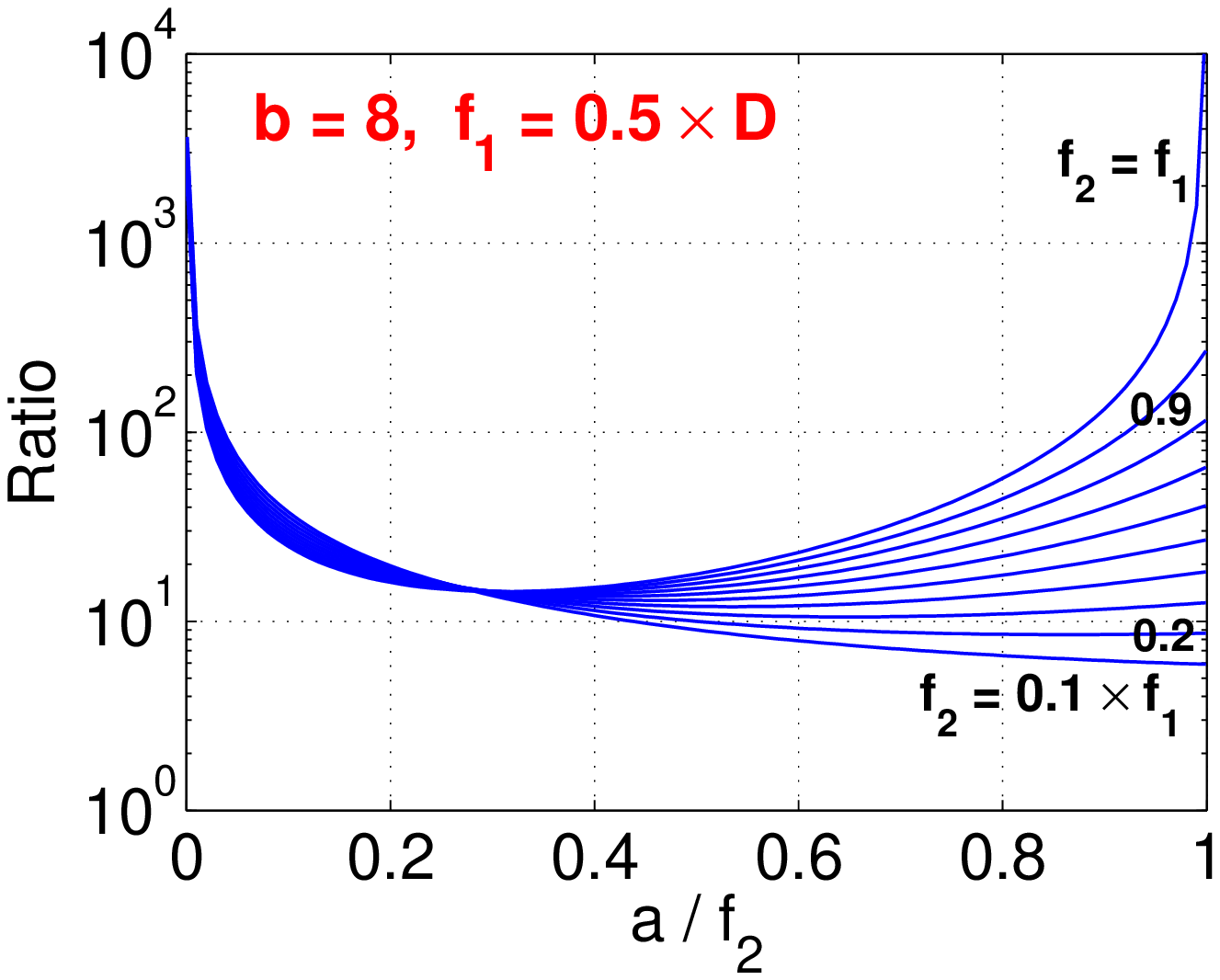}}
{\includegraphics[width = 2  in]{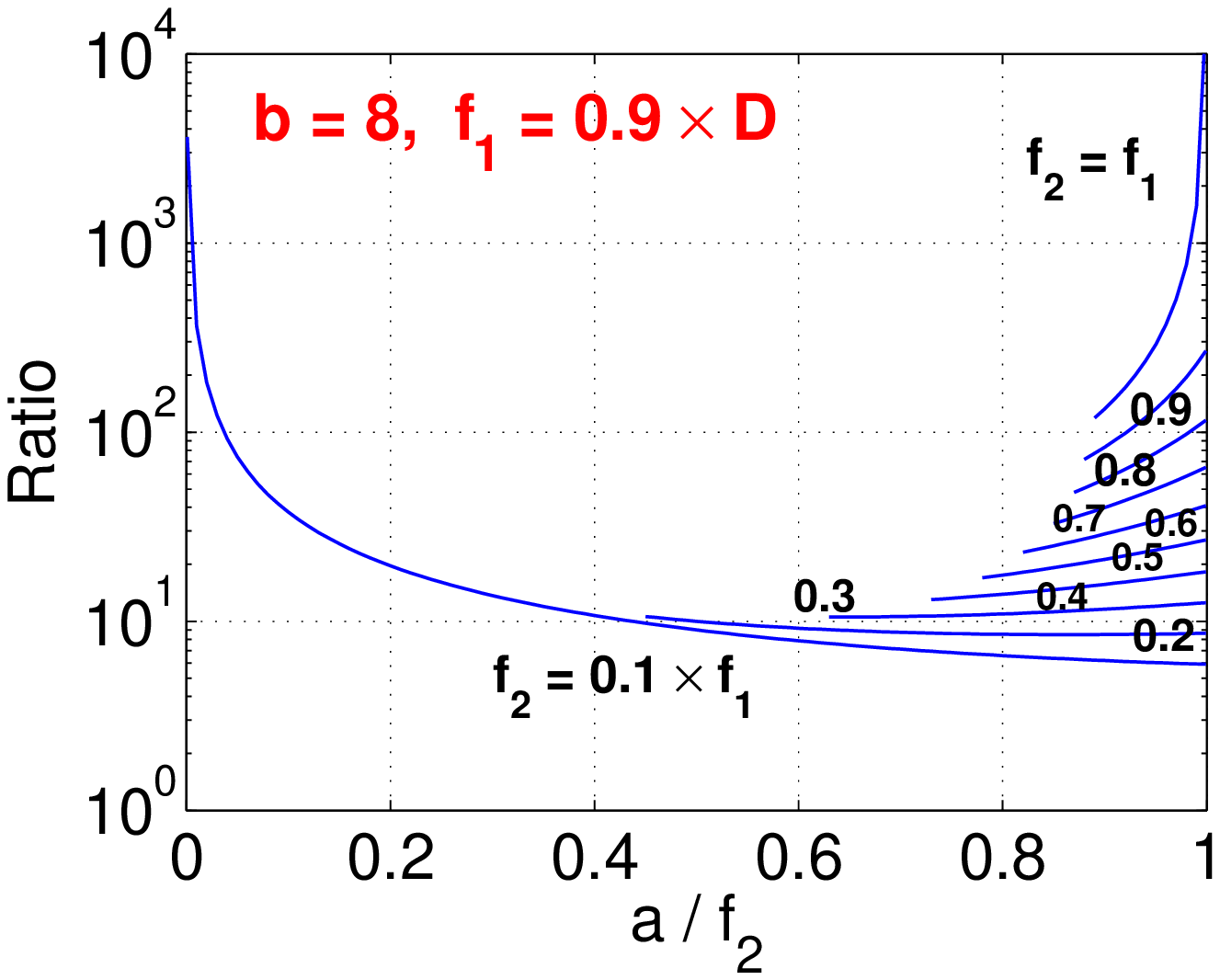}}}

\vspace{-0.3in}
\end{center}
\caption{$G_{vw}$ as defined in (\ref{eqn_G_vw}), for $b=8$. $G_{vw}>1$ means $b$-bit minwise hashing  is more accurate than random projections at the same storage cost. We consider four  $f_1$ values. For each $f_1$, we let $f_2 = 0.1f_1, 0.2f_1, ..., f_1$ and $a = 0$ to $f_2$. }\label{fig_Gb8}
\end{figure}

\begin{figure}[h!]
\begin{center}
\mbox{
{\includegraphics[width = 2  in]{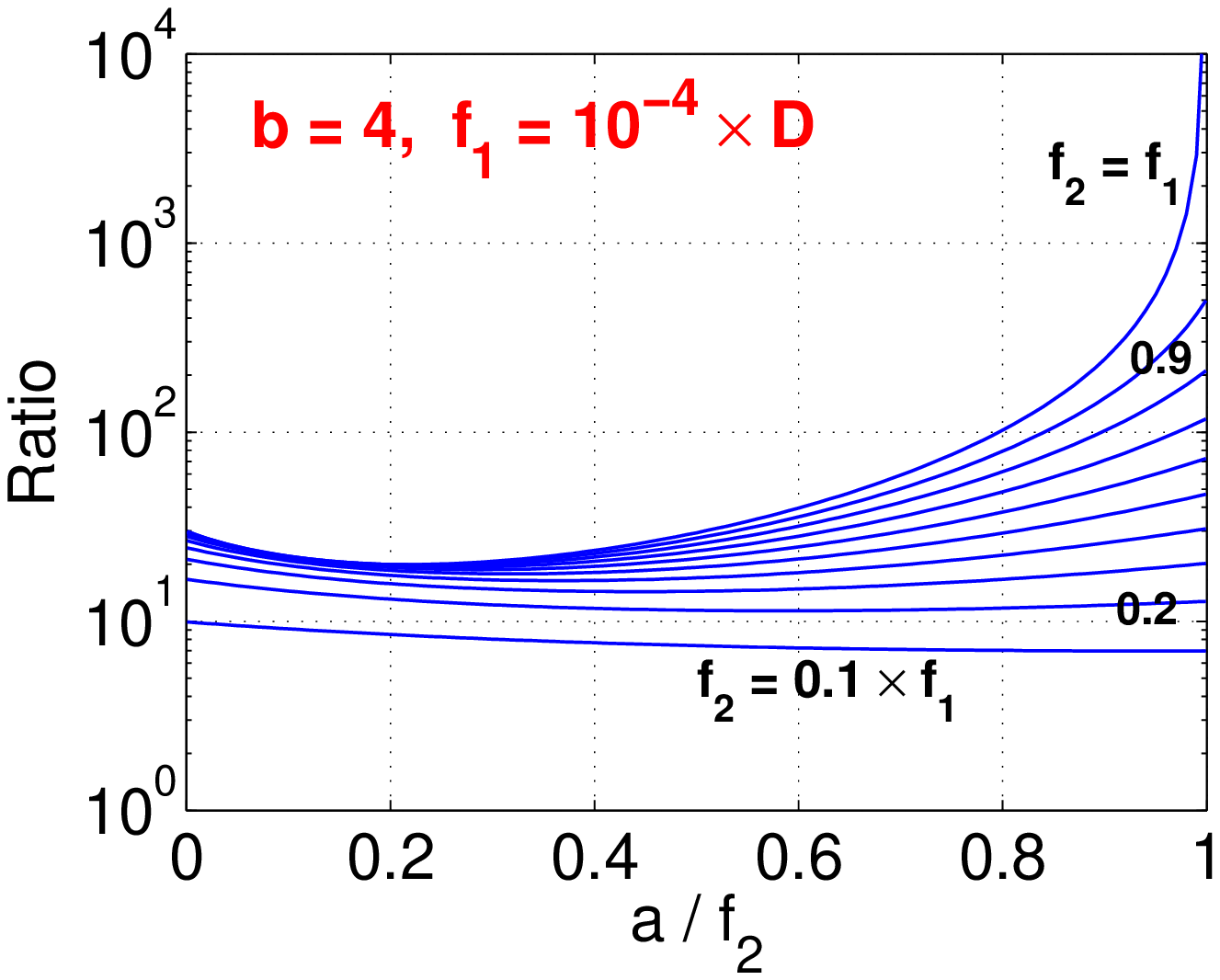}}
{\includegraphics[width = 2  in]{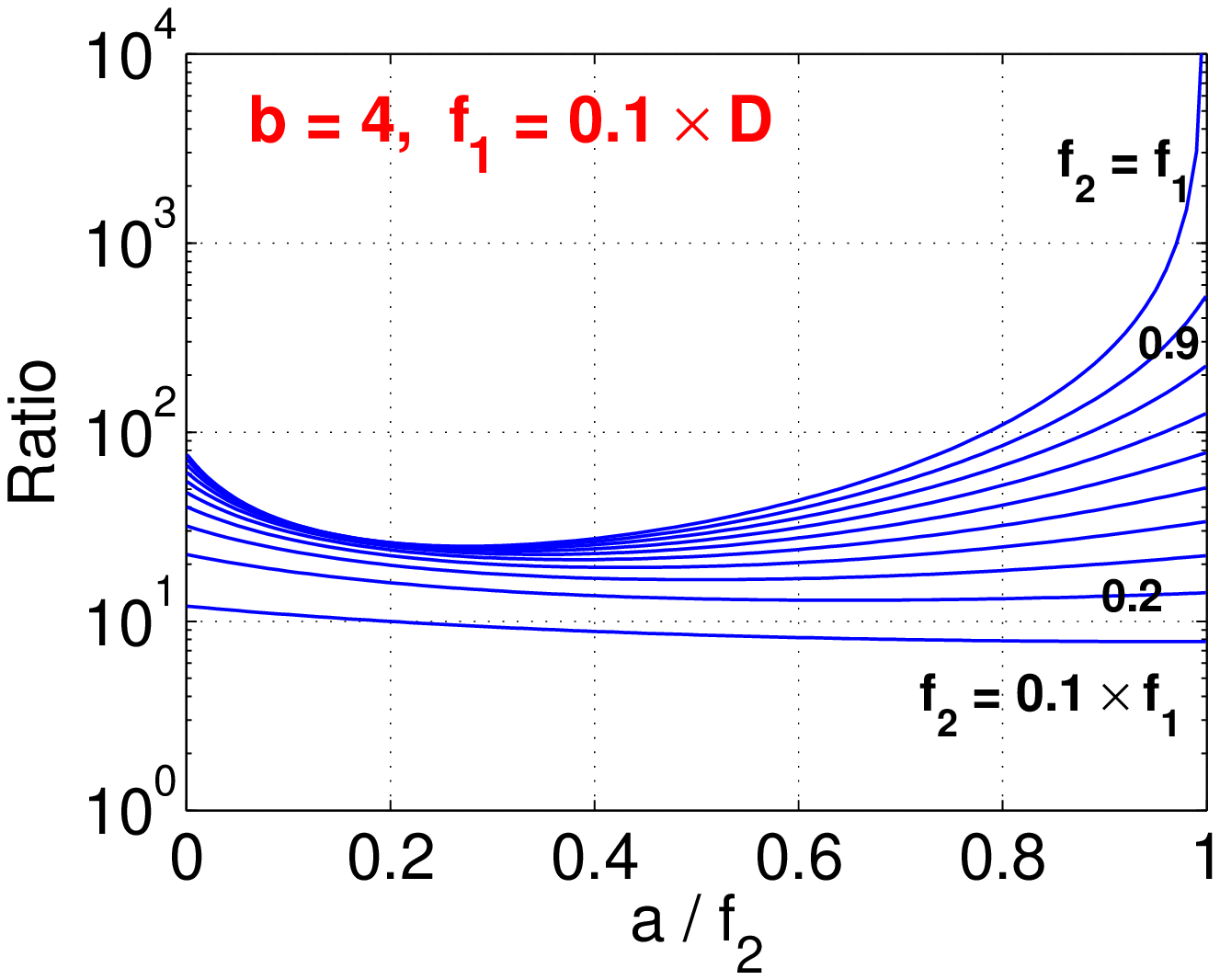}}}

\mbox{
{\includegraphics[width = 2  in]{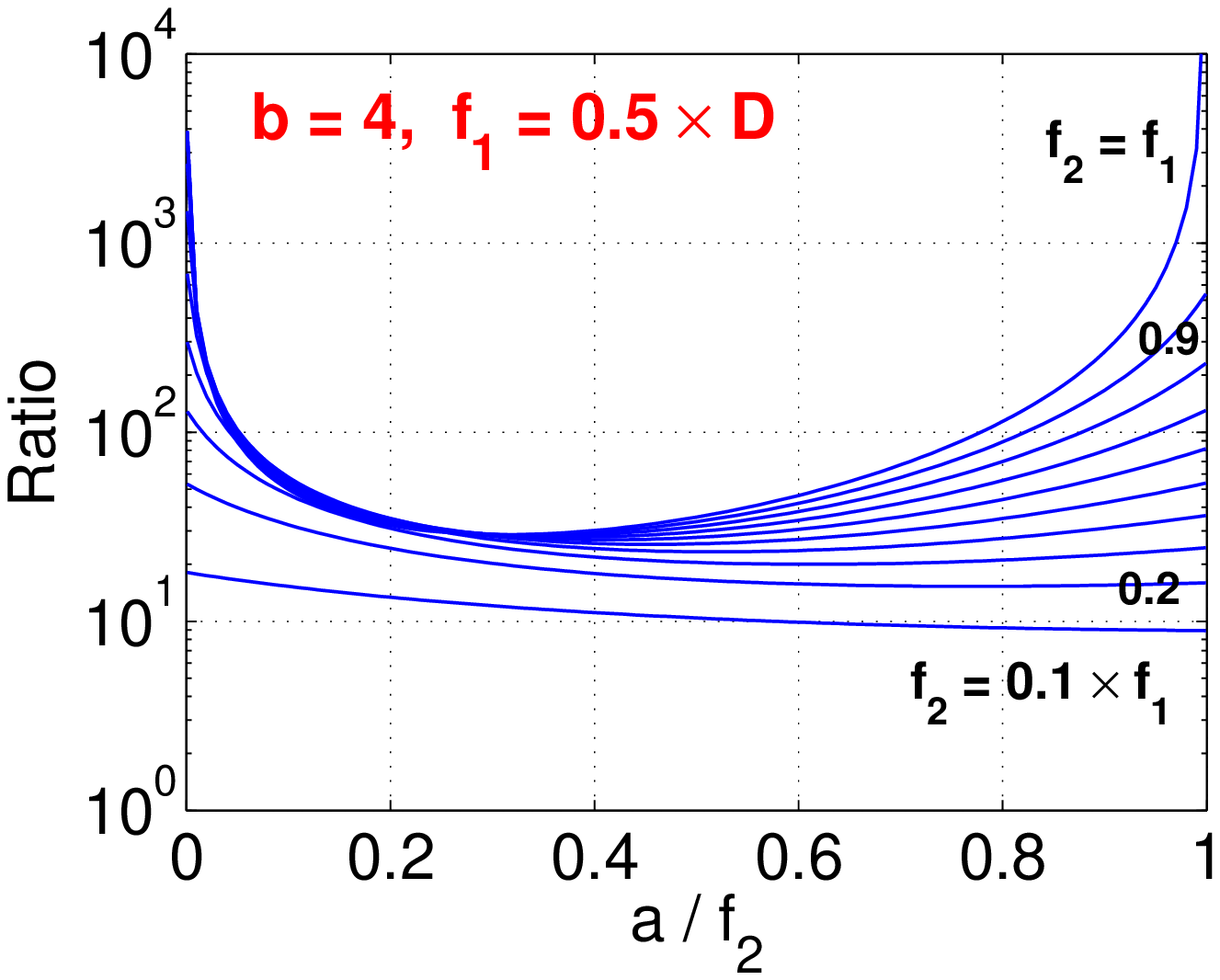}}
{\includegraphics[width = 2  in]{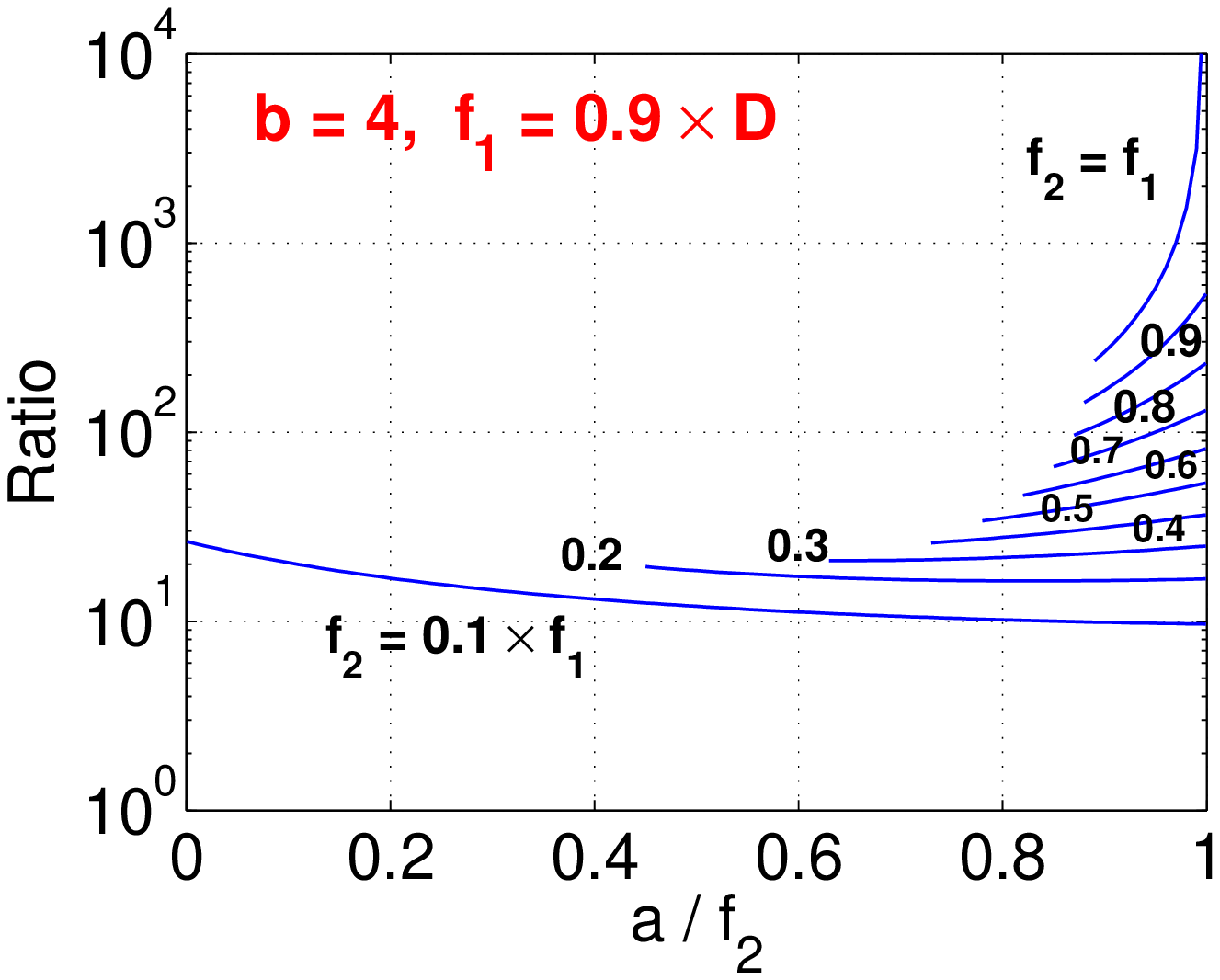}}}

\vspace{-0.3in}
\end{center}
\caption{$G_{vw}$ as defined in (\ref{eqn_G_vw}), for $b=4$. They again indicate the very substantial improvement of $b$-bit minwise hashing over random projections. }\label{fig_Gb4}
\end{figure}

\begin{figure}[h!]
\begin{center}
\mbox{
{\includegraphics[width = 2.0  in]{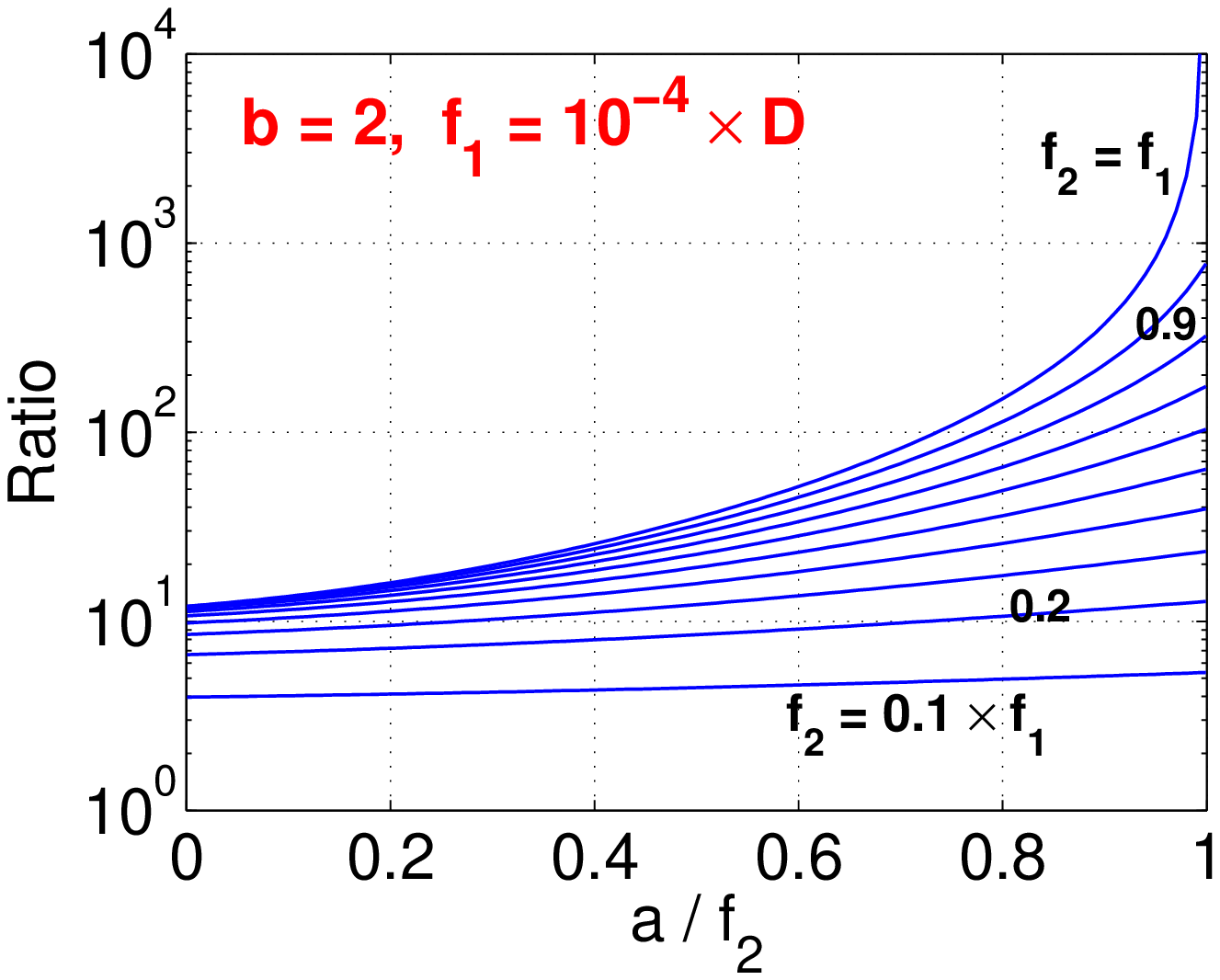}}
{\includegraphics[width = 2.0  in]{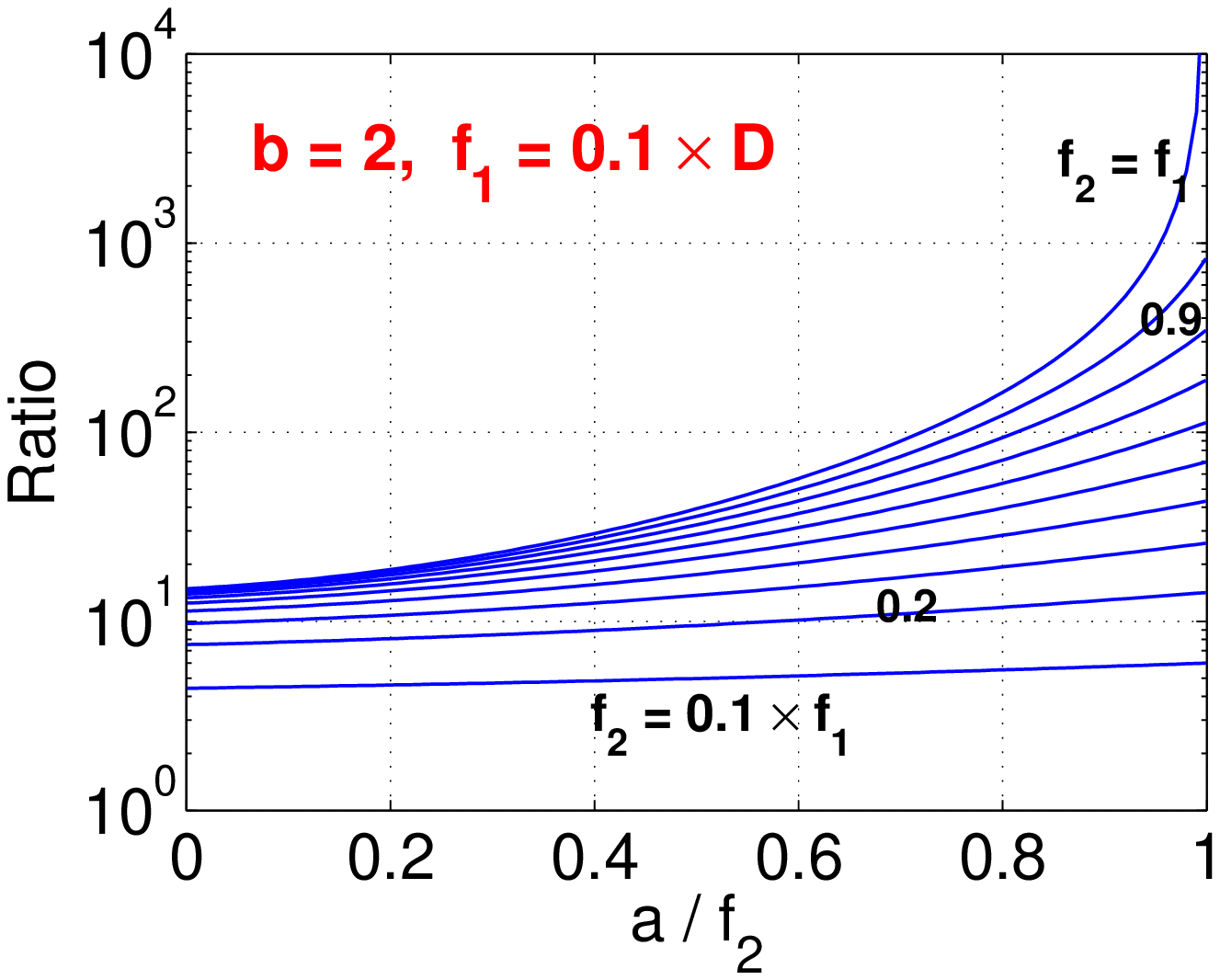}}}

\mbox{
{\includegraphics[width = 2.0  in]{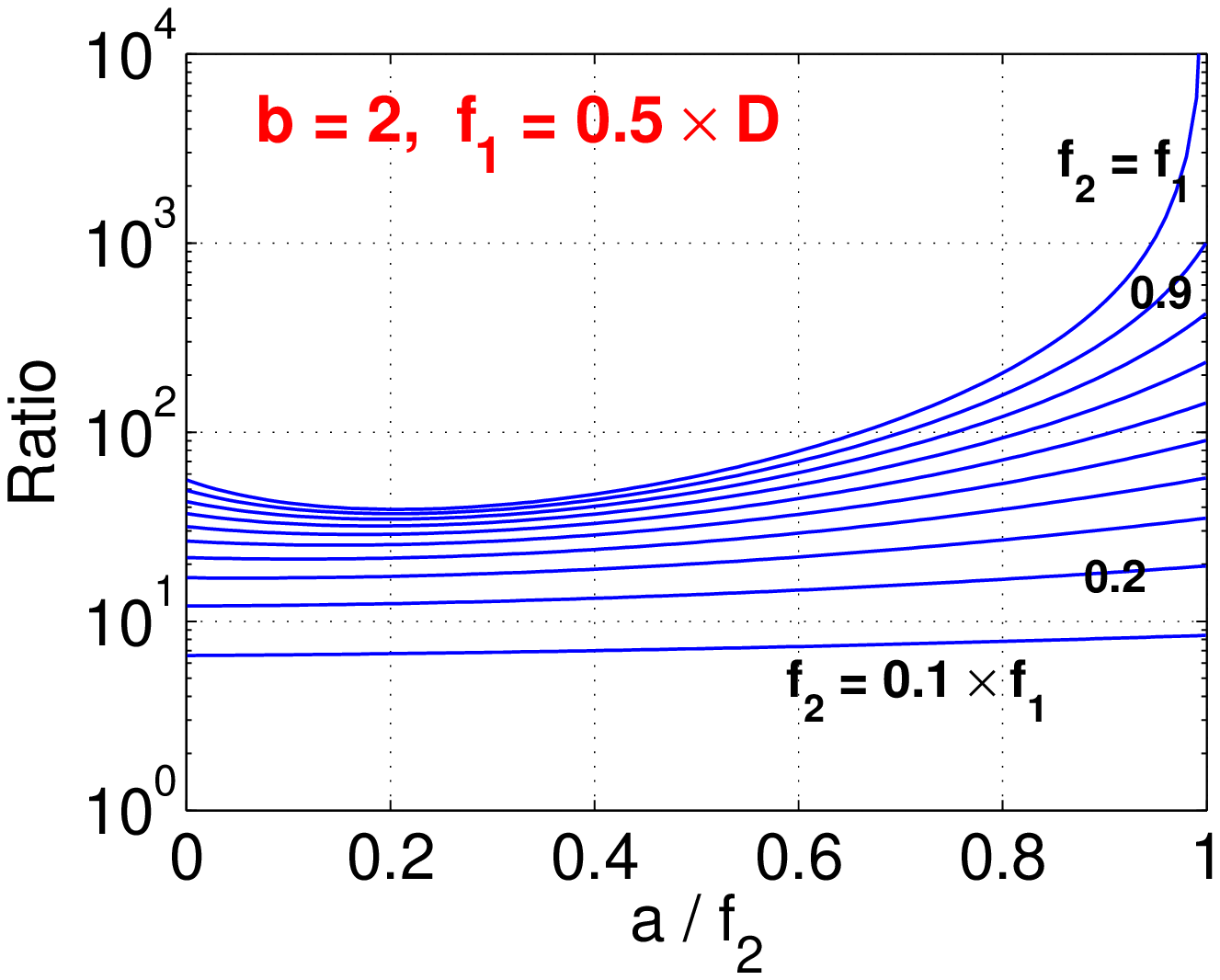}}
{\includegraphics[width = 2.0  in]{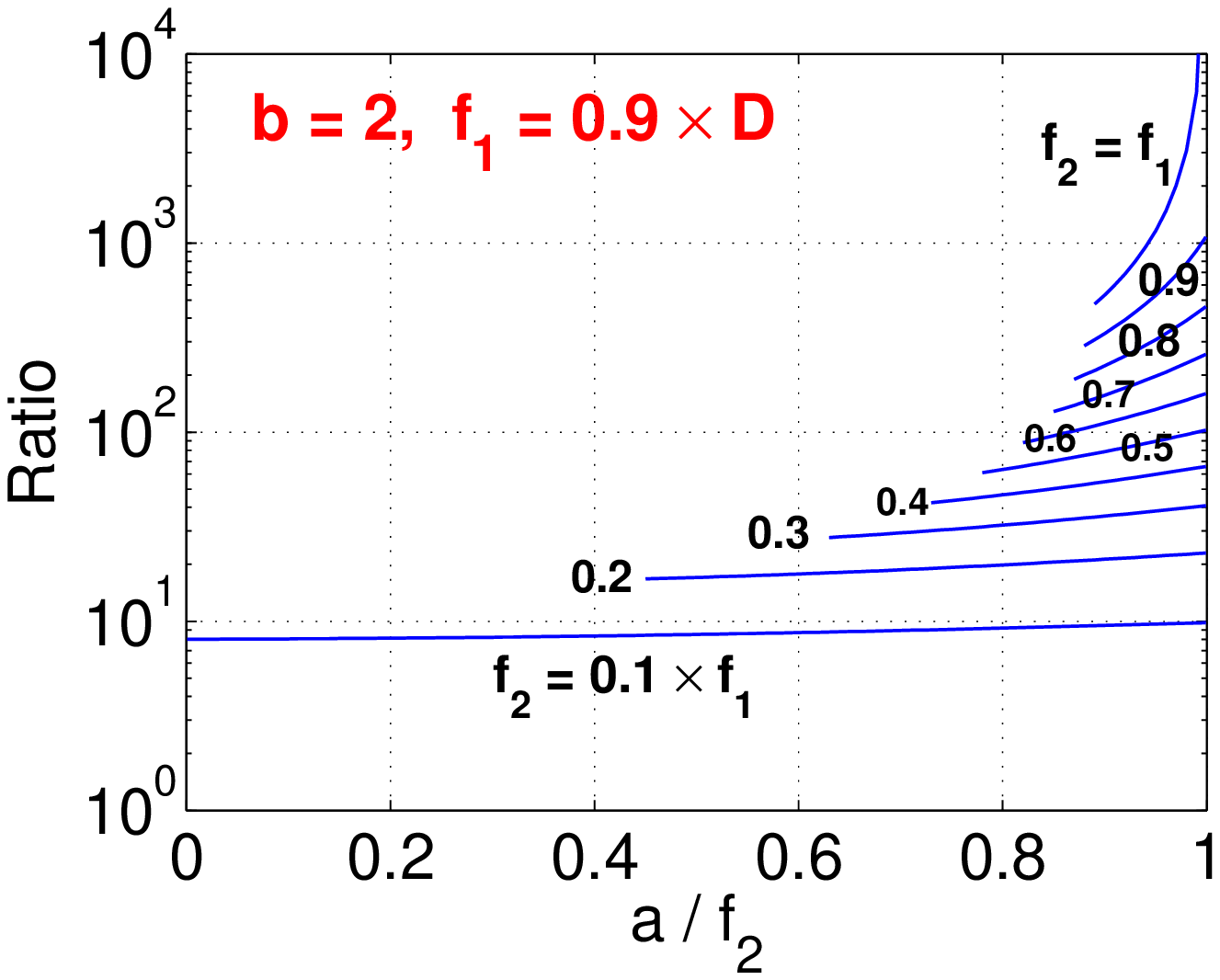}}}

\vspace{-0.3in}
\end{center}
\caption{$G_{vw}$ as defined in (\ref{eqn_G_vw}), for $b=2$.}\label{fig_Gb2}
\end{figure}

\begin{figure}[h!]
\begin{center}
\mbox{
{\includegraphics[width = 2.0  in]{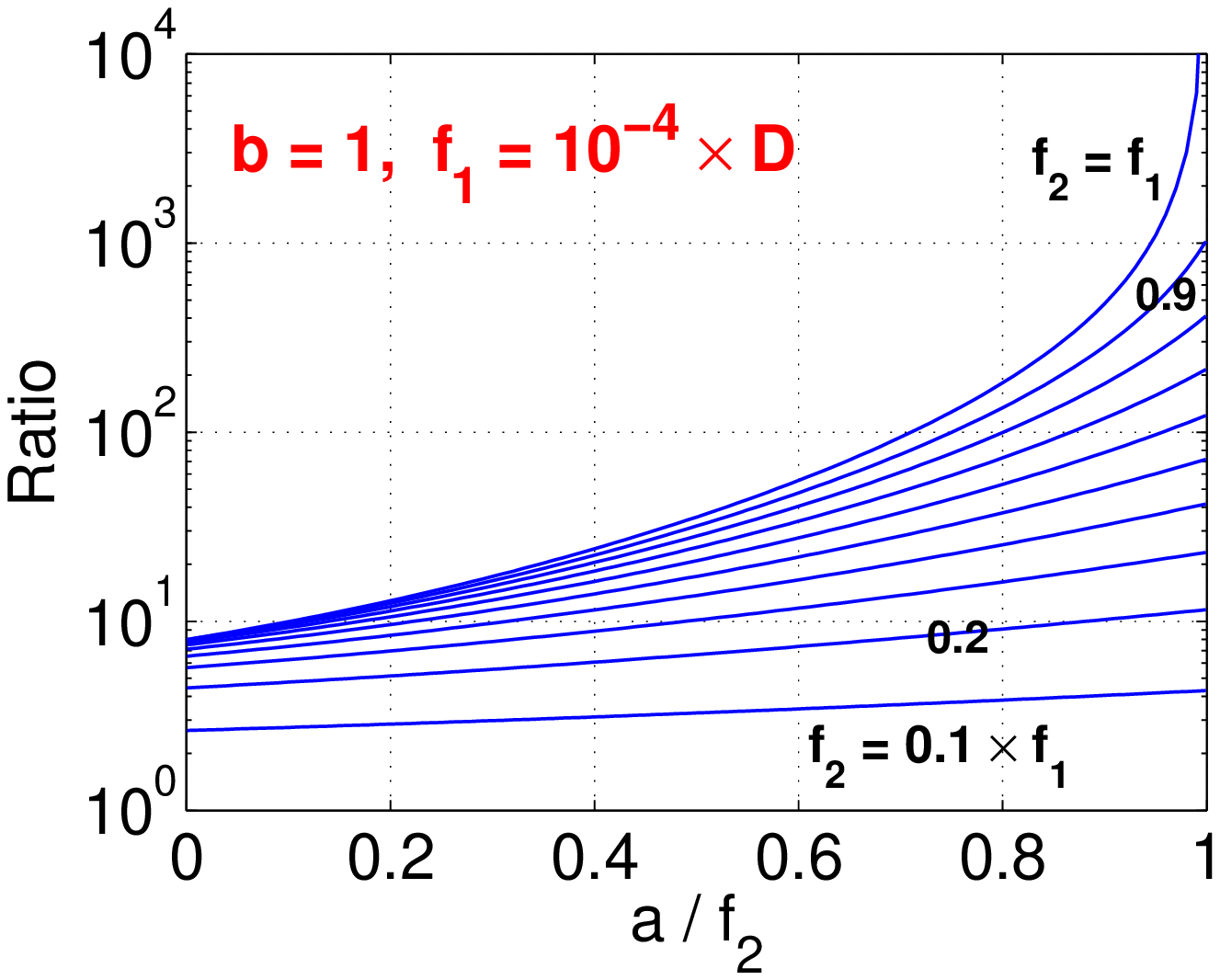}}
{\includegraphics[width = 2.0  in]{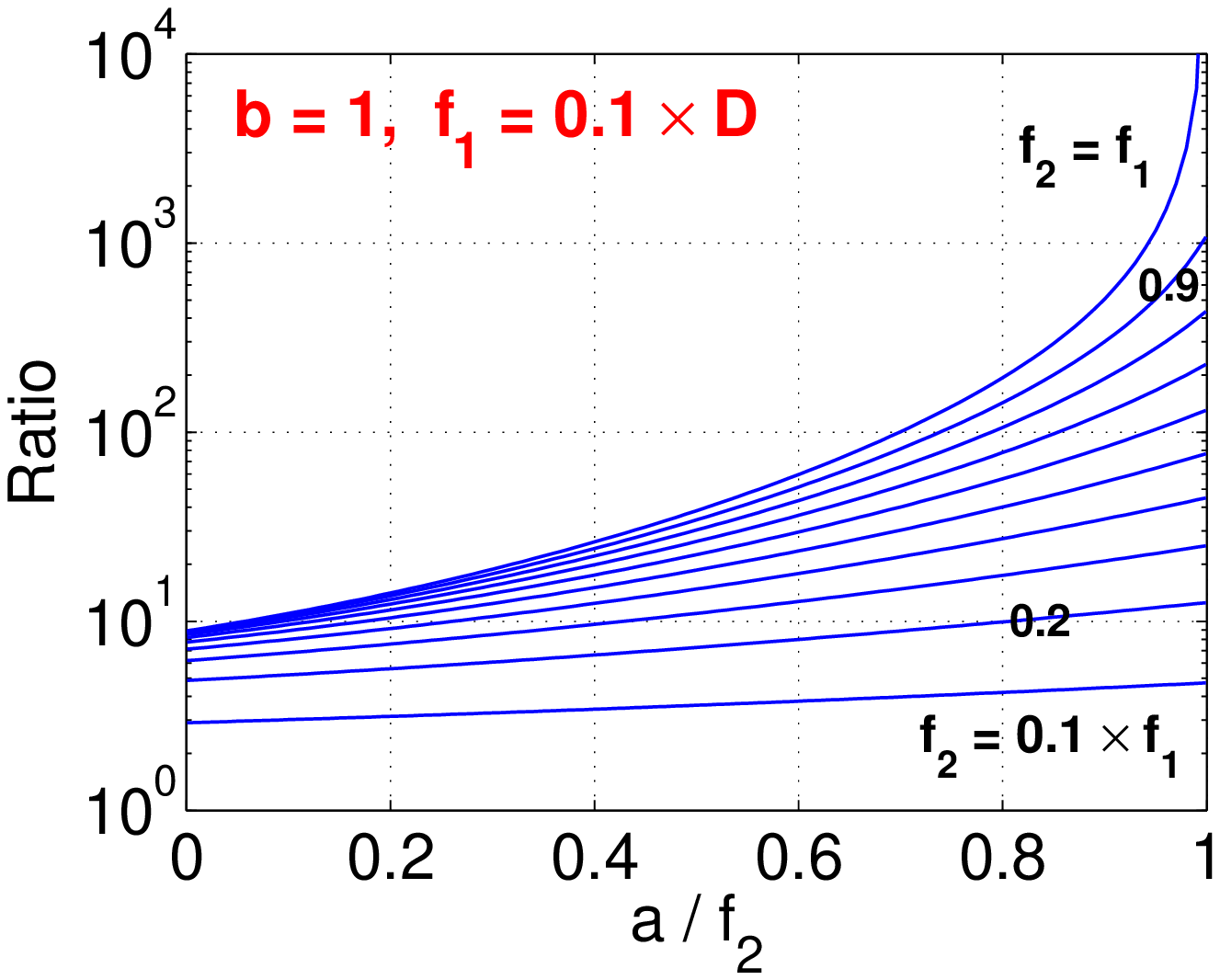}}}

\mbox{
{\includegraphics[width = 2.0  in]{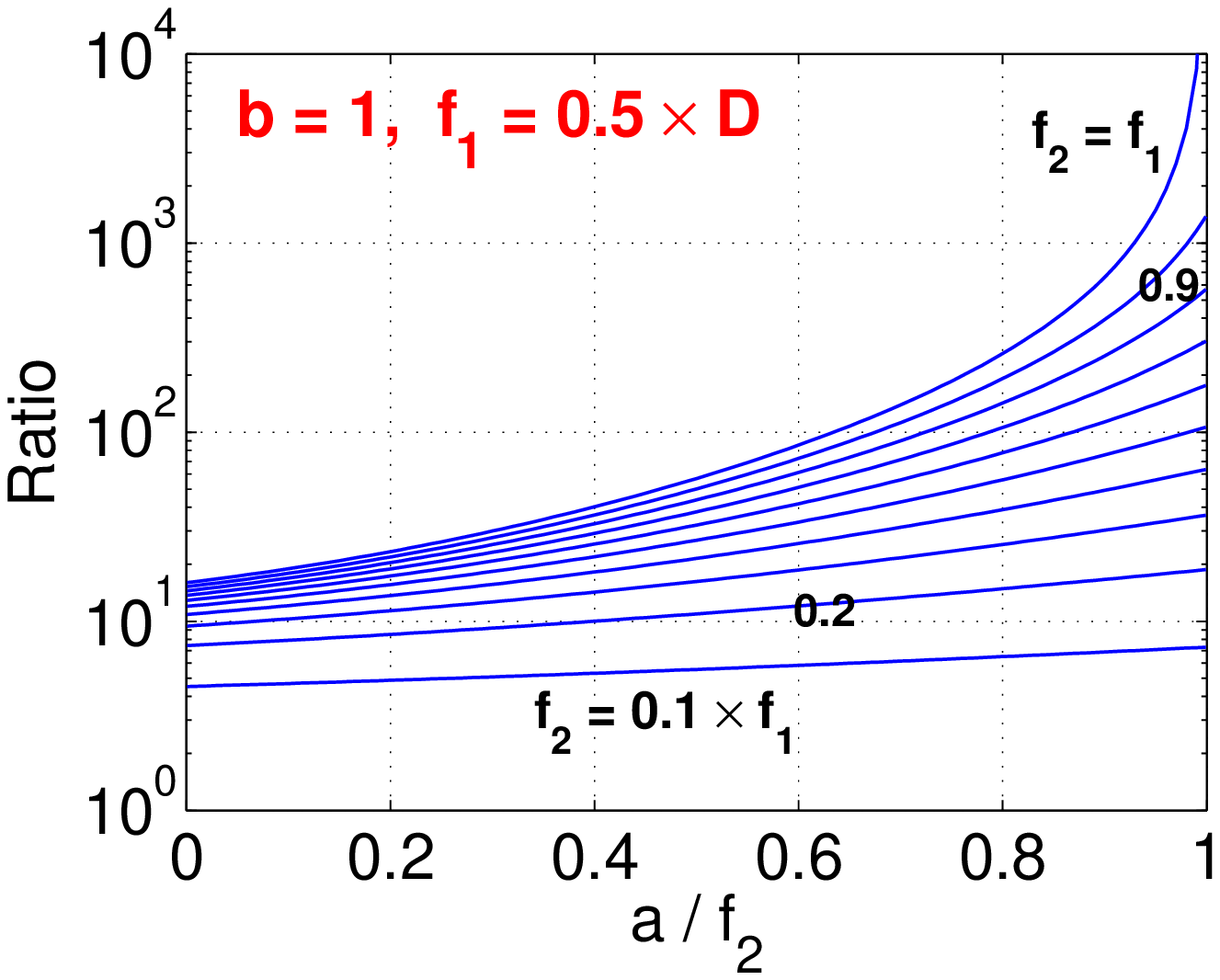}}
{\includegraphics[width = 2.0  in]{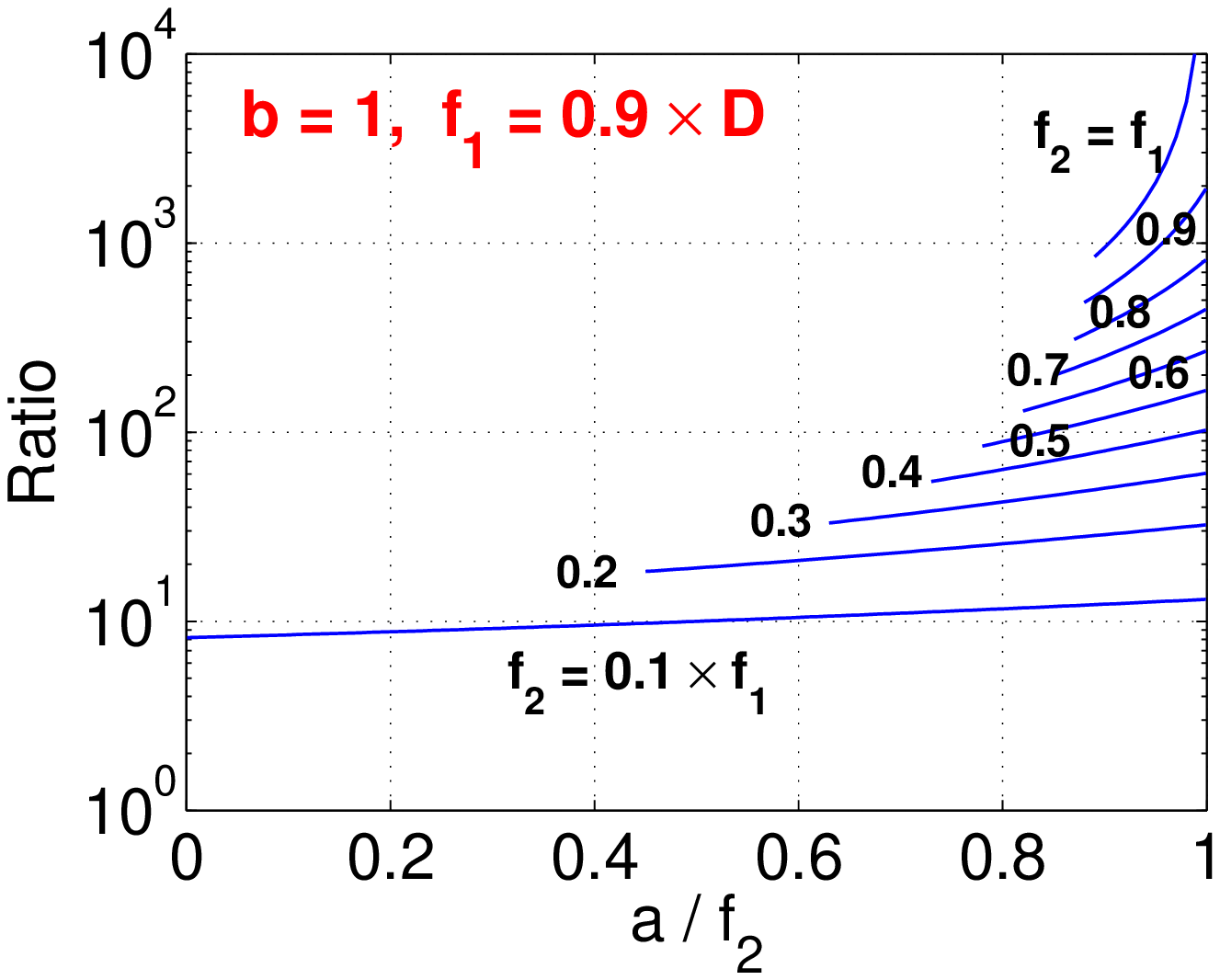}}}

\vspace{-0.3in}
\end{center}
\caption{$G_{vw}$ as defined in (\ref{eqn_G_vw}), for $b=1$. }\label{fig_Gb1}
\end{figure}

Without loss of generality, we can assume $f_2\leq f_1$ (hence $a\leq f_2 \leq f_1$).  Figures~\ref{fig_Gb8} to~\ref{fig_Gb1} display the ratios (\ref{eqn_G_vw}) for $b=8, 4, 2, 1$, respectively. In order to achieve high learning accuracies, $b$-bit minwise hashing requires  $b=4$ (or even 8). In each figure, we plot $G_{BP}$ for  $f_1/D = 0.0001, 0.1, 0.5, 0.9$ and full ranges of $f_2$ and $a$. We can see that $G_{vw}$ is much larger than one (usually 10 to 100), indicating the very substantial advantage of $b$-bit minwise hashing over random projections.

Note that the comparisons are essentially independent of $D$. This is because in the variance of binary random projection (\ref{eqn_G_vw}) the $-2a$ term is negligible compared to $a^2$ in binary data as $D$ is very large. To generate the plots, we used $D=10^6$ (although practically $D$ should be much larger).\\

\noindent\textbf{Conclusion}: \ Our theoretical analysis has illustrated the  substantial improvements of $b$-bit minwise hashing over the VW algorithm and random projections in binary data, often by 10- to $100$-fold. We feel such a large performance difference should be noted by researchers and practitioners in large-scale machine learning.

\end{document}